\documentclass[lettersize,journal]{IEEEtran}
\usepackage{amsmath,amsfonts}
\usepackage{algorithmic}
\usepackage{array}
\usepackage[caption=false,font=normalsize,labelfont=sf,textfont=sf]{subfig}
\usepackage{textcomp}
\usepackage{stfloats}
\usepackage{url}
\usepackage{verbatim}
\usepackage{graphicx}
\usepackage[table]{xcolor}
\usepackage{xspace}

\usepackage{cite}
\definecolor{citecolor}{HTML}{0071bc}
\definecolor{shadecolor}{HTML}{EFEFEF}
\usepackage[colorlinks=true,linkcolor=black,citecolor=black,urlcolor=black]{hyperref}
\usepackage[capitalize,noabbrev]{cleveref}

\usepackage{amsmath,amsfonts,bm}

\def\eqref#1{equation~\ref{#1}}

\def\1{\bm{1}}

\def\rvs{{\boldsymbol{s}}}
\def\rva{{\boldsymbol{a}}}

\DeclareMathAlphabet{\mathsfit}{\encodingdefault}{\sfdefault}{m}{sl}
\SetMathAlphabet{\mathsfit}{bold}{\encodingdefault}{\sfdefault}{bx}{n}

\usepackage{tablefootnote}

\usepackage{tabularray}
\UseTblrLibrary{booktabs}
\usepackage{multirow}

\usepackage{algorithm}
\usepackage{enumitem}

\usepackage{amsmath}
\usepackage{amssymb}
\usepackage{mathtools}
\usepackage{amsthm}

\newcommand{\etal}{\textit{et al}.~}
\newcommand{\ie}{\textit{i}.\textit{e}.}

\newcommand{\etc}{\textit{etc}.}

\theoremstyle{plain}
\newtheorem{theorem}{Theorem}[section]

\newtheorem{lemma}[theorem]{Lemma}

\theoremstyle{definition}

\theoremstyle{remark}

\newcommand{\name}{Offline Decoupled Prioritized Resampling\xspace}
\newcommand{\shortname}{ODPR\xspace}
\newcommand{\algo}{Advantage-based Offline Decoupled Prioritized Resampling\xspace}
\newcommand{\shortalgo}{ODPR-A\xspace}
\newcommand{\algoo}{Return-based Offline Decoupled Prioritized Resampling\xspace}
\newcommand{\shortalgoo}{ODPR-R\xspace}
\newcommand{\YY}[1]{#1}

\def\BibTeX{{\rm B\kern-.05em{\sc i\kern-.025em b}\kern-.08em
    T\kern-.1667em\lower.7ex\hbox{E}\kern-.125emX}}
\usepackage{balance}

\begin{document}

\title{Decoupled Prioritized Resampling for Offline RL}

\author{
    Yang Yue, 
    Bingyi Kang, 
    Xiao Ma, 
    Qisen Yang,
    Gao Huang, \textit{Member, IEEE}, \\
    Shiji Song, \textit{Senior Member, IEEE}, 
    and Shuicheng Yan, \textit{Fellow, IEEE}    
\thanks{
Yang Yue, Qisen Yang, Gao Huang, and Shiji Song are with the Department of Automation, Tsinghua University, Beijing 100084, China. Bingyi Kang, Xiao Ma, and Shuicheng Yan are with Sea AI Lab, Singapore. Email: 
\{yueyang22f,bingykang,yusufma555\}@gmail.com,
yangqs19@mails.tsinghua.edu.cn,
\{shijis, gaohuang\}@tsinghua.edu.cn,
\{yansc\}@sea.com. }
\thanks{\textit{Corresponding author: Shiji Song.}}
}

\maketitle

\vspace{-2mm}
\begin{abstract}

Offline reinforcement learning (RL) is challenged by the distributional shift problem.
To tackle this issue, existing works mainly focus on designing sophisticated policy constraints between the learned policy and the behavior policy.
However, these constraints are applied equally to well-performing and inferior actions through uniform sampling, which might negatively affect the learned policy. 
In this paper, we propose \emph{\name} (\shortname), which designs specialized priority functions for the suboptimal policy constraint issue in offline RL, and employs unique decoupled resampling for training stability.
Through theoretical analysis, we show that the distinctive priority functions induce a provable improved behavior policy by modifying the distribution of the original behavior policy, and when constrained to this improved policy, a policy-constrained offline RL algorithm is likely to yield a better solution.
We provide two practical implementations to balance computation and performance: one estimates priorities based on a fitted value network (\shortalgo), and the other utilizes trajectory returns (\shortalgoo) for quick computation. 
As a highly compatible plug-and-play component, \shortname is evaluated with five prevalent offline RL algorithms: BC, TD3+BC, Onestep RL, CQL, and IQL. Our experiments confirm that both \shortalgo and \shortalgoo significantly improve performance across all baseline methods.
Moreover, \shortalgo can be effective in some challenging settings, \ie, without trajectory information.
Code and pretrained weights are available at \url{https://github.com/yueyang130/ODPR}.

\end{abstract}



\begin{figure*}[ht]
\vspace{-2mm}
 \centering
    \includegraphics[width=0.98\textwidth]{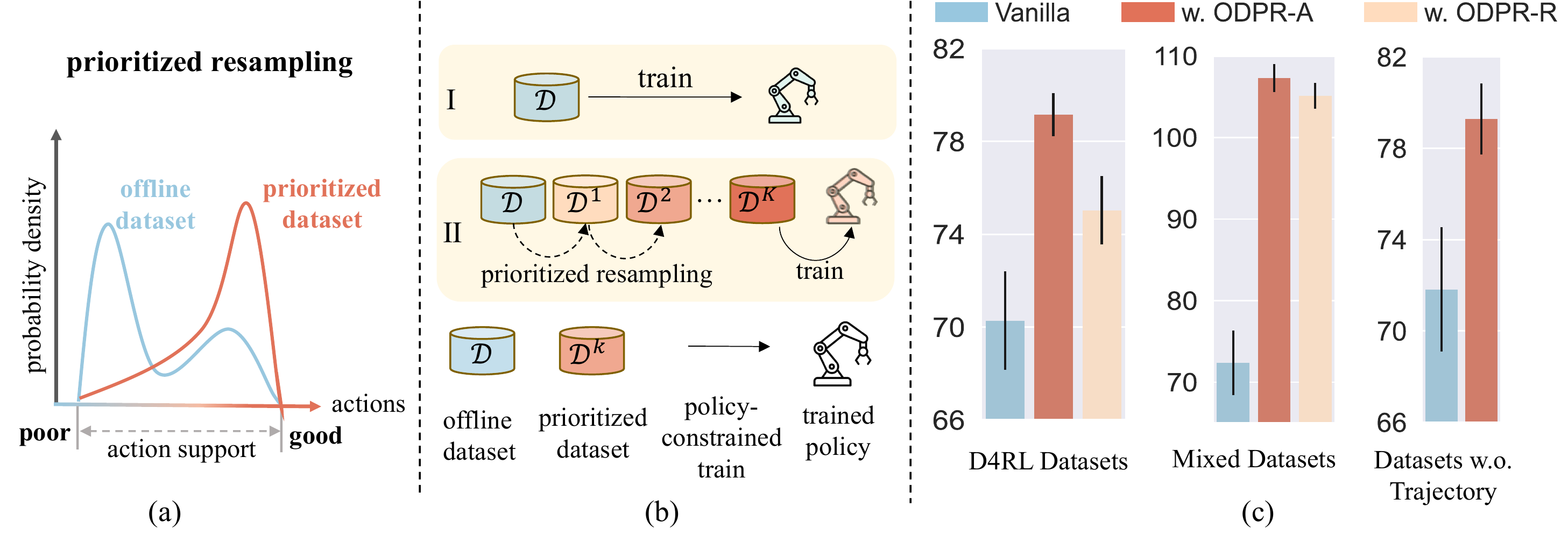}
    \caption{
    (a) \textbf{Prioritized resampling}. Given a state, possible actions are ranked by quality in x-axis. A behavior \textcolor[RGB]{152,199,223}{policy} (in blue) usually follows a multi-modal distribution. A \textcolor[RGB]{225,144,89}{prioritized policy} (in red) is acquired by prioritized resampling which assigns higher weights to better actions. 
    (b) \textbf{Offline RL (\MakeUppercase{\romannumeral 1}) \textit{v.s.} Offline RL with \shortname (\MakeUppercase{\romannumeral 2}).} Different from the vanilla offline RL, \shortname obtains a sequence of better behavior policy by iteratively resampling the current dataset. 
    Then policy-constrained offline RL algorithms are performed on the superior dataset $\mathcal{D}^K$.
    (c) \textbf{The average score of popular offline RL algorithms}. \shortname statistically boost the performance of these algorithms on standard D4RL benchmark and curated mixed datasets abundant with suboptimal data. Further, \shortalgo is effective within datasets without trajectory, where preceding trajectory-based resampling strategies~\cite{hong2023harnessing,chen2021decision} failed.
    }
    \label{fig:overall}
\vspace{-1mm}
\end{figure*}

\vspace{-5pt}
\section{Introduction}

Offline Reinforcement Learning (RL) aims to solve the problem of learning from previously collected data without real-time interactions with the environment~\cite{lange2012batch,prudencio2023survey}. 
However, standard off-policy RL algorithms tend to perform poorly in the offline setting due to the distributional shift problem~\cite{fujimoto2019off}. 
Specifically, to train a Q-value function based on the Bellman optimality equation, these methods frequently query the value of out-of-distribution (OOD) state-action pairs,  which leads to accumulative extrapolation error. 
Most existing algorithms tackle this issue by constraining the learning policy to stay close to the behavior policy that generates the dataset.
These constraints directly operate on the policy densities, such as KL divergence~\cite{jaques2019way, peng2019advantage, wu2019behavior}, Wasserstein distance~\cite{wu2019behavior}, maximum mean discrepancy (MMD)~\cite{kumar2019stabilizing}, and behavior cloning~\cite{td3+bc}.

However, such constraints might be too restrictive as the learned policy is forced to equally mimic bad and good actions of the behavior policy, 
especially in an offline scenario where data are generated by policies with different levels.
For instance, consider a dataset $\mathcal{D}$ with state space $\mathcal{S}$ and action space $\mathcal{A}=\{\rva_1,\rva_2,\rva_3\}$ collected with behavior policy $\beta$. At one specific state $\rvs^*$,  the policy $\beta$ assigns probability $0.2$ to action $\rva_1$, 0.8 to $\rva_2$ and zero to $\rva_3$. However, $\rva_1$ would lead to a much higher expected return than $\rva_2$. Minimizing the distribution distance of two policies can avoid $\rva_3$, but forces the learned policy to choose $\rva_2$ over $\rva_1$, resulting in much worse performance. 
Employing a policy constraint strategy is typically essential to avoid out-of-distribution actions.
However, this necessity often results in a compromise on performance, stemming from a suboptimal policy constraint.
Then the question arises: can we substitute the \textit{behavior policy, \ie, offline dataset} with an improved one, enabling the learned policy to \textit{avoid out-of-distribution actions and achieve superior policy constraint simultaneously}, thereby improving performance? Indeed, as illustrated in~\cref{fig:overall}(a), if we can accurately assess the quality of an action, we can then adjust its density to yield an improved behavior policy.

Based on the above motivation, we propose data prioritization strategies for offline RL, \textit{i.e.}, \emph{\name~(\shortname)}.
\shortname prioritizes data by the action quality, specifically, assigning priority weight proportional to normalized (\ie non-negative) advantage \textemdash~the additional reward that can be obtained from taking a specific action. 
Contrary to online prioritized resampling methods like PER~\cite{PER,kumar2020discor,sinha2022lfiw,brittain2019pser,wang2020striving,liu2021regret}, which mainly aim to accelerate value function fitting, our proposed priority is motivated by the desire to cultivate a superior behavior policy. 
We theoretically demonstrate that a prioritized behavior policy, with our proposed priority functions, yields a higher expected return than the original one.
In practice, we develop two implementations, \emph{Advantage-based \shortname (\shortalgo)} and \emph{Return-based \shortname (\shortalgoo)}. 
As depicted in~\cref{fig:overall}(b), \shortalgo fits a value network from the dataset and calculates advantages with one-step TD error for all transitions, and then iteratively refine current behavior policy based on the previous one.
Similarly, \shortalgoo, a more computation-friendly version, employs trajectory return as the priority when trajectory information is available.
Then, both \shortalgo and \shortalgoo run an offline RL algorithm with the prioritized behavior policy to learn a policy.
To summary, our contributions are two-fold:
\setlist[itemize]{leftmargin=15pt}
\begin{itemize}
\vspace{-1mm}
\item \textbf{Decoupled Prioritized Replay for Offline RL:} 
We introduce a unique class of priority functions specifically tailored for offline RL resampling. 
Furthermore, a distinctive feature of our methodology, different from existing resampling methods in both online~\cite{PER,kumar2020discor,sinha2022lfiw,brittain2019pser,wang2020striving,liu2021regret,oh2018self} and offline RL~\cite{hong2023harnessing,singh2022offline,chen2020bail,wang2018exponentially,liu2021curriculum}, is the incorporation of dual samplers: a uniform one for policy evaluation and a prioritized one for policy improvement and constraint. 
This approach, referred to as \textit{decoupled resampling}, allows obtaining a superior behavior policy while mitigating the off-policy degree in policy evaluation between current policy and the sampling distribution, which reduces potential instability~\cite{fu2019diagnosing, van2018deadly_triad, tsitsiklis1996deadly_tirad}. 
We underscore the significance of decoupled resampling for  offline RL resampling through experiments.

\item \textbf{Empirical Effectiveness:} 
Experimental results, as summarized
in~\cref{fig:overall}(c), reveals that our proposed prioritization strategies, serving as a plug-in orthogonal to algorithmic improvements, boost the performance of prevalent offline RL algorithms BC, CQL, IQL, TD3+BC, and OnestepRL~\cite{CQL,IQL,td3+bc,onestep} across diverse domains in D4RL~\cite{gym, fu2020d4rl}. 
When compared to existing popular resampling methods AW/RW~\cite{hong2023harnessing} and percentage sampling~\cite{chen2021decision} in offline RL, \shortalgo achieves superior performance owing to its unique \textit{fine-grained priorities}. This approach resamples transition by its advantage instead of assigning the same weights to all
transitions in a trajectory as is the case with popular resampling methods.
Furthermore, we demonstrate that, \shortalgo is also effective within datasets without complete trajectories, a scenario where preceding trajectory-based resampling strategies~\cite{hong2023harnessing, chen2021decision,yue2022boosting} in offline RL have been non-funcitonal.
\end{itemize}

\section{\name} 
\label{sec:theoretic_framework}

In this section, we develop \name, which prioritizes transitions in an offline dataset at training according to a class of priority functions. We start with an observation that performing prioritized sampling on a dataset generated with policy $\beta$ is equivalent to sampling from a new behavior $\beta^\prime$. Then, we theoretically justify that $\beta^\prime$ gives better performance than $\beta$ in terms of the cumulative return when proper priority functions are chosen. In the end, we propose two practical implementations of \shortname using transition advantage and return as the priority, respectively. 

\subsection{Preliminaries} 
\label{sec:preliminaries}
\textbf{Reinforcement Learning (RL).} RL addresses the problem of sequential decision-making, which is formulated with a Markov Decision Process $\langle\mathcal{S}, \mathcal{A}, {T}, r, \gamma\rangle$.
Here, $\mathcal{S}$ is a finite set of states; $\mathcal{A}$ is the action space. \YY{The transition dynamics function ${T}(\rvs, \rva, \rvs^\prime)$ defines the probability of moving from state $\rvs$ to the next state $\rvs^\prime$ after taking action $\rva$}. $r(\rvs,\rva)$ and $\gamma\in (0,1]$ are the reward function and the discount factor respectively. 
The policy is denoted by $\pi(\rva|\rvs)$ and an induced trajectory is denoted by $\tau$. 
The goal of RL is to learn a policy maximizing the expected  cumulative discounted reward:
\vspace{-1mm}
\begin{equation}
\label{eqn:rl_goal}
\setlength\abovedisplayskip{3pt}
\setlength\belowdisplayskip{3pt}
\small
     J(\pi) =   \mathbb{E}_{\tau \sim p_\pi(\tau)} \left[\sum_{t=0}^\infty \gamma^t r(\rvs_t, \rva_t) \right].
\end{equation}

\textbf{Offline RL as Constrained Optimization.} Offline RL considers a dataset $\mathcal{D}$ generated with behavior policy $\beta$. Since $\beta$ or $\mathcal{D}$ is fixed throughout training, maximizing $J(\pi)$ is equivalent to maximizing the improvement $J(\pi) - J(\beta)$. It can be measured by:
\begin{lemma}
\label{lmm:pdl}
\emph{(Performance Difference Lemma~\cite{kakade2002approximately}.)} For any policy $\pi$ and $\beta$,
\vspace{-1mm}
\begin{equation}
\label{eqn:PDL}
\setlength\abovedisplayskip{3pt}
\setlength\belowdisplayskip{3pt}
\small
J(\pi) - J(\beta) =  \int_\rvs d_\pi(\rvs) \int_\rva \pi(\rva | \rvs) A^\beta(\rvs, \rva) \ d\rva \ d\rvs,
\end{equation}
where $d_\pi(\rvs) =  \sum_{t=0}^\infty \gamma^t p(\rvs_t = \rvs | \pi)$,
represents the unnormalized discounted state marginal distribution induced by the policy $\pi$, and $ p(\rvs_t = \rvs | \pi)$ is the probability of the state $\rvs_t$ being $\rvs$. 
\end{lemma}
The proof is available in Kakade (2002)~\cite{kakade2002approximately}.

\subsection{Prioritized Behavior Policy}
Consider a dataset $\mathcal{D}$ generated with behavior policy  $\beta$. Let $ \omega(\rvs,\rva)$ denote a weight/priority function for the transition $(\rvs, \rva, \rvs^\prime)$ in $\mathcal{D}$. Then, we define a prioritized behavior policy $\beta^\prime$:
\begin{equation}
\label{eqn:reweighting}
    \beta^\prime(\rva|\rvs) = \frac{\omega(\rvs,\rva) \beta(\rva|\rvs)}{\int_\rva \omega(\rvs,\rva) \beta(\rva|\rvs) d\rva},
\end{equation}
where the denominator is to guarantee $\int_\rva \ \beta^\prime(\rva | \rvs) \ d\rva = 1$. 
As shown in ~\cref{fig:overall}(b), $\beta^\prime$ shares the same action support as $\beta$.
Suppose a dataset produced by prioritized sampling on $\mathcal{D}$ is $\mathcal{D}^\prime$. We have:
\begin{equation}
\label{eq:dqn}
    \mathbb{E}_{(\rvs,\rva) \sim \mathcal{D^\prime}}\left[ \mathcal{L}_{\theta}(\rvs,\rva) \right] = 
    \mathbb{E}_{\rvs \sim \mathcal{D}, \rva \sim \beta^\prime(\cdot|s) }\left[ \mathcal{L}_{\theta}(\rvs,\rva) \right],
\end{equation}
where $\mathcal{L}$ represents a generic loss function, and the constant is discarded as it does not affect the optimization. This equation shows that prioritizing the transitions in a dataset by resampling or reweighting (LHS) can mimic the behavior of another policy $\beta^\prime$ (RHS). 
Intuitively, priority functions, denoted as $\omega(\rvs, \rva)$, should be \textit{non-negative} and \textit{monotonically increase with respect to the quality of the action $\rva$}.
In the context of RL, advantage $A^\beta(\rvs, \rva)$ represents the extra reward that could be obtained by taking the action $\rva$ over the expected return. Therefore, advantage $A^\beta(\rvs, \rva)$, as an action quality indicator, provides a perfect tool to construct $\omega(\rvs,\rva)$. 
We can easily construct many functions that satisfy the above properties.
For instance, consider $\omega(A)$ is linear with respect to the advantage $A^\beta(\rvs, \rva)$ in~\cref{eqn:linear-weight}, expressed as follows:
\vspace{-1mm}
\begin{equation}
\label{eqn:linear-weight}
   \displaystyle \omega(A^\beta(\rvs, \rva)) = C(A^\beta(\rvs, \rva) - \min_{(\rvs, \rva)\in\mathcal{D}} A^\beta(\rvs, \rva)),
\end{equation}
where $C$ is a constant, set to make the sum over the dataset equal to 1. 
\YY{
Given a specific state $s$, if action $a_1$ is of higher quality than action $a_2$, evidenced by $A^\beta(\rvs, \rva_1) > A^\beta(\rvs, \rva_2)$, according to~\cref{eqn:linear-weight}, $\omega(\rvs, \rva_1)$ will be greater than $\omega(\rvs, \rva_2)$. Consequently, the state with better action is assigned a higher priority.}

\subsection{Prioritized Policy Improvement}
\label{sec:policy_improve}
We are ready to show that prioritized sampling can contribute to an improved \textit{learned policy}. We first show the prioritized version $\beta^\prime$ is superior to $\beta$.

\paragraph{Behavior Policy Improvement.} The below theorem underscores that prioritization can improve the original behavior policy $\beta$ if it is a stochastic policy or a mixture of policies, either of which could result in actions of different Q-values. The detailed proof is deferred to Appendix~\ref{appendix:beta_improve}. 
\begin{theorem}\label{thm:beta_improve}
Let $\omega(A)$ be any priority function with non-negative and monotonic increasing properties. Then, we have $J(\beta^\prime) - J(\beta) \geq 0$. If there exists a state $\rvs$, under which not all actions in action support $\{ \rva|\beta(\rva|\rvs) > 0, \rva \in \mathcal{A} \}$ have the same $Q$-value, the inequation strictly holds. 
\end{theorem}

In an extreme case, if the policy constraint is exceptionally strong, causing the learned policy to exhibit performance very similar to the behavior policy, $\pi^{\prime*}$ obviously surpasses $\pi^*$ because $\beta^\prime$ is greater than $\beta$.
 In this way, we have $J(\pi^{\prime*}) \geq J(\pi^*) $, which demonstrates that $\pi^{\prime *}$ is a better solution. Although with an assumption about state distribution, it still offers valuable insights that the constraint induced by prioritized behavior policy has the potential to improve the performance of the learned policy. 
The rationale behind this is straightforward: when starting from a better behavior policy (\cref{thm:beta_improve}), the learned policy is more likely, though not guaranteed, to achieve a higher final performance.  


\subsection{Practical Algorithms}
\paragraph{\algo (\shortalgo).} We approximate priorities by fitting a value function $V^{\beta}_\psi(\rvs)$ for the behavior policy $\beta$ by TD-learning:
\vspace{-1mm}
\begin{equation}
\label{eqn:estimate_value}
    \min_\psi \quad \mathbb{E}_{(\rvs,\rva,\rvs^\prime,r) \sim \mathcal{D}} \left[ \left(r+\gamma  V_\psi(\rvs^\prime) - V_\psi(\rvs)\right)^2 \right].
\end{equation}
\vspace{-1mm}
The advantage for $i$-th transition $(\rvs_i,\rva_i,\rvs_i^\prime,r_i)$ in the dataset is then given by a one-step TD error: 
\begin{equation}
\label{eqn:p_adv}
A(\rvs_i,\rva_i) = r_i + V_\psi(\rvs_i^\prime) - V_\psi(\rvs_i),
\end{equation}
which is similar to the form of priority in online PER, such as the absolute TD error, but differs in whether the absolute value is taken.
\YY{Then the prioritized  is obtained by \cref{eqn:linear-weight}.}
This implementation is referred to as \textit{\shortalgo} in the following. 
The term ``decoupled" will be elucidated subsequently within this section.

\paragraph{\algoo (\shortalgoo)} The limitation of \shortalgo is also clear, \textit{i.e.}, fitting the value network incurs extra  computational cost (refer to~\cref{sec:time}). Therefore, we propose another variant that uses trajectory return as an alternative transition quality indicator. For the $i$-th transition, we find the complete trajectory that contains it, and calculate the return for the whole trajectory $G_i = \sum_{k=0}^{T_i} r_k$. $T_i$ is the length of the trajectory. Then the priority is obtained by
\vspace{-1mm}
{
\begin{equation}
\label{eqn:p_return}
    \omega_i = C(\frac{G_i - G_{\min}}{G_{\max} - G_{\min}} + p_\text{base}),
\end{equation}}
where $G_{\min} = \min_i G_i$ and $G_{\max} = \max_i G_i$. $p_\text{base}$ is a small positive constant that prevents zero weight. $C$ is a constant, set to make the sum equal to 1. 
We term this variant as \textit{\shortalgoo}. \shortalgoo can only work with datasets where the trajectory information is available.

\paragraph{Decoupled prioritized resampling.} After obtaining priority weights, \shortname can be implemented by both resampling and reweighting. The sampling probability or weight is proportional to its priority. We opt for resampling in the main text and also provide the results of reweighting in the Appendix~\ref{appendix:sample-and-weight}.
An offline RL algorithm can be decomposed into three components: policy evaluation, policy improvement, and policy constraint. 
This leads us to the question: where should resampling be applied?
In alignment with our stated motivation, we employ prioritized data for both policy constraint and policy improvement terms to mimic being constrained to a better behavior policy.
However, we found it is crucial to conduct policy evaluation under uniform sampling data. It helps mitigate instability of value estimation that arises from an excessive off-policy degree~\cite{fu2019diagnosing,sutton2018reinforcement,van2018deadly_triad,tsitsiklis1996deadly_tirad}.
Such a prioritization method is termed as \textit{decoupled prioritized resampling}.
For decoupled prioritized resampling, two samplers are employed, one for uniform sampling and one for prioritized sampling.
A more in-depth discussion about why decoupled resampling is crucial can be found in Section \ref{sec:ablation}. The complete algorithm is given in \cref{alg:oper}.

\vspace{-1.2mm}
\begin{algorithm}[h]
  \caption{\name}
  \label{alg:oper}
\begin{algorithmic}[1]
   \STATE \textbf{Require:} Dataset $\mathcal{D} = \{(\rvs,\rva,\rvs',r)_i\}_{i=1}^N$,
   a policy-constrained algorithm $\mathcal{I}$ 
    \STATE \textcolor{red}{\textbf{Stage1:}} Calculate ${\omega}_{i}$ according to ~\cref{eqn:p_adv} or ~\cref{eqn:p_return} ( with trajectory information).
    \STATE \textcolor{blue}{\textbf{Stage2 (Decoupled Resampling):}} Train algorithm $\mathcal{I}$ on dataset  $\mathcal{D}$.  Sample transition $i$ with the priority $\omega_{i}$ for policy constraint and improvement. Uniform sample for policy evaluation.\\
\end{algorithmic}
\end{algorithm}
\vspace{-3.2mm}

\subsection{Improving \shortalgo by Iterative Prioritization}
\label{sec:iteration}

In ~\cref{sec:policy_improve}, we demonstrate a likelihood that enhancing $\beta(\rva|\rvs)$ to $\beta^\prime(\rva | \rvs)$ leads to an improvement in the learned policy through offline RL algorithms. Then, a natural question arises: can we further boost the learned policy by improving $\beta^\prime (\rva | \rvs)$? The answer is yes. Suppose we have a sequence of behavior policies $\beta^{(0)}, \beta^{(1)}, \dots, \beta^{(K)}$ satisfying $\beta^{(k)} (\rva|\rvs) \propto \omega(A^{(k-1)}(\rva, \rvs)) \beta^{(k-1)}(\rva | \rvs) $, where $A^{(k-1)}(\rva, \rvs)$ represents the advantage for policy $\beta^{(k-1)}(\rva | \rvs)$. We can easily justify that the behavior policies are monotonically improving by \cref{thm:beta_improve}: 
\begin{equation*}
J(\beta^{(0)}) \leq J(\beta^{(1)}) \leq J(\beta^{(2)}) \leq \cdots \leq J(\beta^{(K)}).
\end{equation*}
It is reasonable to anticipate, though not guarantee, the following relationship: 
\YY{$\textstyle
J(\pi^{(0)*}) \leq J(\pi^{(1)*}) \leq J(\pi^{(2)*}) \leq \cdots \leq J(\pi^{(K)*})
$}, where $\pi^{(k)*}$ is the optimal solution of policy-constraint policy search. 
We build such a sequence of behaviors from a fixed policy $\beta^{(0)}=\beta$ and its dataset $\mathcal{D}$, which relies on the recursion $\beta^{(k)}(\rva | \rvs) \propto \prod_{j=0}^{k-1} \omega(A^{(j)}(\rva, \rvs)) \cdot \beta^{(0)}(\rva|\rvs)$.

It means that a dataset $\mathcal{D}^{(k)}$ for behavior $\beta^{(k)}$ can be acquired by resampling the dataset $\mathcal{D}$ with weight $\prod_{j=0}^{k-1} \omega(A^{(j)}(\rva, \rvs))$ (normalize the sum to 1). Then, the advantage $A^{(k)}$ can be estimated on $\mathcal{D}^{(k)}$ following \cref{eqn:estimate_value}-\cref{eqn:p_adv}. 
After all iterations, we scale the standard deviation of priorities to a hyperparameter $\sigma$ to adjust the strength of data prioritization.
The full algorithm for this iterative \shortalgo is presented in \cref{alg:adv-reweight}. 
In the experiments, \shortalgo mainly refers to this improved version. It is notable that priorities that are acquired in the first stage can be saved and made public, and then offline RL algorithms could directly use the existing priorities without extra cost.

\begin{algorithm}[thb]
   \caption{\algo}
   \label{alg:adv-reweight}
\begin{algorithmic}[1]
   \STATE \textbf{Require:} Dataset $\mathcal{D} = \{(\rvs,\rva,\rvs',r)_i\}_{i=1}^N$, \ie, behavior policy $\beta^{(0)}$, the number of iterations $K$, standard deviation $\sigma$, and a policy-constrained algorithm $\mathcal{I}$\\
    \STATE \textcolor{red}{\textbf{Stage1:}}  Initialize transition priorities $\omega_{i=1}^N = 1$  \\

    \FOR{step $k$ in \{1, \dots, $K$\}}
        \STATE Evaluate advantage $A^{(k-1)}$ of behavior policy $\beta^{(k-1)}$ by sampling transition $i$ with the probability $\omega_{i}$. \\
        \STATE Calculate $\omega(A^{(k-1)}(\rvs_i, \rva_i))$ by \cref{eqn:linear-weight}. \\
        \STATE \mbox{$\omega_{i} := \omega_{i} * \omega(A^{(k-1)}(\rvs_i, \rva_i))$} \\
    \ENDFOR
    \STATE  Scale the standard deviation of $\omega_{i}$ to $\sigma$ \\
    \STATE \textit{\textcolor{gray}{save $\omega_i$ for utilization in other algorithms and runs}}
    \STATE \textcolor{blue}{\textbf{Stage2(Decoupled Resampling):}}   Train algorithm $\mathcal{I}$ on dataset  $\mathcal{D}$.  Sample transition $i$ with the priority $\omega_{i}$ for policy constraint and improvement. Uniform sample for policy evaluation.\\
\end{algorithmic}
\end{algorithm}

\begin{figure}[h]
\vspace{-2mm}
 \centering
    \includegraphics[width=0.4\textwidth]{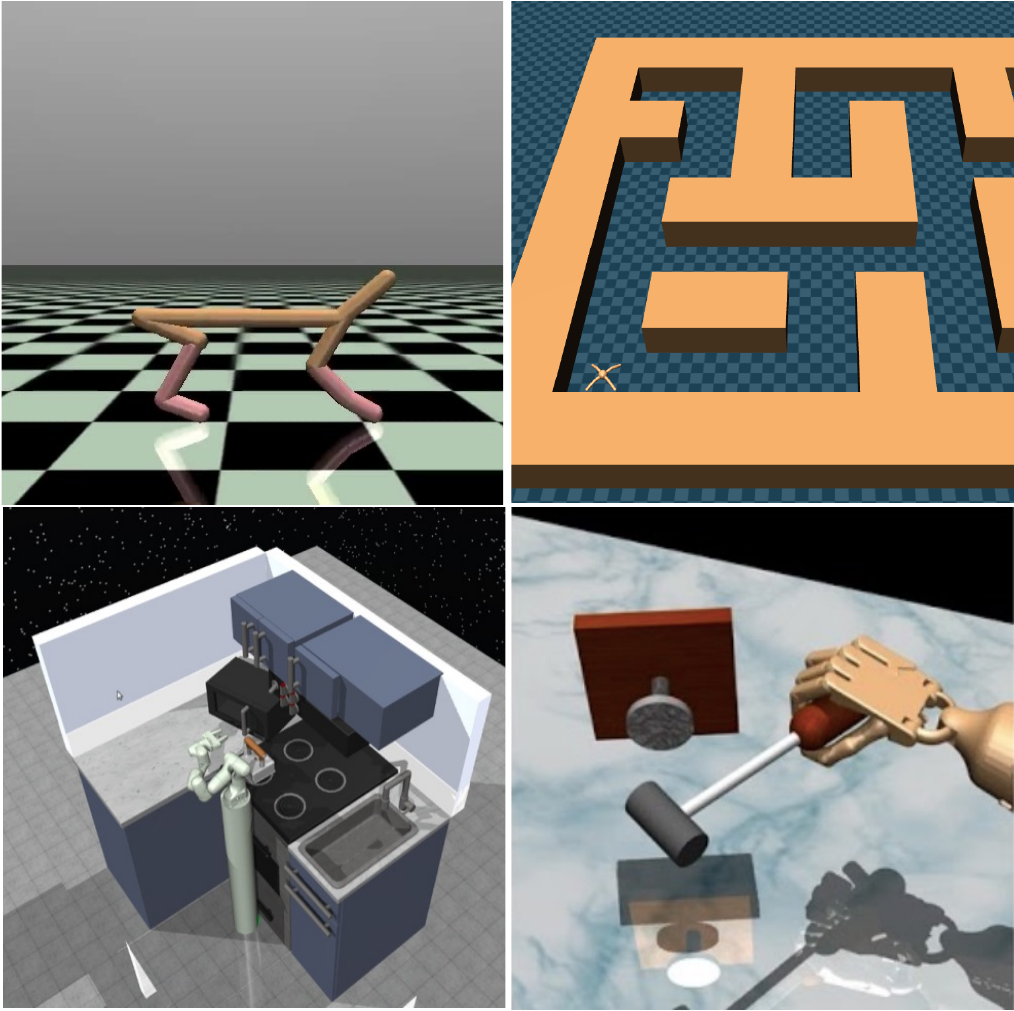}
    \caption{
    A subset of tasks pertinent to real-world applications from the D4RL benchmark~\cite{fu2020d4rl}, including Mujoco locmotion tasks with bipeds or quadrupeds, Maze navigation tasks with an 8-DoF Ant quadruped robot, Kitchen tasks with a 9-DoF Franka robot, and Adroit tasks with a 24-DoF Hand robot\protect\footnotemark.}
    \label{fig:envs}
\vspace{-1mm}
\end{figure}
\footnotetext{Figures are adapted from the D4RL paper~\cite{fu2020d4rl}.}

\begin{figure*}[th]
\vspace{-3mm}
\captionsetup[subfloat]{font=scriptsize}
\includegraphics[width=0.98\textwidth]{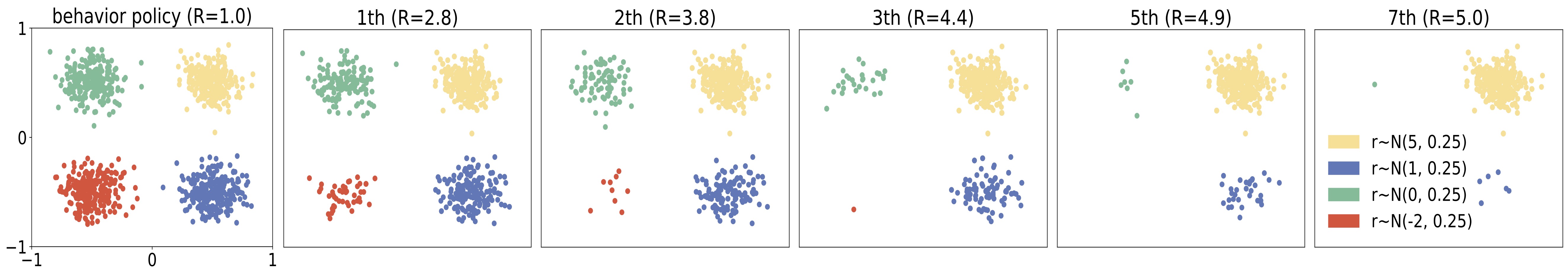}
\caption{
Visualization of the effect of \shortname on prioritized behavior policies in the bandit setting. As the iterations progress, the suboptimal \textcolor[RGB]{238,85,61}{red},
\textcolor[RGB]{102,188,152}{green}, and \textcolor[RGB]{86,121,186}{blue} samples noticeably decrease, leading to the prioritized dataset progressively converging towards the optimal action mode, represented by the \textcolor[RGB]{255,224,148}{yellow} color. The value enclosed in parentheses denotes the average reward.
\label{fig:bandit1}
  }
\end{figure*}

\section{Experiments}
\label{sec:experiments}

Our experiments seek to investigate \shortname from the following perspectives:
\begin{enumerate}
\item \textbf{Visualization.} We initiate our investigation with a toy bandit experiment. It vividly illustrates the impact of prioritized sampling on the behavior policies.

\item \textbf{Effectiveness.} We apply \shortname to state-of-the-art offline RL algorithms to demonstrate its effectiveness across diverse domains in the D4RL benchmark, as well as curated mixed datasets with suboptimal behaviors. These tasks are detailed in~\cref{fig:envs}.

\item \textbf{Ablation Studies.} We delve deeper into the role of decoupled sampling in the success of \shortname. We also examine whether iterative prioritization effectively enhances the behavior policy and the learned policy.

\item \textbf{Comparison to Existing Methods.} 
We differentiate the priority function of \shortname from those used in existing resampling methods in online RL~\cite{PER}.
We also compare \shortname to existing resampling methods in offline RL, such as AW/RW~\cite{hong2023harnessing} and percentage resampling~\cite{chen2021decision}. Furthermore, we investigate whether \shortalgo can enable more challenging scenarios without trajectory information where previous methods have failed to function.
\end{enumerate}
Lastly, we delve into the additional computational overhead introduced by the priority estimation within \shortalgo.

\begin{table*}[thpb]
\centering
\small
\caption{Normalized scores on MuJoCo locomotion v2 tasks. We report the average and the standard deviation (SD) of the total score over 15 seeds. 
The results that have an advantage over the baselines (denoted as vanilla) are printed in \textbf{bold} type. ``m", ``mr", and ``me" are respectively the abbreviations for ``medium", ``medium-replay", and ``medium-expert". ``V", ``A", and ``R" denotes ``vanilla", ``\shortalgo", and ``\shortalgoo".
Standard deviation of individual games can be found at~\cref{sec:result_with_std}. 
}
\label{tab:mujoco}
\begin{tabular}{ccccccccccccc}
\toprule
\multirow{2}{*}{Dataset} & \multicolumn{3}{c}{TD3+BC~\cite{td3+bc}} & \multicolumn{3}{c}{CQL~\cite{CQL}} & \multicolumn{3}{c}{IQL~\cite{IQL}} & \multicolumn{3}{c}{OnestepRL~\cite{onestep}} \\ 
\cmidrule(lr){2-4}\cmidrule(lr){5-7}\cmidrule(lr){8-10}\cmidrule(lr){11-13}
& V     & A              & R & V     & A              & R              & V     & A              & R              & V     & A              & R              \\
\cmidrule(lr){1-1}\cmidrule(lr){2-13}
halfcheetah-m  & 48.3  & \textbf{50.0}  & \textbf{48.6}           & 48.2  & 48.3           & 48.1           & 47.6  & 47.5           & 47.6           & 48.4  & \textbf{48.6}           & 48.4           \\
hopper-m       & 57.3  & \textbf{74.1}  & \textbf{59.1}           & 72.1  & \textbf{72.7}           & \textbf{74.9}           & 64.1  & \textbf{66.0}  & \textbf{66.4}  & 57.2  & \textbf{64.8}  & \textbf{58.2}           \\
walker2d-m     & 84.9  & 84.9           & 84.2           & 82.1  & \textbf{83.9}           & 80.7           & 80.0  & \textbf{83.9}  & 78.3           & 77.9  & \textbf{85.1}  & \textbf{80.9}           \\
halfcheetah-mr & 44.5  & \textbf{45.9}          & 44.6           & 45.2  & 45.4           & \textbf{46.1}  & 43.4  & 43.0           & \textbf{44.0}  & 37.5  & \textbf{42.9}  & \textbf{39.7 }          \\
hopper-mr      & 58.0  & \textbf{88.7}  & \textbf{77.4}  & 96.1  & 94.2           & 92.3           & 88.4  & \textbf{95.3}  & \textbf{99.9}  & 90.1  & 82.6           & 90.6           \\
walker2d-mr    & 72.9  & \textbf{88.2}  & \textbf{82.7}           & 82.3  & \textbf{85.9}  & 81.7           & 69.1  & \textbf{82.7}  & \textbf{79.1}  & 58.2  & \textbf{72.4}  & \textbf{63.7}           \\
halfcheetah-me & 92.4  & 83.3           & 93.9           & 62.1  & \textbf{70.7}  & \textbf{84.3}  & 82.9  & \textbf{92.7}  & \textbf{93.5}  & 94.1  & 94.2  & 93.9  \\
hopper-me      & 99.2  & \textbf{107.3} & \textbf{106.7} & 82.9  & \textbf{105.1} & \textbf{97.2}  & 97.2  & \textbf{105.1} & \textbf{107.2} & 80.5  & \textbf{99.4}  & \textbf{98.8}  \\
walker2d-me    & 110.2 & \textbf{111.7}          & 110.1          & 110.0 & 107.9          & 109.6          & 109.4 & \textbf{111.6} & \textbf{110.7}          & 111.1 & \textbf{112.5} & 111.4          \\
\midrule
\rowcolor[HTML]{EFEFEF}
total          & 667.7 & \textbf{734.1} & \textbf{707.3} & 681.0 & \textbf{714.1} & \textbf{714.9} & 682.1 & \textbf{727.8} & \textbf{726.7} & 655.0 & \textbf{702.5} & \textbf{685.6} \\
\rowcolor[HTML]{EFEFEF}
SD(total)      & 18.4  & 10.4           & 7.9            & 15.3  & 6.2            & 14.9           & 22.3  & 11.2           & 8.9            & 21.7  & 6.2            & 16.7           \\
\bottomrule
\end{tabular}
\vspace{-2mm}
\end{table*} 

\subsection{Visualization of Toy Bandit Problem}
We consider a bandit task, where the action space is 2D continuous, $\textstyle \mathcal{A}\!=\![-1,1]^2$~\cite{wang2023diffusion} and as a bandit has no states, the state space $\mathcal{S}\!=\!\emptyset$. The offline dataset is as the first figure in ~\cref{fig:bandit1} shows. Please see Appendix~\ref{appendix:bandit_setup} for details.
The goal of the bandit task is to learn the action mode with the highest expected reward from the offline dataset.

We first show that prioritized datasets are improved over the original one in \cref{fig:bandit1}.
The red samples with the lowest reward are substantially reduced in the first two iterations.
After iterating five times, the suboptimal blue and green samples also significantly diminish. The average return of the prioritized dataset is increased to 4.9, very close to the value of optimal actions. In the 7th iteration, suboptimal actions almost disappear. 
Since the reward is exactly the return in bandit, \shortalgoo is the 1st prioritized behavior policy of \shortalgo, which raises the average return from 1.0 to 2.8.

\begin{figure}[ht]
\vspace{-1mm}
\centering
\includegraphics[width=0.48\textwidth]{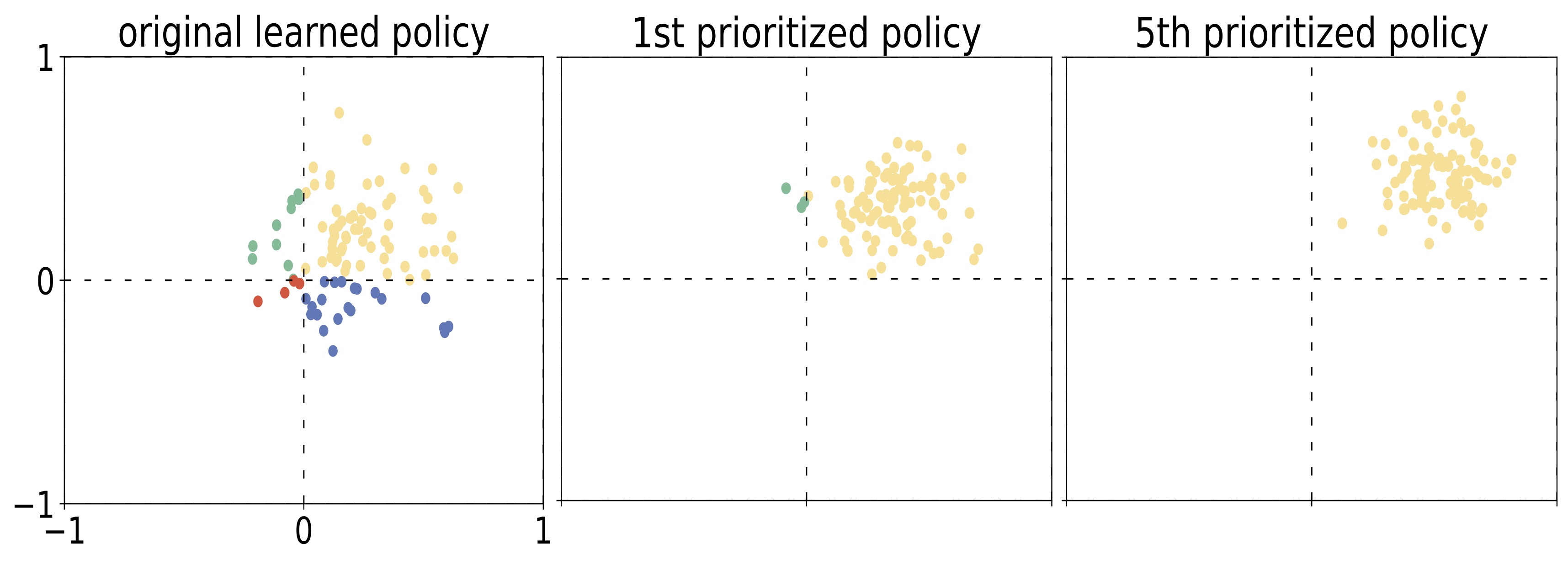}
\vspace{-3mm}
\caption{The left figure represents TD3+BC learning on the original dataset, which failed to find the optimal action. The middle and right figures depict the learned policies obtained by running TD3+BC on the first and fifth prioritized datasets, respectively. These policies demonstrate convergence towards nearly optimal and optimal modes, respectively.}
\label{fig:bandit2}
\end{figure}

Next, we show how offline RL algorithms can be improved by \shortalgo. As \cref{fig:bandit2} shows, when trained on the original dataset, TD3+BC failed to produce the optimal policy since it is negatively affected by suboptimal actions and converges to $(0.2,0.2)$, the mean of four modes (policy constraint) but biased towards the best action (policy improvement). 
However, if combined with \shortalgo~(iteration K=5), it successfully finds the optimal mode.

\subsection{Performance on standard D4RL Benchmark}

In this section, we present experimental results conducted on the D4RL benchmark to empirically demonstrate the capability of \shortname to enhance the performance of widely-used offline RL algorithms across a variety of domains.

\textbf{Experiment setups.} As discussed in~\cref{sec:policy_improve}, behavior cloning (BC), as a special case of offline RL, can be improved by \shortname. In addition, \shortname is a general plug-and-play training scheme that improves a variety of state-of-the-art (SOTA) offline RL algorithms. In our work, we choose four widely adopted algorithms as case studies, CQL, OnestepRL, IQL, and TD3+BC. 

\shortname's priority weights are generated in the first stage and then can be reused among baselines and seeds, saving computation. However, to assess the variance of \shortname and verify the generalization ability of \shortname to different algorithms, we organize experiments by sharing priority weights among baselines but not seeds. Specifically, we take seed=1 to compute \shortalgo weights, and then apply these weights and seed=1 to run TD3+BC, IQL, \etc We subsequently repeat this process with the next random seed.
Detailed experiment settings can be found in Appendix~\ref{appendix:oper_setup} and~\ref{appendix:offrl_setup}.



\begin{table}[ht]
\centering
\small
\caption{Averaged normalized scores of Behavior Cloning (BC) on MuJoCo locomotion v2 tasks over 15 seeds. }
\label{tab:mujoco_bc}
\begin{tabular}{cccc}
\toprule
\multirow{2}{*}{Dataset} &  \multicolumn{3}{c}{BC}     \\ 
\cmidrule(lr){2-4}
& V     & A              & R              \\
\midrule
halfcheetah-m  & 42.7  & \textbf{46.5}   & 42.6           \\
hopper-m       & 48.3  & \textbf{57.4}    & \textbf{52.2}     \\
walker2d-m     & 73.3  & \textbf{83.8}  & 70.1           \\
halfcheetah-mr & 33.4  & \textbf{41.6}   & \textbf{39.1} \\
hopper-mr      & 31.1  & \textbf{56.1}  & 30.3           \\
walker2d-mr    & 26.5  & \textbf{81.2}  & \textbf{48.2}   \\
halfcheetah-me & 62.8  & \textbf{95.4}  & \textbf{81.1}   \\
hopper-me      & 52.3  & \textbf{110.7} & \textbf{71.2}   \\
walker2d-me    & 106.4 & \textbf{110.9}  & \textbf{107.4}    \\
\rowcolor[HTML]{EFEFEF}
total          & 476.8 & \textbf{683.6} & \textbf{542.2} \\
\rowcolor[HTML]{EFEFEF}
SD(total)      & 17.7  & 8.1            & 18.2      \\
\bottomrule
\end{tabular}
\end{table}

\textbf{Mujoco locomotion.} 
\cref{tab:mujoco_bc} reveals that \shortname induces a better offline dataset, from which behavior cloning produces a behavior policy with higher performance. Further, \cref{tab:mujoco} shows that even though the state-of-the-art algorithms have achieved a strong performance, \shortalgo and \shortalgoo can further improve the performance of \textbf{\textit{all}} algorithms by large margins. Specifically, with \shortalgo, TD3+BC achieves a total score of 734.1 from 667.7. In addition, IQL, when combined with \shortalgo and \shortalgoo, also reaches 727.8 and 726.7 points, respectively.
We observe that \shortalgo generally performs better than \shortalgoo. This is potentially because \shortalgo is improved by iterative prioritization while \shortalgoo simply utilizes trajectory returns.  Interestingly, \shortname occasionally attains a smaller standard deviation than the vanilla, mainly due to its ability to achieve higher and more stable scores in
some difficult environments. Another notable observation is that although TD3+BC performs worse than IQL and CQL in their vanilla implementations, TD3+BC eventually obtains the highest performance boost with \shortalgo and achieves the best performance with a score of 734.1. 
The reason might be that TD3+BC directly constrains the policy with a BC term, which is easier to be affected by negative samples.
 
\begin{figure}[th]
    \centering
    \includegraphics[width=0.24\textwidth]{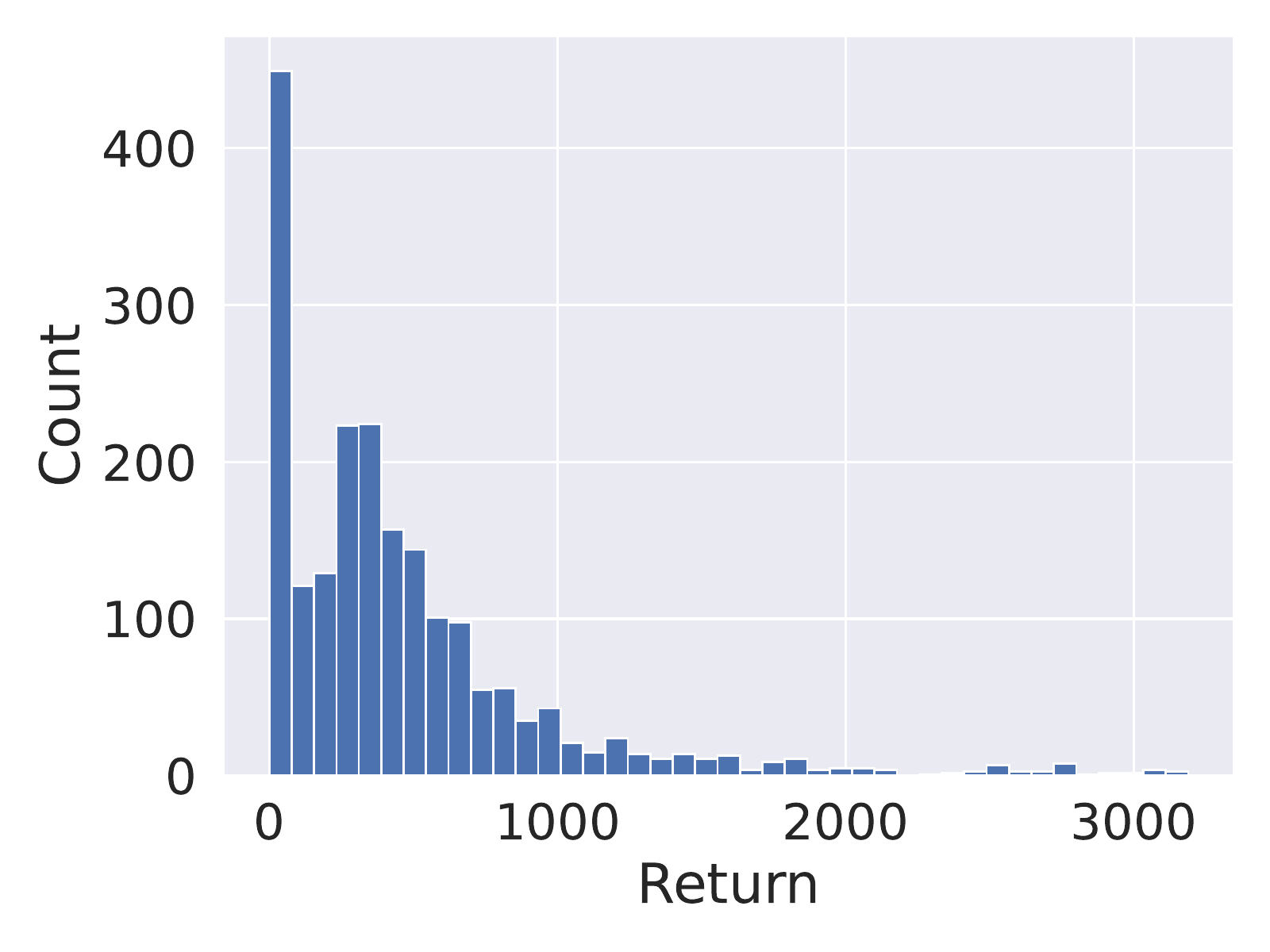}
    \includegraphics[width=0.24\textwidth]{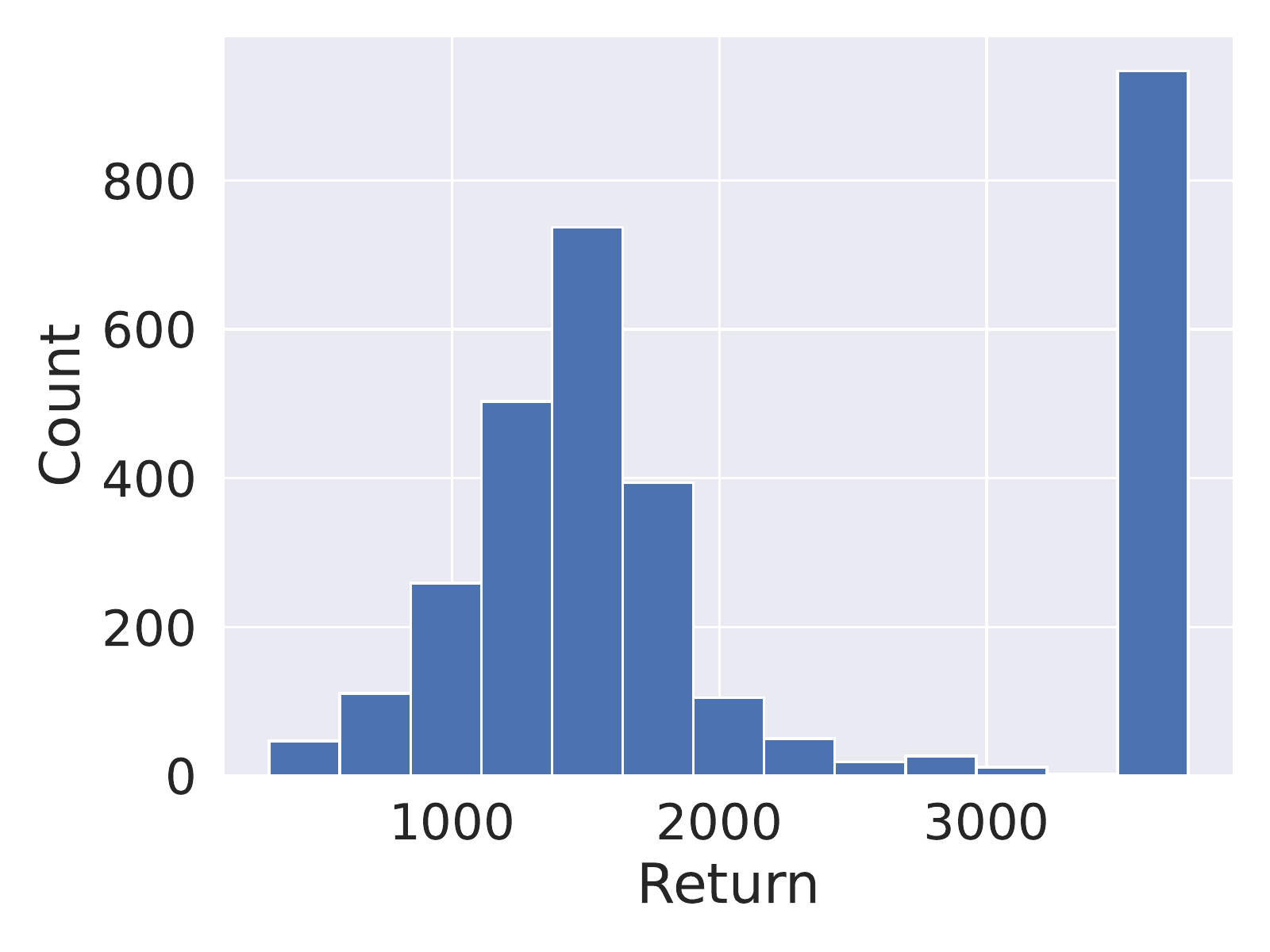}
    \vspace{-18pt}
    \caption{Trajectory Return Distributions of hopper-medium-replay (left) and hopper-medium-expert (right). Medium-replay datasets usually have a long-tailed distribution, and medium-expert often display two peaks. Both are composed of policies with varying quality.}
    \label{fig:part_dist_vis}
\end{figure}

\textbf{Discussions on data prioritizing.} In particular, we observe that on the locomotion tasks, the performance boost of \shortalgo and \shortalgoo mainly comes from the ``medium-replay" and ``medium-expert" level environments. To better understand this phenomenon, we visualize trajectory return distributions of hopper on these two levels in~\cref{fig:part_dist_vis}. The visualizations suggest that these tasks have a highly diverse data distribution.
This is consistent with our intuition that the more diverse the data quality is, the more potential for the data to be improved through data prioritizing by quality.

\textbf{Curated mixed datasets.}
To validate our core proposition, we also constructed a mixed dataset by mixing random and expert datasets and examine how \shortname perform when the behavior policy is a mixture of policies.~\cref{tab:mix} reflects that the vanilla TD3+BC underperforms within these mixed datasets. In contrast, \shortalgo and~\shortalgoo manages to target beneficial actions, resulting in a performance that stands on par with a pure expert. This underscores the significant potential of applying \shortname in real-world applications, where data are often collected from users with diverse skill levels.
\begin{table}[htb!]
\centering
\small
\caption{
Mixed datasets comprising random and expert data with suboptimal behaviors (10 seeds).}
\label{tab:mix}
\begin{tabular}{cccc}
\toprule
\multirow{2}{*}{Mixed Dataset} &  \multicolumn{3}{c}{TD3+BC}     \\ 
\cmidrule(lr){2-4}
& Vanilla  & \shortalgo    & \shortalgoo         \\
\midrule
halfcheetah  & 89.3  & \textbf{102.1} & \textbf{96.6 }               \\
hopper      & 102.2  & \textbf{109.4} & \textbf{108.7 }               \\
walker2d     & 25.6  & \textbf{110.4} & \textbf{110.0}     \\
\rowcolor[HTML]{EFEFEF}
total          & 217.1±11.8 & \textbf{321.9±5.2} & \textbf{315.3±4.7} \\
\bottomrule
\end{tabular}
\end{table}

\textbf{Antmaze, Kitchen and Adroit.}
In addition to the locomotion tasks, we evaluate our methods in more challenging environments.
Given that IQL achieves the absolute SOTA performance in these domains and other algorithms, e.g., CQL, do not give an ideal performance in these domains, we pick IQL as a case study. We present the results in~\cref{tab:iql}. Similarly, we observe that both \shortalgo and \shortalgoo can further improve the performance of IQL on all three domains. In the most challenging Antmaze environments, \shortalgo and \shortalgoo successfully improve the most difficult medium and large environments. For Kitchen and Adroit tasks, we have observed a similar trend of improvement. 
Considering the inherent variability of these tasks with larger variances, we conducted t-tests at a significance level of 0.05. The results demonstrate that the performance improvements achieved by \shortname are statistically significant.

\begin{table}[htbp]
  \centering
  \small
  \setlength{\tabcolsep}{2pt} 

\caption{Averaged normalized scores on Antmaze, Kitchen, and Adroit tasks over 15 seeds. The results for AW/RW~\cite{hong2023harnessing} were taken directly from the official source\protect\footnotemark. PS represents percentage sampling~\cite{chen2021decision}.}
  \label{tab:iql}
  \begin{tabular}{c l cccccc }
    \toprule
    &           &  IQL & AW & RW & PS & \shortalgo & \shortalgoo   \\
    \midrule
    \multirow{6}{*}{\rotatebox[origin=c]{90}{antmaze}}
    &umaze          & 88.5 & \textbf{90.7} & 66.6 & 0 & 85.5$\pm$4.4 & 87.8$\pm$3.0 \\
    &u-diverse  & 63.1 & \textbf{75.3} & 48.3 & 0 & \textbf{70.8$\pm$7.8} & 66.0$\pm$7.8 \\
    &med-play    & 70.5 & 61.3 & 8.3  & 0 & \textbf{76.1$\pm$5.1} & \textbf{72.0$\pm$5.4} \\
    &med-diverse & 58.5 & 22.0 & 10.0 & 0 & \textbf{71.8$\pm$6.6} & \textbf{74.2$\pm$9.4} \\
    &large-play     & 44.1 & 23.3 & 10.0 & 0 & 40.0$\pm$5.3 & \textbf{49.6$\pm$4.0} \\
    &large-diverse  & 42.0 & 9.3  & 11.6 & 0 & \textbf{48.0$\pm$4.0} & 43.0$\pm$4.9 \\
    \rowcolor[HTML]{EFEFEF}
    &antmaze total  & 366.7 & 281.9 & 154.8 & 0 & \textbf{392.2$\pm$19.1} & \textbf{392.6$\pm$20.4} \\
    \midrule
    \multirow{3}{*}{\rotatebox[origin=c]{90}{kitchen}}
    &complete-v0    & 65.9 & 26.3 & 24.2 & 5.4 & 64.2$\pm$6.1 & 62.7$\pm$7.8 \\
    &partial-v0     & 51.4 & \textbf{73.1} & \textbf{73.8} & \textbf{72.5} & 66.5$\pm$13.2 & \textbf{69.5$\pm$6.9} \\
    &mixed-v0       & 50.3 & 47.8 & 50.8 & 49.2 & 52.1$\pm$6.7 & 49.9$\pm$3.3 \\
    \rowcolor[HTML]{EFEFEF} 
    &kitchen total  & 167.6& 147.2 & 148.8 & 127.1 & \textbf{182.8$\pm$15.9} & \textbf{182.1$\pm$14.4} \\ 
    \midrule
    \multirow{2}{*}{\rotatebox[origin=c]{90}{pen}}
    &human-v0           & 73.1 & 81.1 & 79.5 & 28.3 & 72.9$\pm$15.5 & \textbf{83.0$\pm$17.2} \\
    &cloned-v0          & 42.1 & \textbf{92.3} & \textbf{96.6} & 54.6 & 61.2$\pm$13.6 & 66.6$\pm$21.4 \\
\midrule
\rowcolor[HTML]{EFEFEF} 
& all total  & 649.5  & 612.5 & 479.7 & 210.0 & \textbf{709.1} &  \textbf{724.3}  \\
    \bottomrule
  \end{tabular}
\end{table}

\footnotetext{
Note that there are inconsistencies between the AW/RW results from the referenced paper~\cite{hong2023harnessing} and its corresponding GitHub repository~\url{https://github.com/Improbable-AI/harness-offline-rl}. The results reported in the paper appear to be unreliable as they present the original IQL results as significantly lower than those reported in the IQL paper~\cite{IQL}. We have chosen to use the results from the GitHub repository, as they align more closely with the results in the IQL paper.
We rerun the code with 15 seeds for missing values.
}


\subsection{Ablation Studies}
\label{sec:ablation}

\textbf{Effect of the number of iterations K.} 
\YY{In~\cref{sec:iteration}, we introduce the number of iteration $K$ to iteratively derive a final prioritized behavior policy $\beta^{(K)}$.}
In \cref{tab:iterations}, as the iteration progresses, the overall performance of BC and TD3+BC combined with \shortalgoo on the locomotion tasks continues to increase.
For BC, performance declines when K = 5, which may be due to some transitions having disproportionately large or small weights, which affects the gradient descent's convergence.  
However, the improvement is significant even with just one iteration compared to the original algorithm. We typically choose a value of K between 3 and 5 for the best performance.

\begin{table}[ht]
\begin{center}
\small
\setlength{\tabcolsep}{2pt} 
\vspace{-3pt}
\caption{Effect of the number of iterations $K$ on \shortalgo (15 seeds). 
}\label{tab:iterations}
\begin{booktabs}{
  colspec = {c||c|ccccc},
}
\toprule
$K$         & vanilla & 1   & 2 & 3  & 4 & 5 \\
\midrule
BC       &   476.8   &  651.0  & 674.5 &   664.5 & \textbf{683.6} & 662.1 \\
\midrule
TD3+BC &  667.7 &	711.2 &	706.3 &	719.7 &	725.1 &\textbf{734.1} \\
\bottomrule
\end{booktabs}
\end{center}
\end{table}

\textbf{Comparison with longer training.}
\shortalgo requires additional computational cost to calculate priorities. To provide a fair comparison, we ran vanilla TD3+BC for twice as long. We found that TD3+BC converges rapidly, and the results at 2M steps were similar to those at 1M steps (677.7 vs. 672.7). This indicates that the superior performance of \shortalgo is not due to extra computation, but rather stems from the improved policy constraint.

\begin{table}[ht]
\begin{center}
\centering
\small
\setlength{\tabcolsep}{1.8pt} 
\vspace{-1mm}
\caption{Effect of decoupled resampling on Mujoco Locomition tasks. Results are the total scores of 9 tasks with 15 seeds. 
``+CNT" denotes using the prioritized sampling only for policy constraint term\protect\footnotemark.
``+DR" represents using the prioritized sampling for policy constraint and policy improvement, \ie, decoupled resampling.
``+all" denotes using the prioritized sampling for all three terms. Very low scores are marked with red. }
\label{tab:which_term}
\begin{booktabs}{
  colspec = {l|c|cccc},
}
\toprule
  & vanilla & &  +CNT     & +DR & +ALL   \\
\hline
TD3+BC& 667.7 & \shortalgo  & 723.7 & \textbf{734.1} &  \textbf{731.7}    \\
    &  & \shortalgoo  & \textcolor{red}{608.7} & \textbf{707.6} &  \textbf{707.3}      \\
\hline
CQL& 681.0 &\shortalgo   & 674.9  & \textbf{714.1} &     \textcolor{red}{652.8}   \\
   &     &\shortalgoo  & 672.3  &  \textbf{714.9} &  \textbf{708.3}    \\
\hline
IQL& 682.1& \shortalgo   & - &  \textbf{727.8} &    \textcolor{red}{674.3}    \\
   &    &\shortalgoo  & - & 706.5  &   \textbf{726.7}     \\
\hline
Onestep& 655.0& \shortalgo  & - & \textbf{702.5} &     \textcolor{red}{658.1}   \\
   &    &\shortalgoo  & - & \textbf{685.6}  &   \textbf{681.1}     \\
\bottomrule
\end{booktabs}
\end{center}
\end{table}

\footnotetext{IQL and OnestepRL utilize weighted regression, coupling policy constraint and improvement together.}

\textbf{Analysis of decoupled resampling.} 
A basic recipe of offline RL algorithms comprises policy evaluation, policy improvement, and policy constraint. 
Since \shortname focuses on producing a better behavior policy for policy constraint, it seems natural to solely apply prioritized data to the constraint term.
However, this does not always improve performance.
For instance, as~\cref{tab:which_term} shows, on TD3+BC with \shortalgoo, only prioritizing data for constraint results in a dramatic drop. 
We observed that it suffers from extrapolation error and results in Q-value overestimation in several tasks. 
We suspect it is because, when only prioritizing the constraint term, it imposes a weaker constraint on low-priority actions, while the policy improvement remains unchanged. As a result, the extrapolation error of low-priority samples accumulates. 
To validate our hypothesis, if we clip the priority weights less than 1 to 1, the results will be much better (608.7 v.s. 672.9). However, clipping gives a biased estimation to the weights and hinders the performance of \shortalgo. 
A more straightforward and effective solution is to apply data prioritization to both improvement and constraint terms. 

For applying data prioritization to policy evaluation, we empirically found that usually it severely degrades the performance except for few cases. 
We hypothesize that data prioritization changes the state-action distribution of the dataset and intensifies the degree of off-policy between the current policy and the sampling distribution. Although it does not harm policy learning, it might potentially cause instability in policy evaluation when combined with bootstrap and function approximators~\cite{sutton2018reinforcement, van2018deadly_triad, tsitsiklis1996deadly_tirad, yue2023understanding}. 
This also explains why \shortalgo is severely impaired by data prioritization for policy evaluation, whereas \shortalgoo is not. 
The underlying cause can be that \shortalgo evaluates the policy on more off-policy data obtained by multiple iterations. 

Similar effect of decoupled resampling on Antmaze are observed in \cref{tab:term-ablation2}, where prioritizing for policy evaluation significantly impairs performance. The prioritized resampling for all terms including policy evaluation culminates in near-zero scores in 4 out of the 6 tasks in~\shortalgo. It also causes inferior performance in~\shortalgoo.
In conclusion, decoupled resampling plays a pivotal role in performance and stability of resampling for offline RL.

\begin{table}[htbp]\centering
\small
\setlength{\tabcolsep}{2pt} 
\caption{Effect of decoupled resampling on Antmaze.}

\begin{tabular}{cccccc}
\toprule
&  \multirow{2}{*}{vanilla IQL}    & \multicolumn{2}{c}{\shortalgo} & \multicolumn{2}{c}{\shortalgoo}   \\
\cmidrule(lr){3-4}\cmidrule(lr){5-6}
&                & +DR & +ALL & +DR & +ALL  \\
\midrule
umaze                            &  88.5   &  85.5  & 77.3  & 87.8 & \textbf{89.2} \\
umaze-diverse                   &  63.1  &  \textbf{70.8} & 67.7 & 66.0 & \textbf{79.8}\\
medium-play                     &  70.5 & \textbf{76.1} & 0 & 72.0 & 68.4 \\
medium-diverse                 &  58.5  & \textbf{71.8} & 0 & \textbf{74.2} & 38.4\\
large-play                     &   44.1  & 40.0 & 0 & \textbf{49.6}  & 14.6 \\
large-diverse                  &  42.0    &  \textbf{48.0} & 0 & 43.0 & 37.6\\
\midrule
\rowcolor[HTML]{EFEFEF}
antmaze total                  &   366.7  & \textbf{392.2} & \textcolor{red}{145.0} & \textbf{392.6} & \textcolor{red}{328.0}\\
\bottomrule
\end{tabular}
\label{tab:term-ablation2}
\end{table}

\subsection{Comparison to Existing Methods}

\begin{figure}[ht]
\begin{center}
\vspace{-3mm}
  \centering
\includegraphics[width=0.35\textwidth]{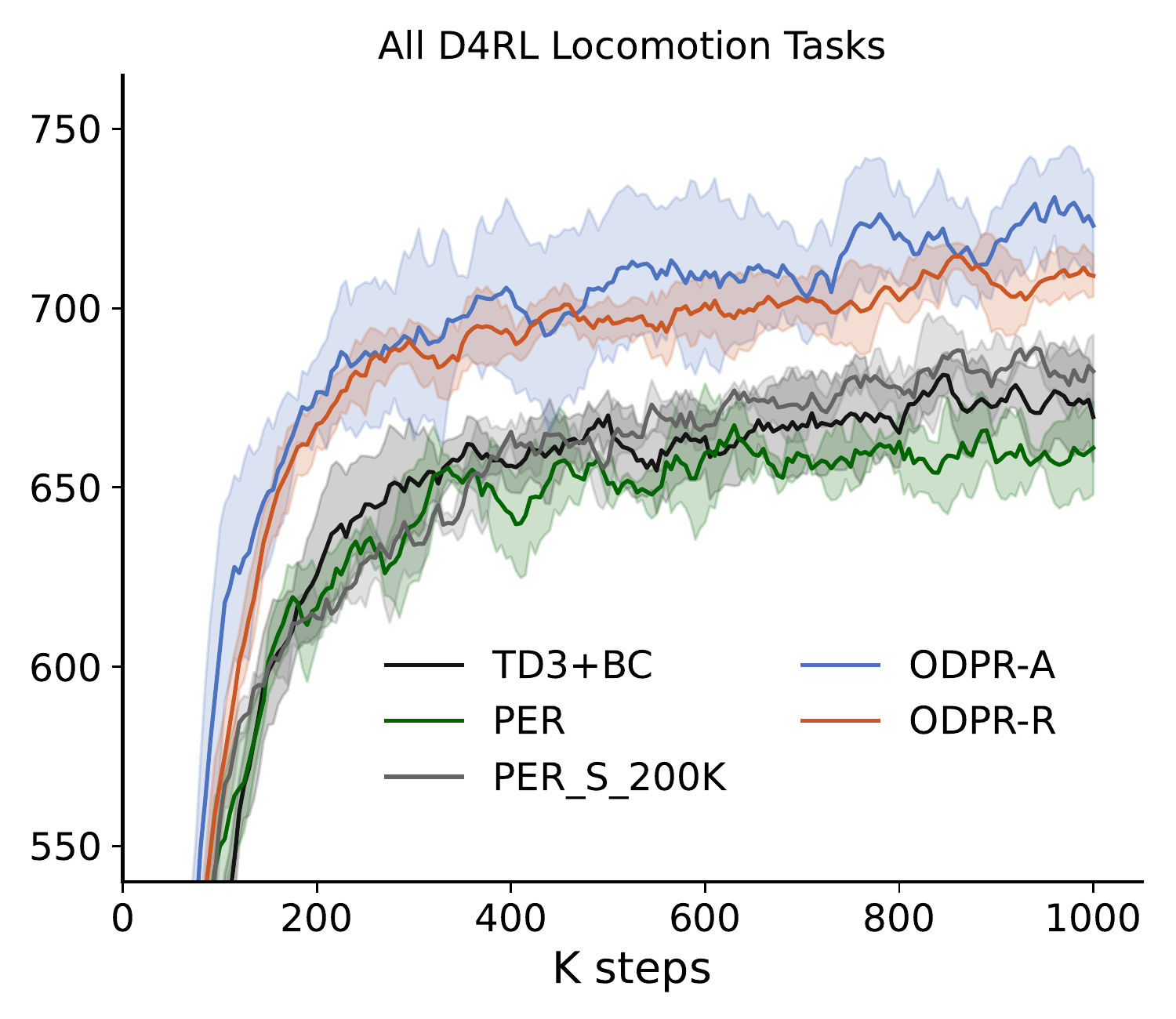}
  \vspace{-10pt}
    \caption{
    \YY{Compare the priority function of \shortname, PER~\cite{PER}, and PER\_S\_200K~\cite{zhang2023efficient}. The aggregate curves over 9 Mujoco Locomotion tasks are reported.}
    }
  \label{fig:per-compare}
\vspace{-3mm}
\end{center}
\end{figure}

\textbf{Comparison with Prioritized Sampling Methods.} We evaluate \shortname against the online prioritized sampling algorithm PER~\cite{PER}, which uses the \textbf{absolute} TD-error for dynamic priority assignment in accelerating value function fitting, and \YY{the efficient experience buffer for offline RL PER\_S\_200K~\cite{zhang2023efficient}}. 
Consider a sample with large \textbf{negative} TD errors (\ie, advantage), Unlike PER and PER\_S\_200K give high priorities to them while \shortname discourages them. Specifically, PER thinks the sample contains more information for value fitting, while \shortname thinks the sample’s action is not good behavior.
In \cref{fig:per-compare}, each curve represents the average performance across 9 Mujoco locomotion tasks. The data indicate that PER marginally reduces the effectiveness of TD3+BC in offline Mujoco settings. \YY{PER\_S\_200K shows a slight improvement over both PER and vanilla TD3+BC. In contrast, \shortname significantly enhances performance, clearly outperforming the other methods.}

\begin{figure}[ht]
\begin{center}
\vspace{-2mm}
  \centering
  \includegraphics[width=0.3\textwidth]{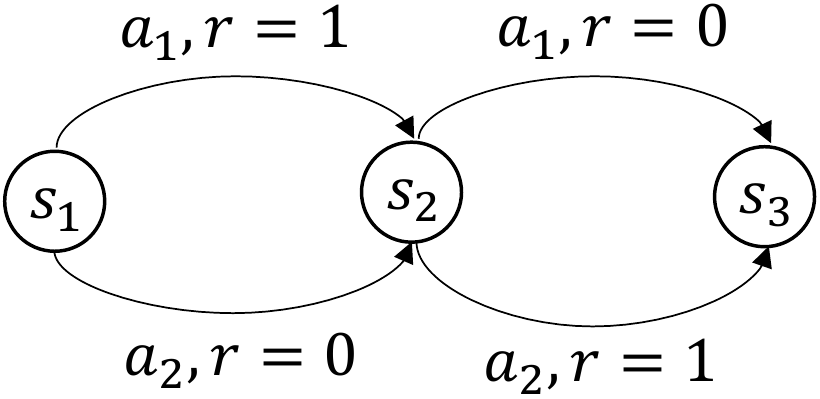}
  \caption{The toy example for trajectory concatenation.}
  \label{fig:fine-grain}
\vspace{-3mm}
\end{center}
\end{figure}

\textbf{Comparison with offline sampling methods.}
Existing resampling works such as AW/RW~\cite{hong2023harnessing} and percentage sampling~\cite{chen2021decision} assigned the same weights to \textit{all transitions in a trajectory} according to the trajectory return.
Instead, \shortalgo resamples each transition according to its advantage, which we term \textit{fine-grained priorities}. 
The fine-grained priorities also enable \shortalgo to combine favorable segments of suboptimal trajectories, thereby forming improved behavior. This concept is illustrated by~\cref{fig:fine-grain}, where we consider two suboptimal trajectories in the dataset: $\{s_1, a_1, 1, s_2, a_1, 0, s_3\}, \{s_1, a_2, 0, s_2, a_2, 1, s_3\}$. Since two trajectories yield equal returns, AW/RW would not affect the original dataset. In contrast, \shortalgo calculates normalized advantage as priority weights, assigning near-zero weight to $\{s_1, a_2, 0, s2\}, \{s_2, a_1, 0, s_3\}$, and consequently deriving an optimal dataset $\{s_1, a_1, 1, s2\}, \{s_2, a_2, 1, s_3\}$. 

Moreover, we substantiate the enhanced performance yielded by fine-grained priorities by comparing \shortalgo with AW, employing TD3+BC and IQL in the D4RL Mujoco experiments. As outlined in~\cref{tab:aw-mujoco}, \shortalgo outperforms AW in 8 out of 9 tasks in both TD3+BC and IQL, highlighting its effectiveness in achieving superior performance.

\begin{table}[!ht]
\vspace{-2mm}
\small
\setlength{\tabcolsep}{2pt} 
\centering
\caption{Compare \shortalgo with AW. AW resamples all transitions in a trajectory with the same probability without fine-grained priorites.  We maintained identical hyperparameters for both \shortalgo and AW to ensure a fair comparison. \shortalgo achieves the best score in 8 out of 9 tasks.}
\label{tab:aw-mujoco}
\begin{tabular}{ccccccc}
\toprule
& \multicolumn{3}{c}{TD3+BC~\cite{td3+bc}}                            & \multicolumn{3}{c}{IQL~\cite{IQL}}                                                               \\
\cmidrule(lr){2-4}\cmidrule(lr){5-7}
& vanilla & \shortalgo & AW            & vanilla  & \shortalgo & AW \\
\midrule
halfcheetah-m  & 48.2  & \textbf{50.0}  & 48.6          & \textbf{47.6} & \textbf{47.5}  & 45.1            \\
hopper-m       & 58.8  & \textbf{74.1}  & 61.1          & 64.3          & \textbf{73.5}  & 62.7           \\
walker2d-m     & 84.3  & \textbf{84.9}  & 80            & 79.9          & \textbf{83.1}  & 76.1         \\
halfcheetah-mr & 44.6  & \textbf{45.9}  & 45.1          & 43.4          & \textbf{44.1}  & 42.2          \\
hopper-mr      & 58.1  & \textbf{88.7}  & 87.4          & 89.1          & \textbf{103}   & 93.3          \\
walker2d-mr    & 73.6  & \textbf{88.2} & 80.3 & 69.6          & \textbf{81}    & 62.6             \\
halfcheetah-me & 93.0  & 83.3           & \textbf{97.7 }         & 83.5          & 88.9           & \textbf{93.7}  \\
hopper-me      & 98.8  & \textbf{107.3} & 102.9         & 96.1          & \textbf{100.3} & 93.3           \\
walker2d-me    & 110.3 & \textbf{111.7} & 110.2         & 109.2         & \textbf{111.4} & 108.8           \\
\rowcolor[HTML]{EFEFEF}
total          & 669.7 & \textbf{734.1} & 708.7         & 682.7         & \textbf{732.8} & 677.8          \\
\bottomrule
\end{tabular}
\vspace{-1mm}
\end{table}

Except fine-grained priorities which enables~\shortalgo to concatenate suboptimal segments, a vital differentiation between \shortname and AW/RW~\cite{hong2023harnessing} lies in decoupled resampling. Unlike AW/RW, \shortalgoo does not prioritize policy evaluation, opting instead to employ two separate samplers. As we verify in the ablation study, prioritizing policy evaluation can intensify the degree of off-policy and lead to potential performance degradation. The AW column of~\cref{tab:iql} illustrates this point, where we compared \shortalgoo/\shortalgo and AW using the default hyperparameters of IQL for a fair comparison. In Antmaze, while AW outperformed \shortname slightly in two simpler tasks, it significantly lagged in four more challenging tasks, resulting in a substantial deficit in total performance. 
This contrast highlights the value of decoupled resampling, an approach unique to \shortname.

\begin{figure}[!ht]
    \centering
    \includegraphics[width=0.5\textwidth]{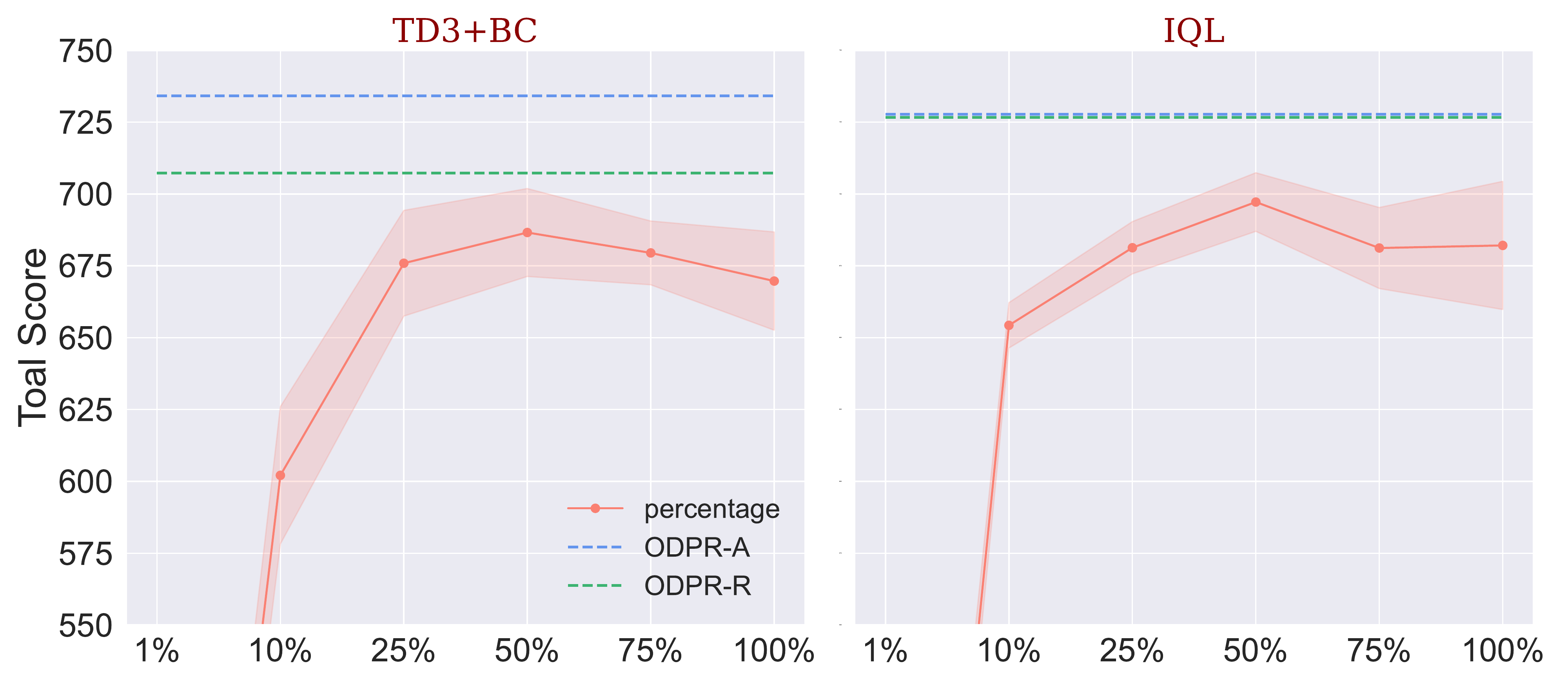}
    \caption{Compare \shortname and percentage sampling~\cite{chen2021decision} on mujoco locomotion based on TD3+BC and IQL. 50\% is the optimal value for percentage sampling, still far behind \shortname. Note that lines of \shortalgo and \shortalgoo in IQL overlap.}
    \label{fig:percentage-sampling}
\end{figure}

Percentage sampling~\cite{chen2021decision} runs algorithms such as TD3+BC and IQL on only top X\% data ordered by trajectory returns.
In~\cref{fig:percentage-sampling}, we tested values of 1\%, 10\%, 25\%, 50\%, and 75\% and found that 50\% is nearly the optimal value for both TD3+BC and IQL. However, 50\% percentage sampling still underperforms \shortname-A/R (695.2 v.s. 727.8/726.7 in IQL; 687.6 v.s. 734.1/707.3 in TD3+BC), further highlighting the effectiveness of \shortname.
For Antmaze, Kitchen, and Adroit tasks, we adopt a value of 10\%, following the approach in~\cite{hong2023harnessing}. As indicated in~\cref{tab:iql}, percentage sampling yields considerably lower scores. We attribute this to the fact that only a small fraction of data is utilized, leading to potential information loss.

\textbf{Applicability to Datasets Lacking Trajectory.}  
Even in difficult scenarios without trajectory information, a context in which AW/RW and percentage sampling failed to function without trajectory return, \shortalgo can still be employed due to its fine-grained property.
To illustrate the benefit of \shortalgo in such a scenario, we constructed transition-based datasets by randomly sampling 50\% of transitions from the D4RL Mujoco datasets.~\cref{tab:part_transition} demonstrates that \shortalgo improves the total score of TD3+BC from 647.2 to 712.7, enhancing performance in 7 out of 9 tasks. Similar enhancements were observed in the IQL case. This evidence reveals that \shortalgo can significantly boost popular offline RL algorithms even without access to trajectory information.

\begin{table}
\begin{center}
\vspace{-4mm}
\small
\caption{Performance of \shortalgo on \textbf{50\%-transition datasets in the absence of trajectory}. Existing offline resampling methods, such as AW/RW~\cite{hong2023harnessing} and percentage sampling~\cite{chen2021decision}, are unable to function without trajectory data.} 
\label{tab:part_transition}
\begin{tabular}{ccccc}
\toprule
\multirow{2}{*}{50\% Dataset} & \multicolumn{2}{c}{TD3+BC~\cite{td3+bc}}     & \multicolumn{2}{c}{IQL~\cite{IQL}}  \\
\cmidrule(lr){2-3}\cmidrule(lr){4-5}
        10 seeds                 & V       & A        & V & A         \\
\midrule
halfcheetah-m        & 48.5          & \textbf{49.9}  & 44.1    & \textbf{47.3}  \\
hopper-m             & 60.4          & \textbf{72.9}  & 57.0    & \textbf{68.8}  \\
walker2d-m           & 66.0          & \textbf{82.5}  & 67.7    & \textbf{81.7}  \\
halfcheetah-mr       & 43.2          & \textbf{45.2}  & 35.5    & \textbf{39.2}  \\
hopper-mr            & 76.4          & 74.9           & 78.5    & \textbf{91.2}  \\
walker2d-mr          & 53.3          & \textbf{89.6}  & 55.6    & \textbf{75.2}  \\
halfcheetah-me       & \textbf{89.1} & 77.7           & 90.8    & 89.9           \\
hopper-me            & 100.0         & \textbf{108.4} & 106.9   & \textbf{108.9}          \\
walker2d-me          & 110.3         & \textbf{111.6} & 109.4   & \textbf{112.2} \\
\rowcolor[HTML]{EFEFEF}
total                & 647.2         & \textbf{712.7} & 645.5   & \textbf{714.4} \\
\rowcolor[HTML]{EFEFEF}
SD (total)           & 25.2          & 13.1           & 23.9    & 14.7  \\
\bottomrule    
\end{tabular}
\vspace{-2mm}
\end{center}
\end{table}

\YY{
\subsection{How does \shortname Perform When Value Estimations Are Inaccurate?}
Value estimation can vary significantly depending on the dynamics of the environment and the quality of the data available~\cite{kumar2020discor}.
Although value estimation in \cref{eqn:estimate_value} works well in D4RL environments, we investigate how does \shortalgo performs when value estimations are inaccurate, which bears significance for the real application of \shortalgo in scenarios. 
We conducted experiments where we intentionally introduced bias by adding Gaussian noise $N(0, \sigma)$ or uniform noise $[0,2\sqrt{3} \sigma]$ to the estimated advantage of each state-action pair in the dataset. This simulates scenarios with imperfect advantage estimation. $\sigma$ is the standard deviation of the unmodified advantages within the dataset, representing a relatively substantial perturbation to advantage estimations.
As shown in~\cref{tab:adv_noise}, \shortalgo~maintains stable performance despite significant noise, demonstrating its robustness under noisy advantage estimation conditions. 
}


\begin{table}[htb!]
\vspace{-1mm}
\centering
\small
\caption{Mujoco Total Score of \shortalgo with TD3+BC under conditions of inaccurate advantage estimation.}
\begin{tabular}{cc}
\hline
Method & Score \\
\hline
TD3+BC & 669.7 \\
\shortalgo & 734.1 \\
\shortalgo + Gaussian Noise & 733.2 \\
\shortalgo + Uniform Noise & 720.8 \\
\hline
\end{tabular}
\label{tab:adv_noise}
\vspace{-1mm}
\end{table}

\begin{table}[htb!]
\vspace{-1mm}
\centering
\small
\caption{Mujoco Total Score of \shortalgo with TD3+BC under conditions of highly noisy state.}
\begin{tabular}{lcc}
\hline
Method & No Noise & w. State Noise \\
\hline
TD3+BC & 669.7 & 470.3 \\
\shortalgoo & 707.3 (+37.6) & 504.7 (+34.4) \\
\shortalgo & 734.1 (+64.4) & 515.7 (+45.2) \\
\hline
\end{tabular}
\label{tab:state_noise}
\vspace{-1mm}
\end{table}

\YY{
\subsection{Performance of \shortname in Highly Noisy Data Environments}
We assessed the robustness of \shortname by introducing Gaussian noise \(N(0, 0.1)\) to the states in the dataset. The results, presented in \cref{tab:state_noise}, show that although both TD3+BC and ODPR experience declines in performance with noisy states, ODPR-A and ODPR-R maintain substantial performance advantages of 45.2 and 34.4 points, respectively, over TD3+BC. These findings underscore the efficacy of ODPR even under conditions of significant data noise.
}

\subsection{Computational Cost}
\label{sec:time}
Training the value network to estimate priority in \shortalgo (when K=4) does not necessitate four times the computational resources compared to standard TD3+BC.
Comparing to TD3+BC, \shortalgo requires no actor update and does not need to query the actor during policy evaluation, making it less time-intensive per step. In our tests conducted on an NVIDIA 3090, 1M gradient steps took 69 minutes for the official TD3+BC, whereas our JAX implementation of \shortalgo took only 28 minutes for K=4, translating to a 40\% increase in time. 
Moreover, we highlight that the priority weights, which are relevant only to the dataset and independent of the algorithms, need to be generated only once by \shortalgo and can be reused across different algorithms.
We have made the priority weights of the D4RL datasets publicly available, enabling researchers to directly utilize them to enhance their algorithms without extra computational costs. 
In conclusion, \shortalgo does not significantly increase the computational burden.

\YY{For \shortalgo, we traverse each trajectory in the dataset and calculate its return by summing the rewards before training begins, which incurs very slight additional computation compared to the training. For example, processing a large dataset of 1 million transitions takes approximately one minute.
}
 \section{Related Works }
\label{sec:related-works}
\textbf{Offline RL with Behavior Regularization.}
To alleviate the distributional shift problem, a general framework employed by prior offline RL research is to constrain the learned policy to stay close to the behavior policy. 
Many works~\cite{jaques2019way, wu2019behavior} opt for KL-divergence as policy constraint.
Exponentially advantage-weighted regression (AWR), an analytic solution of the constrained policy search problem with KL-divergence, is adopted by AWR~\cite{peng2019advantage}, CRR~\cite{wang2020critic} and AWAC~\cite{nair2020awac}.
IQL~\cite{IQL} follows AWR for policy improvement from the expectile value function that enables multi-step learning.
BEAR~\cite{kumar2019stabilizing} utilizes maximum mean discrepancy (MMD) to approximately constrain the learned policy in the support of the dataset, while Wu~\etal~\cite{wu2019behavior} find MMD has no gain over KL divergence.
Other variants of policy regularization include the use of Wasserstein distance~\cite{wu2019behavior} and BC~\cite{td3+bc, wang2023diffusion, chen2022offline, yang2023hundreds}.
An alternative approach to regularize behavior involves modifying the Q-function with conservative estimates~\cite{CQL, buckman2020pessimism, yu2021combo, sac-n, lb-sac, ghasemipour2022so,ma2021conservative}. 

\textbf{Prioritized Sampling.} 
Many resampling methods, i.e., prioritization methods, have been proposed for RL in an online setting, including PER~\cite{PER}, DisCor~\cite{kumar2020discor}, LFIW~\cite{sinha2022lfiw}, PSER~\cite{brittain2019pser}, ERE~\cite{wang2020striving}, and ReMER\cite{liu2021regret}. 
These methods mainly aim to expedite temporal difference learning.
SIL~\cite{oh2018self} only learns from data with a discounted return higher than current value estimate.
Strategies focused on offline-to-online settings are explored by~\cite{lee2022offline} and~\cite{pong2022offline}. 
In offline RL, schemes based on imitation learning~(IL) aim to learn from demonstration, naturally prioritizing data with high return. These approaches include data selection~\cite{chen2020bail, liu2021curriculum} and weighted imitation learning~\cite{wang2018exponentially}.
BAIL~\cite{chen2020bail} estimates the optimal return, based on which good state-action pairs are selected to imitate. 
For RL-based learning from offline data,
CQL (ReDS)~\cite{singh2022offline} is specifically designed for CQL to reweight the data distribution.
It works via a modified CQL constraint where the values of bad behaviors are being penalized.This approach is non-trivial to transfer to algorithms like TD3+BC or IQL, since they lack existing components to penalize certain actions.
In contrast, \shortname serves as a plug-and-play solution, designed to enhance a broad range of offline RL algorithms.
\shortname might be the preferable choice for tasks where other SOTA policy constraint algorithms outperform CQL, such as in the Antmaze or computation-limited environments.
AW/RW~\cite{hong2023harnessing} and ReD~\cite{yue2022boosting} proposed to reweight the entire trajectories according to their returns.
Although sharing some conceptual similarities, our method offers a more fine-grained approach by resampling transitions rather than entire trajectories.
Another distinction lies in our use of two samplers for promoting performance and stability, a uniform sampler for policy evaluation, and a prioritized sampler for policy improvement and policy constraint.



\section{Conclusion and Limitation}
\label{conclusion}
This paper proposes a plug-and-play component \shortname for offline RL algorithms by prioritizing data according to our proposed priority function. 
Furthermore, we theoretically show that a better policy constraint is likely induced with \shortname.
We develop two practical implementations, \shortalgo and \shortalgoo, to compute priority. 
Thanks to the advantages of decoupled resampling and fine-grained priorities, extensive experiments show that \shortname effectively enhance the performance of popular RL algorithms compared to other resampling methods. Furthermore, \shortalgo proves to be effective even in challenging scenarios without trajectory.
The iterative computation of \shortalgo adds an extra computational burden. In this paper, \shortalgo mitigates this issue by sharing weights across algorithms and efficient JAX implementation for value fitting. 
More efficient methods, such as implementing resampling and value fitting simultaneously to converge to the limit, remains a promising avenue for future research.

\section{Acknowledge}
This work is supported in part by the National Natural Science Foundation of China under Grants 42327901.



\appendices


\section{Proof and Derivation}

\subsection{Proof of \texorpdfstring{\protect\cref{thm:beta_improve}}{Proof of Behavior Policy Improvement}}
\label{appendix:beta_improve}

\textbf{Behavior Policy Improvement Guarantee}
Following Performance Difference Lemma (\cref{eqn:PDL}), we have
\begin{align} 
    J(\beta^\prime) - J(\beta) = \int_\rvs d_{\beta^\prime}(\rvs) \int_\rva \beta^\prime(\rva | \rvs) A^\beta(\rvs, \rva) \ d\rva \ d\rvs.
\end{align}
For simplicity, we use $A^\beta$ instead of $A^\beta(\rvs, \rva)$.  The inner integral is:
\begin{align} 
     & \int_\rva \beta^\prime(\rva | \rvs) A^\beta \ d\rva \\
     =& \int_\rva \frac{\omega(A^\beta) \beta(\rva|\rvs)}{\int_\rva \omega(A^\beta) \beta(\rva|\rvs) d\rva} A^\beta \ d\rva \label{eqn:bc1} \\
     =& \frac{\int_\rva \left( \omega(A^\beta) - \omega(0) \right) \beta(\rva|\rvs) A^\beta \ d\rva}{{\int_\rva \omega(A^\beta) \beta(\rva|\rvs) d\rva}}. \label{eqn:bc2}
\end{align}
The derivation from~\cref{eqn:bc1} to~\cref{eqn:bc2} utilizes the property of advantage $\int_\rva A^\beta \beta(\rva|\rvs) d\rva$ = 0 and $\omega(0)$ is a constant with respect to action. The sign of the integrand in \cref{eqn:bc2} depends on $\left( \omega(A^\beta) - \omega(0) \right) A^\beta$. Since $\omega(A^\beta)$ is monotonic increasing with respect to $A^\beta$,  $A^\beta$ and $\omega(A^\beta) - \omega(0)$ have an identical sign. The integrand is always non-negative, which implies that $J(\beta^\prime) - J(\beta) \geq 0$ always holds. 
If there exists a state $\rvs$, under which not all actions in action support $\{ \rva|\beta(\rva|\rvs) > 0, \rva \in \mathcal{A} \}$ have zero advantage, the inequation strictly holds.  By the definition of advantage, all actions have zero advantage if and only if all actions have the same $Q$-value.
To summarize, \cref{thm:beta_improve} suggests that policy improvement is ensured if the current policy is weighted according to its normalized advantage. This concept echoes the core principle of policy gradient methods that optimize the likelihood of actions in proportion to the magnitude of their advantage.

\section{Experiment Setup}
\label{appendix:exp-setup}

\subsection{Toy Bandit}
\label{appendix:bandit_setup}
 The offline dataset is collected by four 2D Gaussian distribution policies with means $\textstyle \boldsymbol{\mu}\! \in\! \{(0.5, 0.5), (0.5, -0.5), (-0.5, 0.5), (-0.5, -0.5)\}$, standard deviations $\textstyle \boldsymbol{\sigma} \!=\! (0.10, 0.10)$, and correlation coefficient $\rho=0$. 
The reward of each action is sampled from a Gaussian distribution, whose mean is determined by its action center and the standard deviation is 0.5.
Each policy contributes 250 samples to the dataset, imitating real scenarios where various policies collect data. 

\subsection{\shortname}
\label{appendix:oper_setup}
\paragraph{\shortalgo experiment setup.}
In \cref{eqn:estimate_value}, we adopt double value network to fit the value function ~\cite{Double-Q, td3}. 
Specifically, we employ a one-step bootstrap for value function fitting. 
For each iteration in value function fitting, we utilize 0.5M gradient steps. Pure value fitting comsumes less computation time than policy iteration.
While this lengthy number of gradient steps ensures convergence, a lesser number may suffice in practice.
For Mujoco tasks, BC and TD3+BC prioritize data using priority weights from the 4th and 5th iterations, respectively, achieving optimal results (refer to \cref{tab:iterations}). 
For other algorithms, \ie, CQL, IQL, and OnestepRL, we do not perform hyperparameter search and, following BC, simply set the number of iterations to 4.

\textit{Scale Deviation.} \shortalgo~utilizes \cref{eqn:linear-weight} to calculate weights. However, most of the weights are close to 1, weakening the effect of data prioritization.
Therefore, we adjust the standard deviation of weights to the hyperparameter $\sigma$ through a process of normalization and affine transformation.
Across all Mujoco tasks, we set $\sigma$ to 2.0. For Antmaze tasks, we use $\sigma=5.0$, while for Kitchen and Pen tasks, we set $\sigma=0.5$.

\paragraph{\shortalgoo experiment setup.}
The base priority $p_\text{base}$ in~\cref{eqn:p_return} is set to zero across all tasks, with the exception of Antmaze where it is assigned a value of 0.2. This exception is made because the trajectory return in Antmaze can either be zero or one. If $p_\text{base}$ were set to zero, all trajectories with a return of zero would be disregarded. 

\paragraph{Resampling Implementation.} 
PER~\cite{PER} employs a dynamic update of priorities during training and implements a sophisticated "sum-tree" structure to optimize efficiency. Conversely, \shortname first computes priorities from the offline dataset in an initial stage, and subsequently employs static priorities throughout training. This static priority sampling is straightforwardly implemented using ``np.random.choice".
While the utilization of a sum tree allows for sampling from a list of probabilities in $O(log n)$ time, an aspect independent of online versus offline scenarios, the use of ``np.random.choice" incurs a linear $O(n)$ cost for sampling from the same list of probabilities.
However, it's important to note that the single operation of np.random.choice" is time-efficient. Additionally, since \shortname's priorities are static, we can pre-sample the entire index list, thus accelerating the sampling process due to the parallelized implementation of np.random.choice". In the context of training on a D4RL dataset (comprising 1M data points with priority, 1M gradient steps, and a batch size of 256), it only takes approximately 53.8 seconds to execute all ``np.random.choice" operations. This adds a negligible time cost to the overall training process. 
Furthermore, the implementation of \shortname demands merely 10 lines of code changes, which contrasts with the more complex implementation of a sum tree. Therefore, from a practical standpoint, implementing ``np.random.choice" is fairly straightforward and only marginally increases the actual runtime.
Lastly, for larger dataset sizes (such as 10M or 100M) and a larger number of gradient steps, the additional time cost can be further mitigated by parallelizing the index generation process with the agent training process.

\subsection{Offline RL algorithms}
\label{appendix:offrl_setup}
For a fair comparison with the baselines, we implement \shortname on top of the official implementation of OnestepRL, TD3+BC, and IQL; for CQL, we use a reliable third-party implementation\footnote{\url{https://github.com/young-geng/JaxCQL}}, which, unfortunately, causes a slight discrepancy with PyTorch version results reported in the CQL paper. We run every algorithm for 1M gradient steps and evaluate it every 5,000 steps, except for antmaze-v0 environments, in which we evaluate every 100,000 steps. For antmaze-v0, each evaluation contains 100 trajectories; for others, 10 trajectories are used following IQL~\cite{IQL}.
For OnestepRL, we choose exponentially weighted regression as the policy improvement operator. The original paper~\cite{onestep} does not specify the value of temperature $\tau$. Therefore, we search on a small set$\{0.1, 0.3, 1.0, 3.0, 10.0, 30.0\}$ and use $\tau=1$ because it can reproduce the reported results.

\section{Additional Experiments}

\subsection{Ablation Study of Two Samplers}
\label{appendix:two-sampler}
\begin{table}[htbp]
\centering
\small
\caption{Ablation Study of Two samplers.}
\label{tab:two-sampler}
\begin{tabular}{lccc}
\toprule
& vanilla & two samplers & \shortalgo \\
\midrule
mujoco-v2 total & 667.7 & 674.9 & \textbf{734.1} \\
\bottomrule
\end{tabular}
\end{table}
As shown in \cref{sec:ablation}, for \shortalgo, the best results are obtained by decoupled resampling, where two samplers are employed, one for uniform sampling and one for prioritized sampling.
We conduct experiments here to prove that the introduction of two samplers itself does not improve performance, and the improvement comes from prioritized replay.
The original TD3+BC with one uniform sampler scored 667.7 points on Mujoco locomotion. 
Then we use two uniform samplers for the actor (policy constraint and improvement) and critic (policy evaluation), respectively. The result (674.9 points) is quite similar to the vanilla one.
Then we use a prioritized sampler for the actor and a uniform sampler for the critic, respectively, achieving a high score of 734.1 points.
It implies that the improvement comes from prioritized replay rather than two samplers.

\subsection{Results with Standard Deviation}
\label{sec:result_with_std}
Results with standard deviation are reported in \cref{tab:mujoco_std}. For Mujoco locomotion tasks, \shortname consistently achieves a performance boost, as evidenced by the non-overlapping deviation intervals. 
\begin{table}[ht]
   \centering
\small
\caption{Averaged normalized scores of TD3+BC on MuJoCo locomotion v2 tasks over 15 seeds.}
\label{tab:mujoco_std}
\begin{tabular}{cccc}
\toprule
\multirow{2}{*}{Dataset} & \multicolumn{3}{c}{TD3+BC}  \\ 
\cmidrule(lr){2-4}
& V     & A              & R \\
\cmidrule(lr){1-1}\cmidrule(lr){2-4}
halfcheetah-m  & 48.3 ± 0.1   & \textbf{50.0 ± 0.1}   & 48.6 ± 0.1           \\
hopper-m       & 57.3 ± 1.4   & \textbf{74.1 ± 2.8}   & 59.1 ± 1.2           \\
walker2d-m     & 84.9 ± 0.6   & 84.9 ± 0.3            & 84.2 ± 0.3           \\
halfcheetah-mr & 44.5 ± 0.2   & 45.9 ± 0.4            & 44.6 ± 0.4           \\
hopper-mr      & 58.0 ± 5.8   & \textbf{88.7 ± 5.5}   & \textbf{77.4 ± 4.2}  \\
walker2d-mr    & 72.9 ± 8.7   & \textbf{88.2 ± 2.0}   & 82.7 ± 2.2           \\
halfcheetah-me & 92.4 ± 0.5   & 83.3 ± 3.0            & 93.9 ± 0.7           \\
hopper-me      & 99.2 ± 6.3   & \textbf{107.3 ± 4.1}  & \textbf{106.7 ± 3.1} \\
walker2d-me    & 110.2 ± 0.2  & 111.7 ± 0.2           & 110.1 ± 0.1          \\
\midrule
\rowcolor[HTML]{EFEFEF}
total          & 667.7 ± 18.4 & \textbf{734.1 ± 10.4} & \textbf{707.3 ± 7.9} \\
\bottomrule
\end{tabular}
\vspace{-2mm}
\end{table}

\subsection{How does \shortname Perform When the Behavior Policy Exhibits Similar and Poor Performance}
\label{appendix:similar-policy}
We ran a comparative study using vanilla TD3+BC and the \shortalgoo variation on three D4RL random datasets. As displayed in~\cref{tab:random}(b), \shortname's performance mirrors the vanilla. This observation aligns with our key hypothesis that \shortname's enhancements are primarily sourced from the behavioral diversity inherent in the dataset. In contrast, the random dataset, generated through a poor policy, exhibits limited behavioral variance.

\begin{table}[htb!]
\centering
\small
\caption{\shortalgoo on random dataset (10 seeds).}
\label{tab:random}
\begin{tabular}{ccc}
\toprule
\multirow{2}{*}{Random Dataset} &  \multicolumn{2}{c}{TD3+BC}     \\ 
\cmidrule(lr){2-3}
& Vanilla     & \shortalgoo         \\
\midrule
halfcheetah  & 9.8  & 10.3                \\
hopper      & 8.4  & 8.3                 \\
walker2d     & 0.9  & 1.0       \\
\rowcolor[HTML]{EFEFEF}
total          & 19.1± 1.8 & 19.6± 1.6 \\
\bottomrule
\end{tabular}
\vspace{-2mm}
\end{table}


\subsection{The effect of Hyperparameter \texorpdfstring{$\sigma$}{σ}}
Intuitively, if the weights of all transitions are close to 1, our methods degrade to the vanilla offline RL algorithm. Only when the standard deviation of the weights of transitions is relatively large, \shortname can take effect. We observed that the standard deviation of original \shortalgo~weights in~\cref{eqn:p_adv} is typically small, ranging from approximately 0.02 to 0.2 across different environments. In contrast, the standard deviation of \shortalgoo~weights falls within a suitable range of around 0.3 to 1.0.
Thus, the standard deviation of \shortalgo~weights needs to be scaled, while \shortalgoo~can work without scaling. We test the effect of $\sigma$ on three environments where \shortalgo gives the clearest improvements. In \cref{tab:std}, we demonstrate how the performance of \shortalgo~is influenced by the hyperparameter $\sigma$, which the standard deviation of weights will be scaled to. We select 2 as the default value for $\sigma$.

\begin{table}[htpb]\centering
\small
\vspace{-3pt}
\caption{Effect of $\sigma$ on \shortalgo. The results come from TD3+BC with 15 seeds. ``w.o. scale" denotes disabling scaling.
}\label{tab:std}
\begin{tabular}{c c cccc  }
\toprule
$\sigma$         & vanilla &  w.o. scale  & 0.5 & 2.0  & 4.0 \\
\midrule
hopper-mr     & 57.9 & 70.1 & 73.8 & \textbf{88.7} & \textbf{88.9} \\
walker-mr     & 73.1 & 81.9 & 84.9 & \textbf{88.2} &  86.6 \\
hopper-me     & 98.5 & 99.1   & \textbf{106.9} & \textbf{107.3} & 105.1  \\
\midrule
\rowcolor[HTML]{EFEFEF}
total           & 229.5 & 251.1 & 265.6 & \textbf{284.2} &  \textbf{280.6} \\
\bottomrule
\end{tabular}
\vspace{-2mm}
\end{table}

\subsection{Resampling v.s. Reweighting}
\label{appendix:sample-and-weight}
\begin{figure*}[!htb]
\vspace{-3mm}
    \scriptsize
	\centering
	\begin{tabular}{c c c}
	    halfcheetah-medium-v2 	 & hopper-medium-v2 & walker2d-medium-v2  \\
		\includegraphics[width=0.16\textwidth]{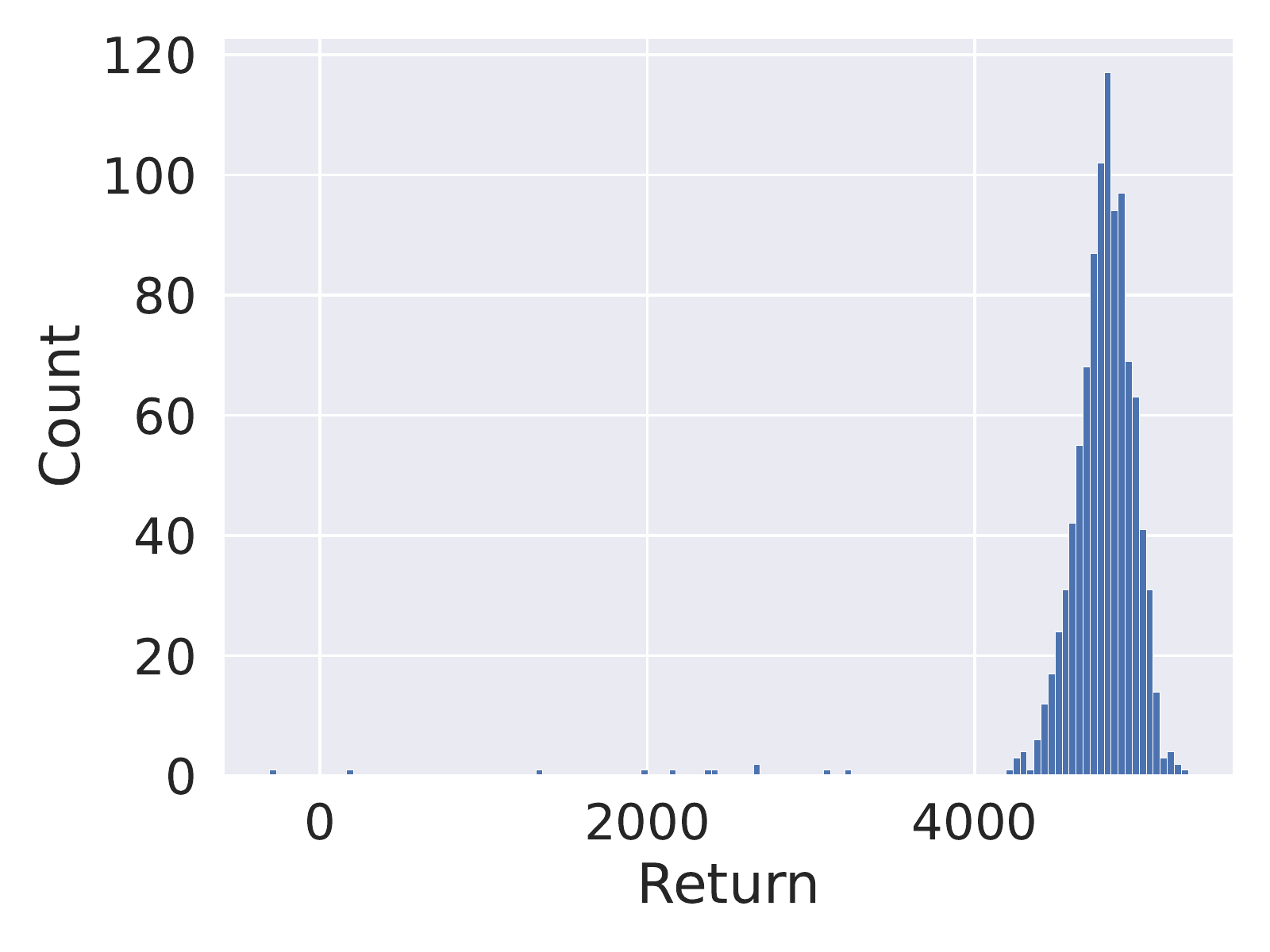} &
		\includegraphics[width=0.16\textwidth]{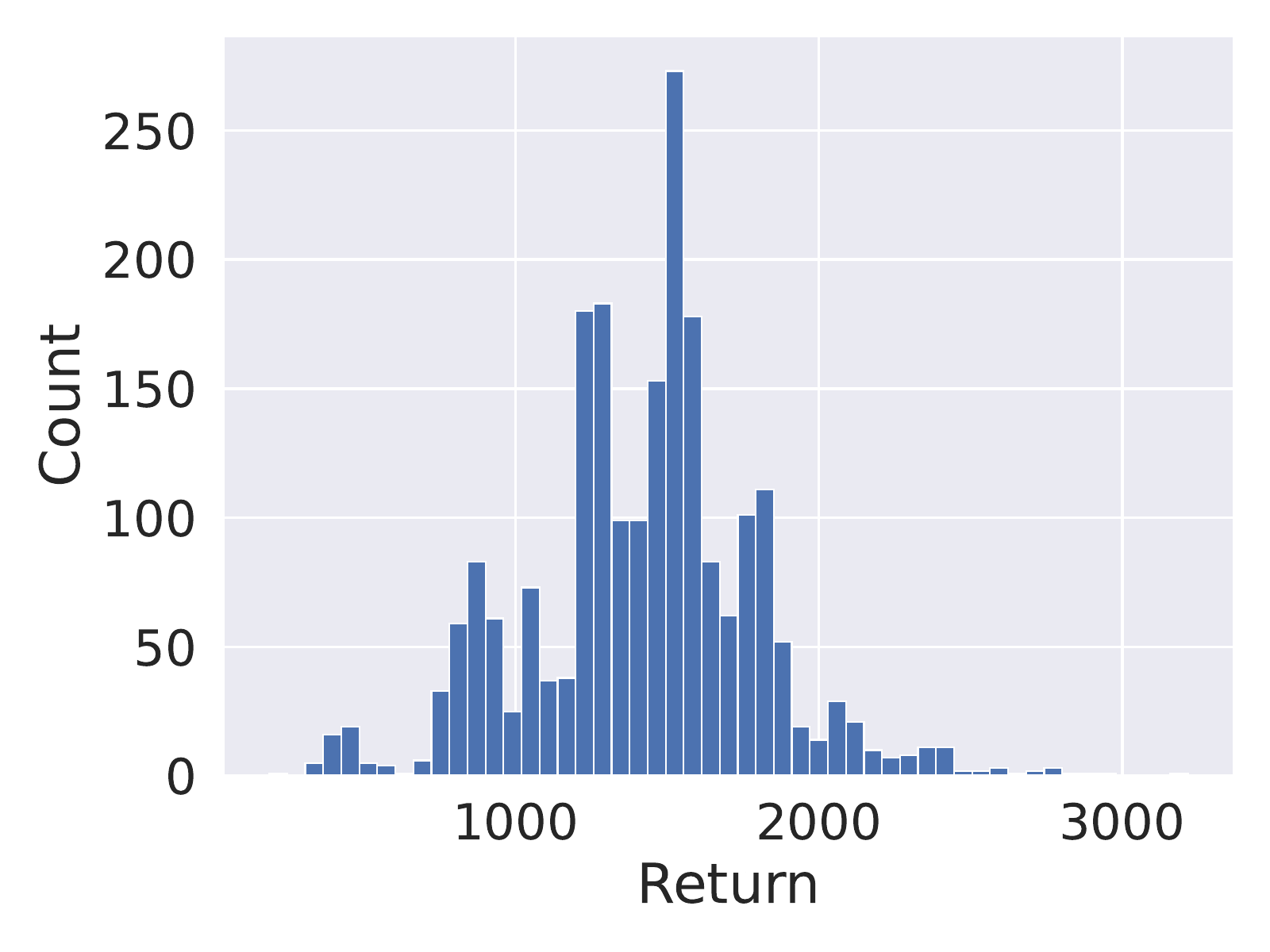}&
		\includegraphics[width=0.16\textwidth]{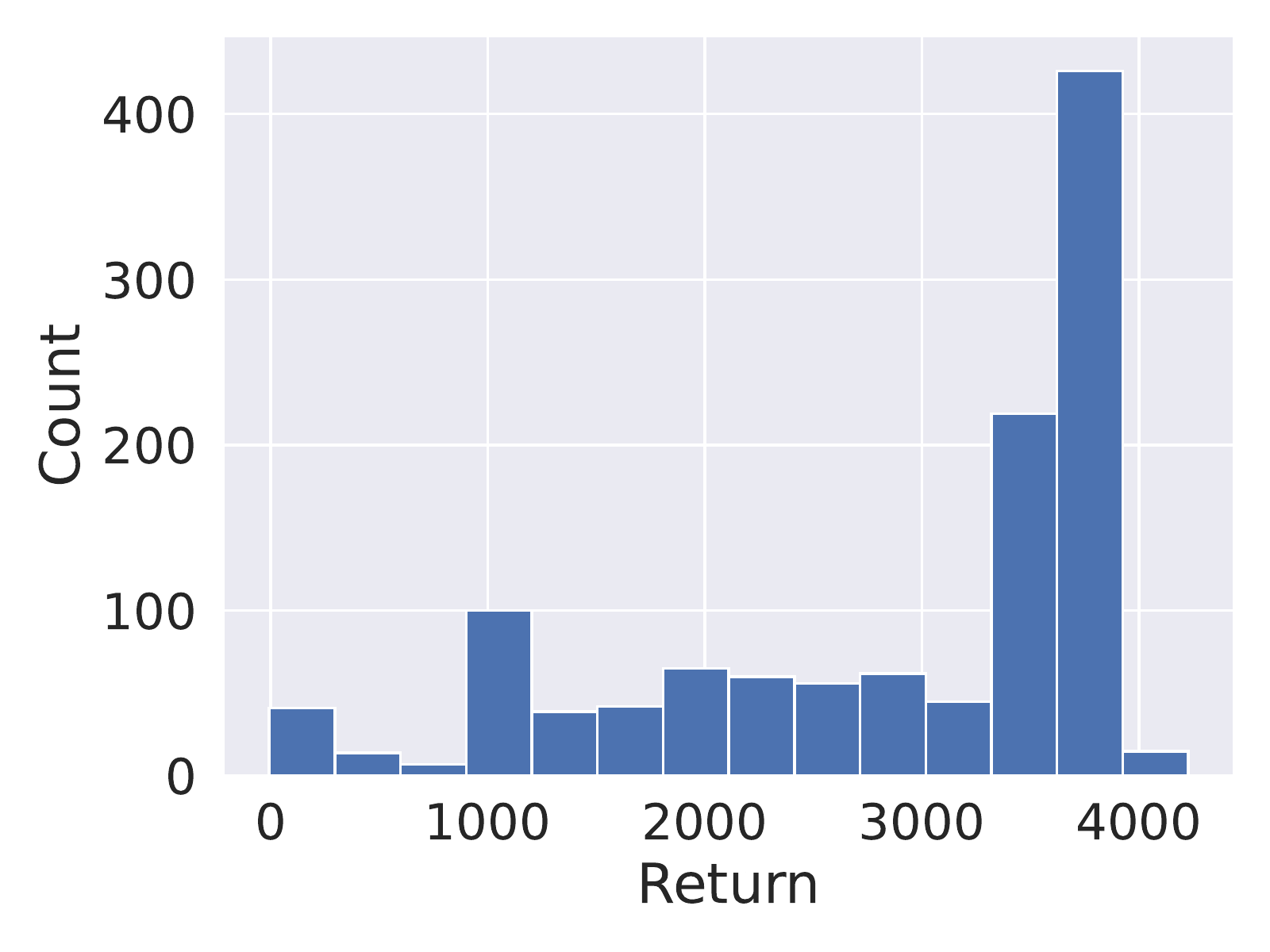} \\
		
		halfcheetah-medium-replay-v2 & hopper-medium-replay-v2 & walker2d-medium-replay-v2 \\
		\includegraphics[width=0.16\textwidth]{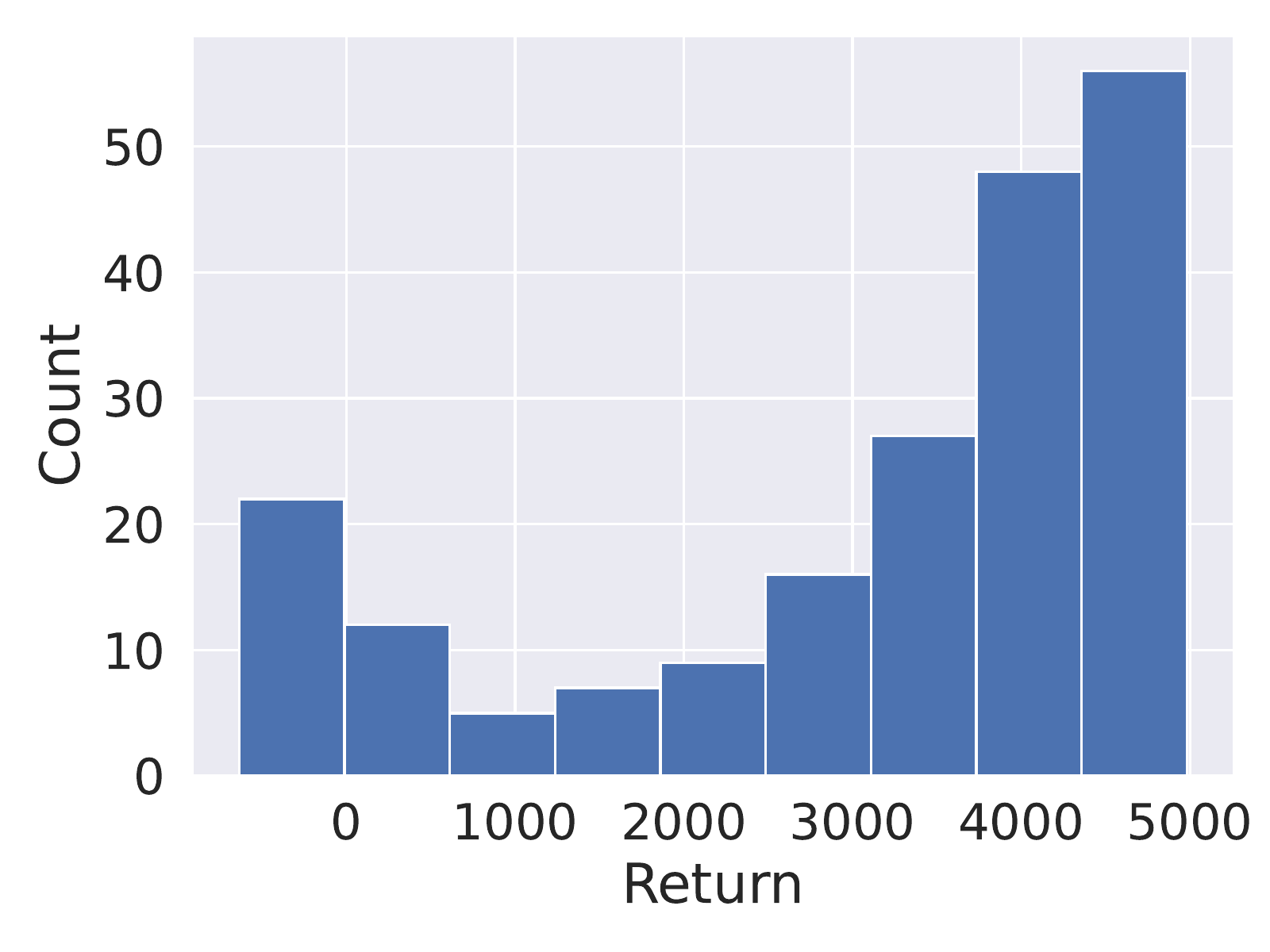}&
		\includegraphics[width=0.16\textwidth]{image/dist/hopper-medium-replay-v2.pdf} &
		\includegraphics[width=0.16\linewidth]{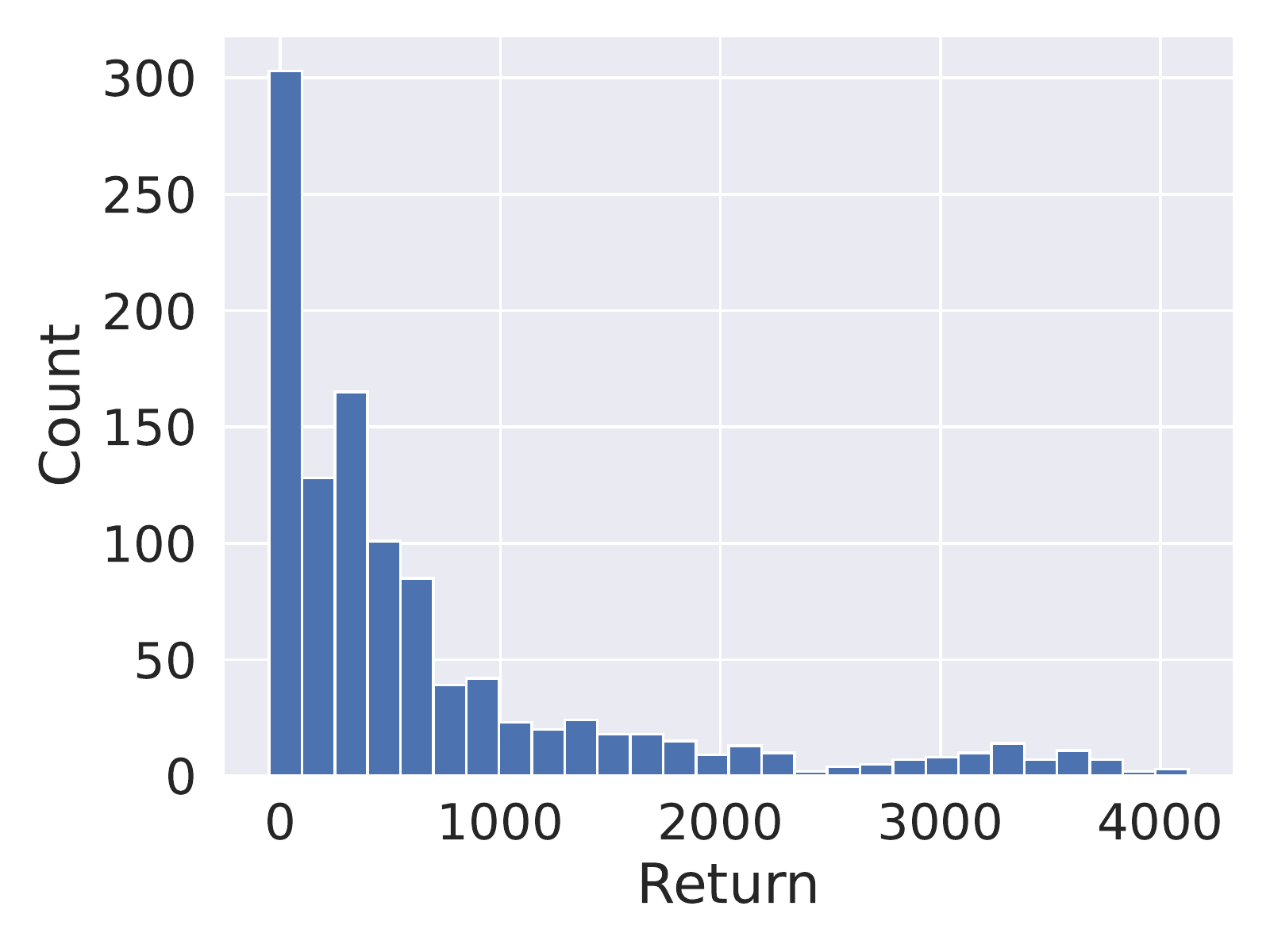} \\
        
        halfcheetah-medium-expert-v2 & hopper-medium-expert-v2 & walker2d-medium-expert-v2 \\
        
		\includegraphics[width=0.16\linewidth]{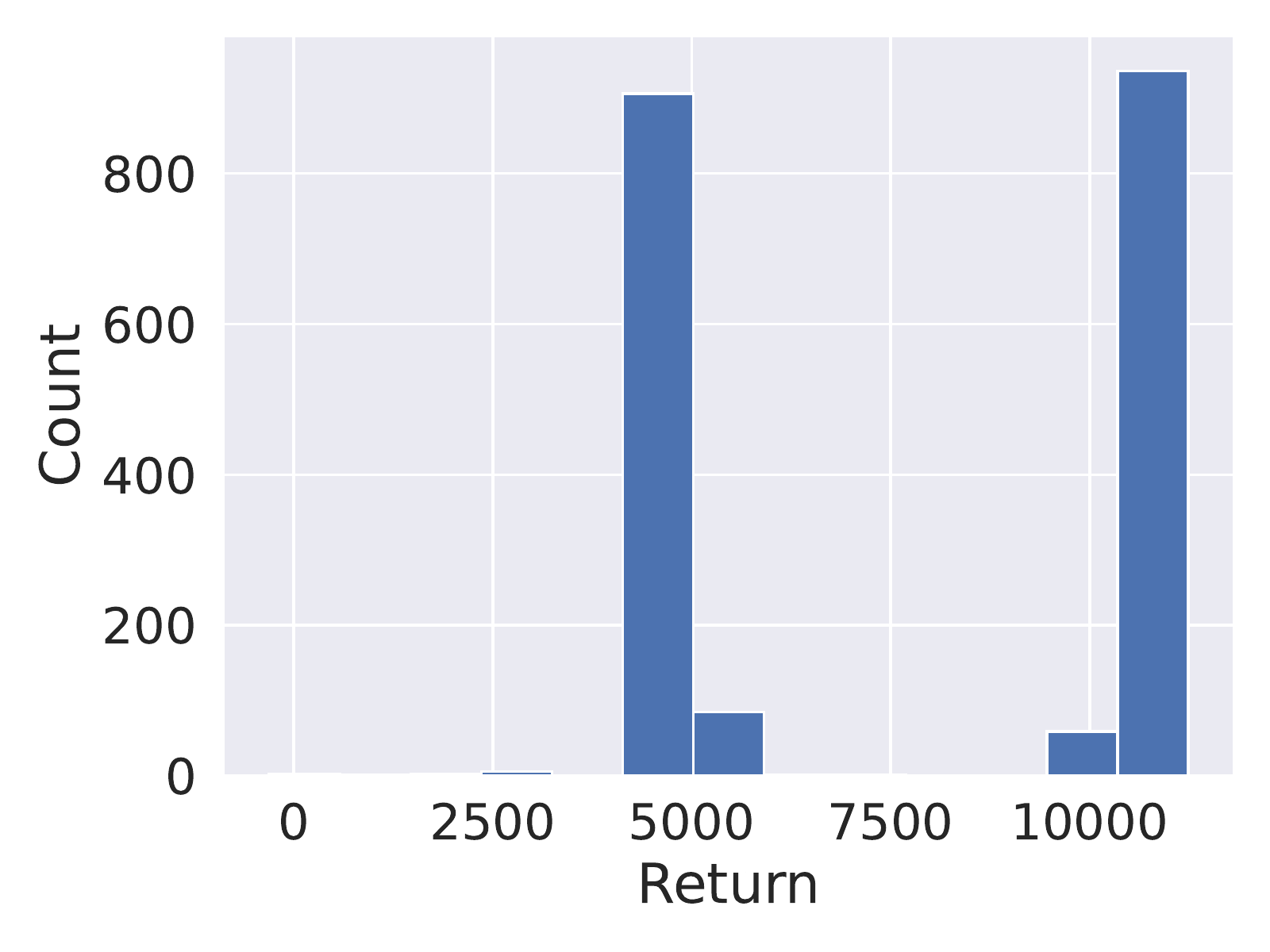}&
		\includegraphics[width=0.16\linewidth]{image/dist/hopper-medium-expert-v2.pdf}&
		\includegraphics[width=0.16\linewidth]{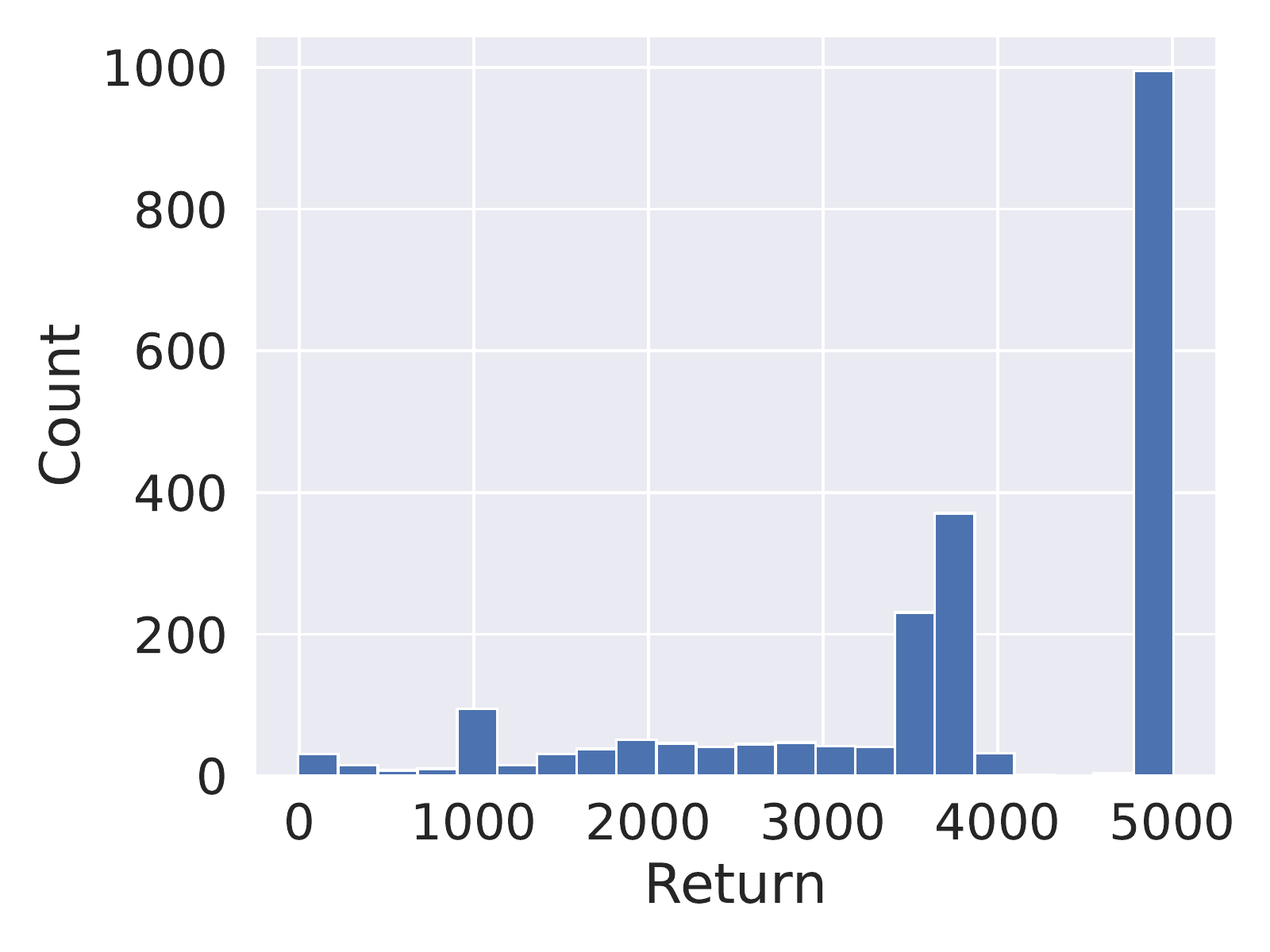} \\
		&(a) Mujoco Locomation &\\ \\
		
		antmaze-umaze-v0 & antmaze-umaze-diverse-v0 & antmaze-medium-play-v0 \\
		\includegraphics[width=0.16\linewidth]{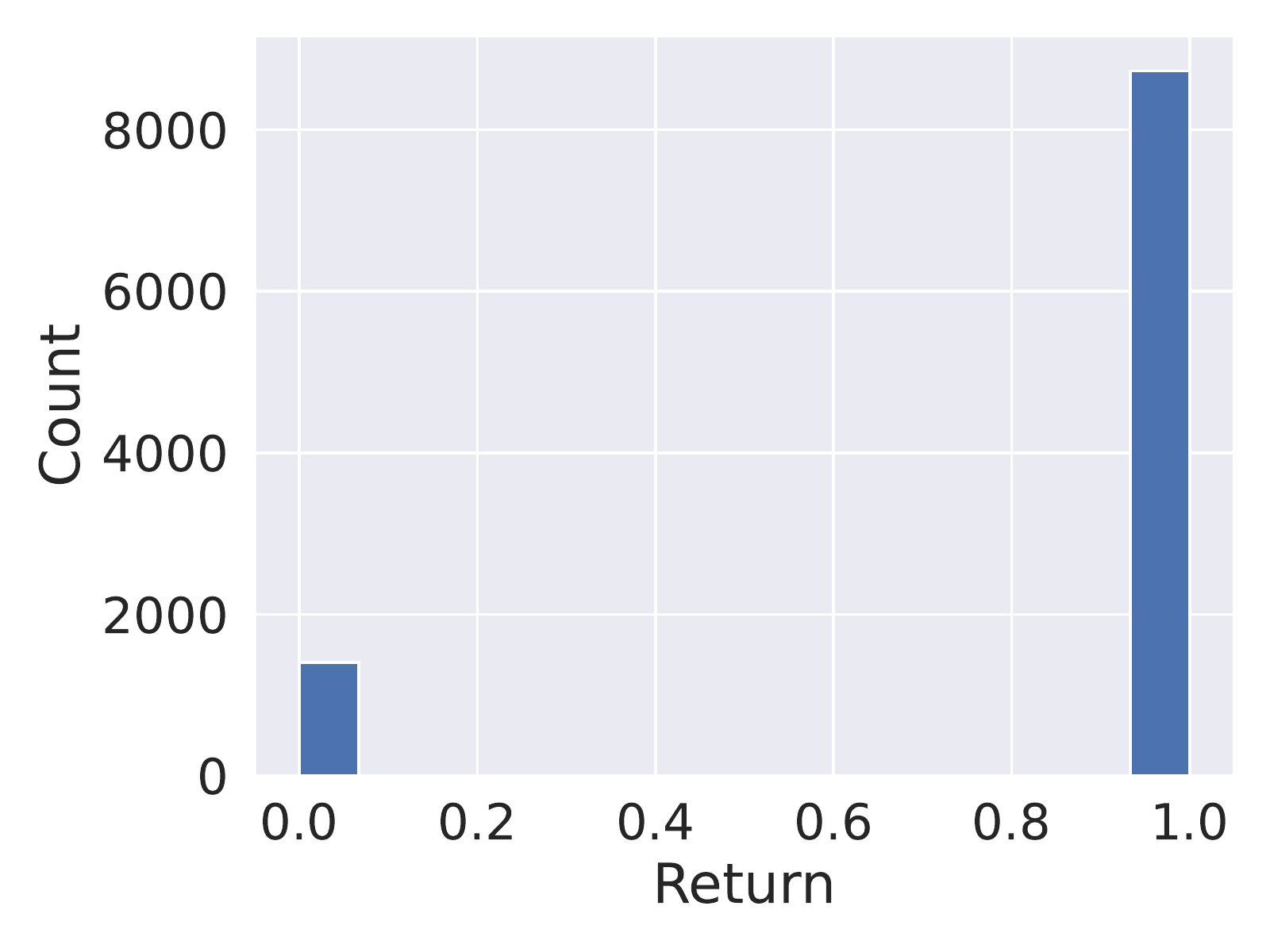} &
		\includegraphics[width=0.16\linewidth]{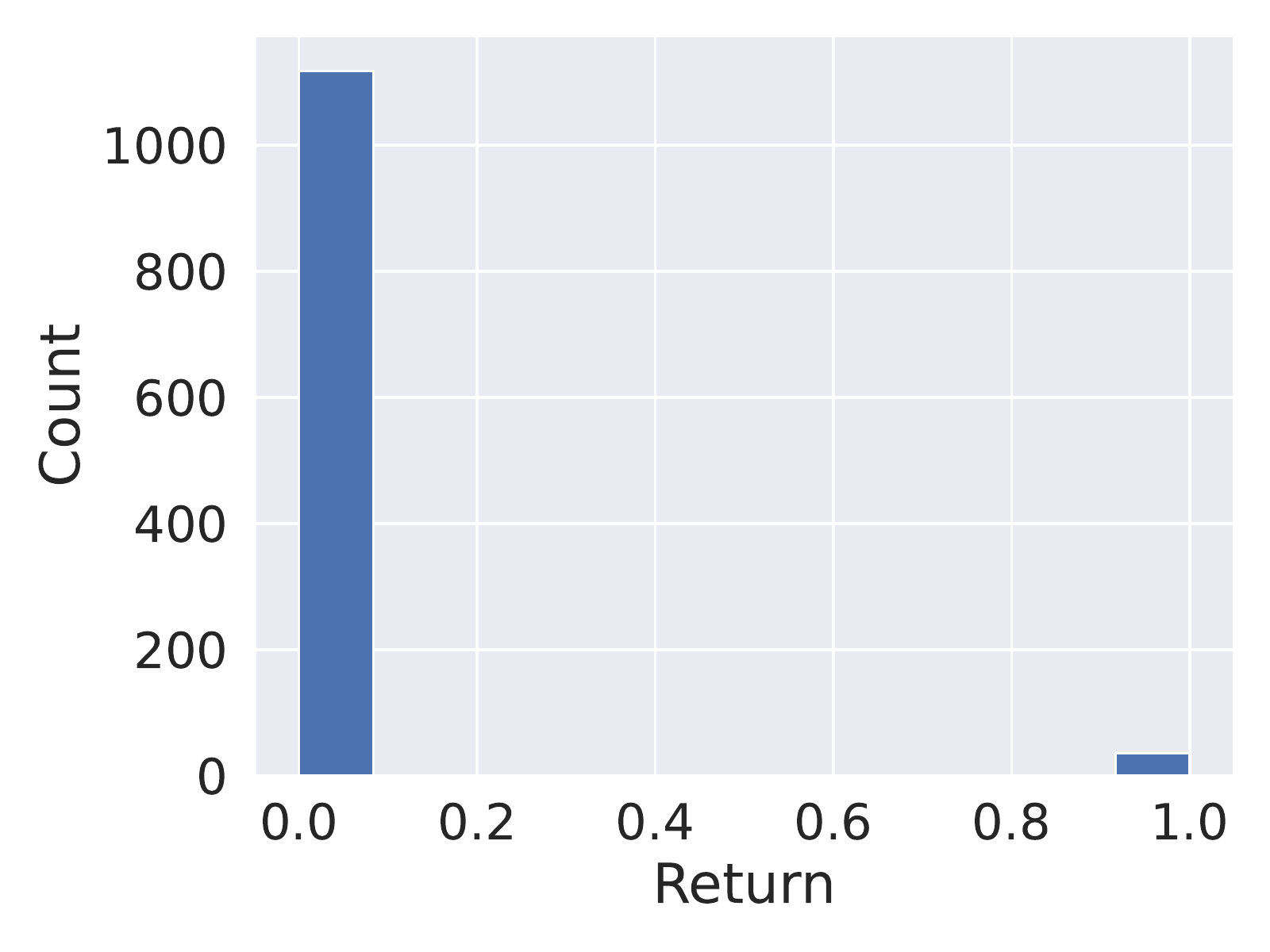}&
		\includegraphics[width=0.16\linewidth]{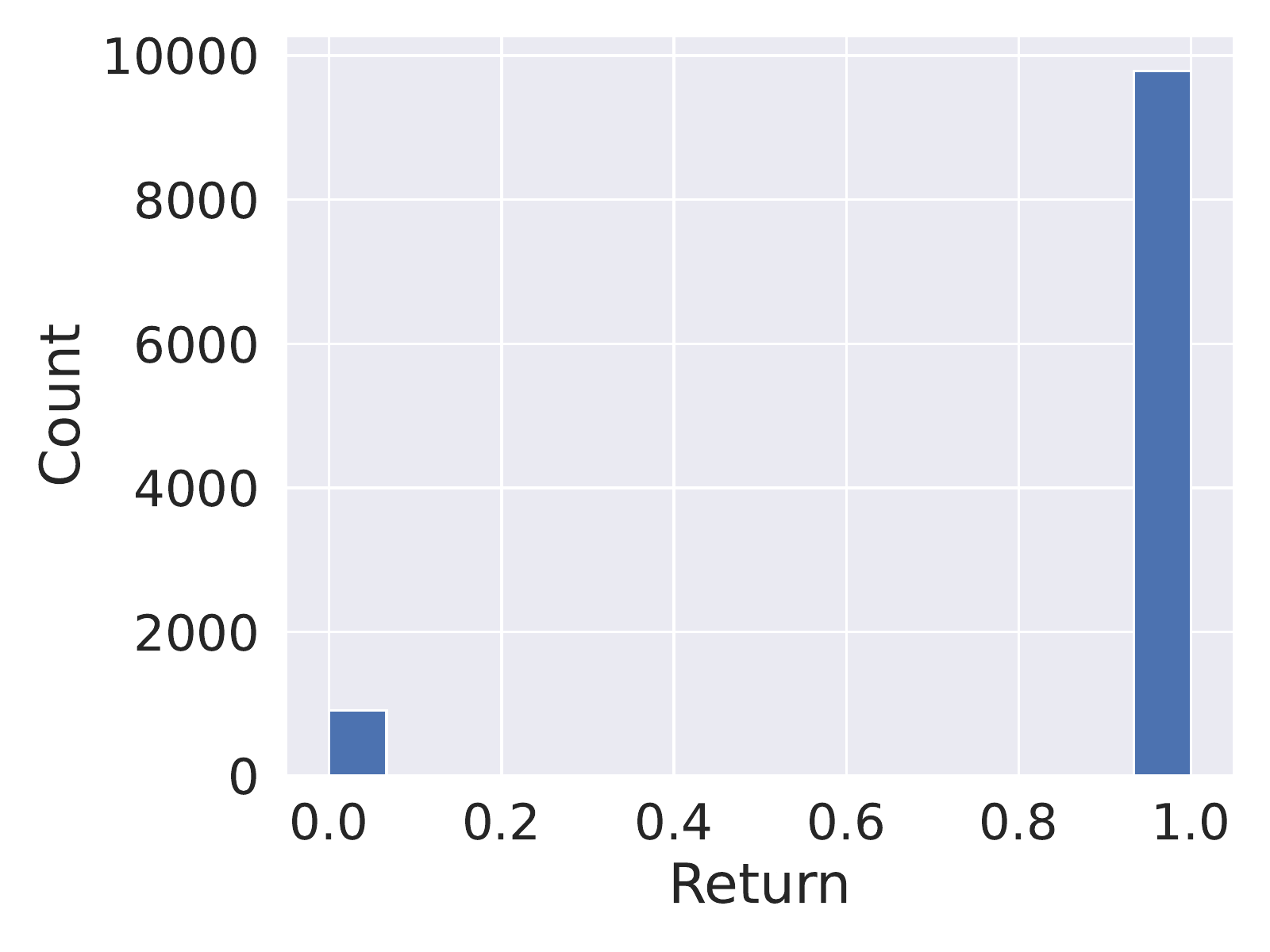} \\
		
		 antmaze-medium-diverse-v0 & antmaze-large-play-v0 & antmaze-large-diverse-v0 \\
		\includegraphics[width=0.16\linewidth]{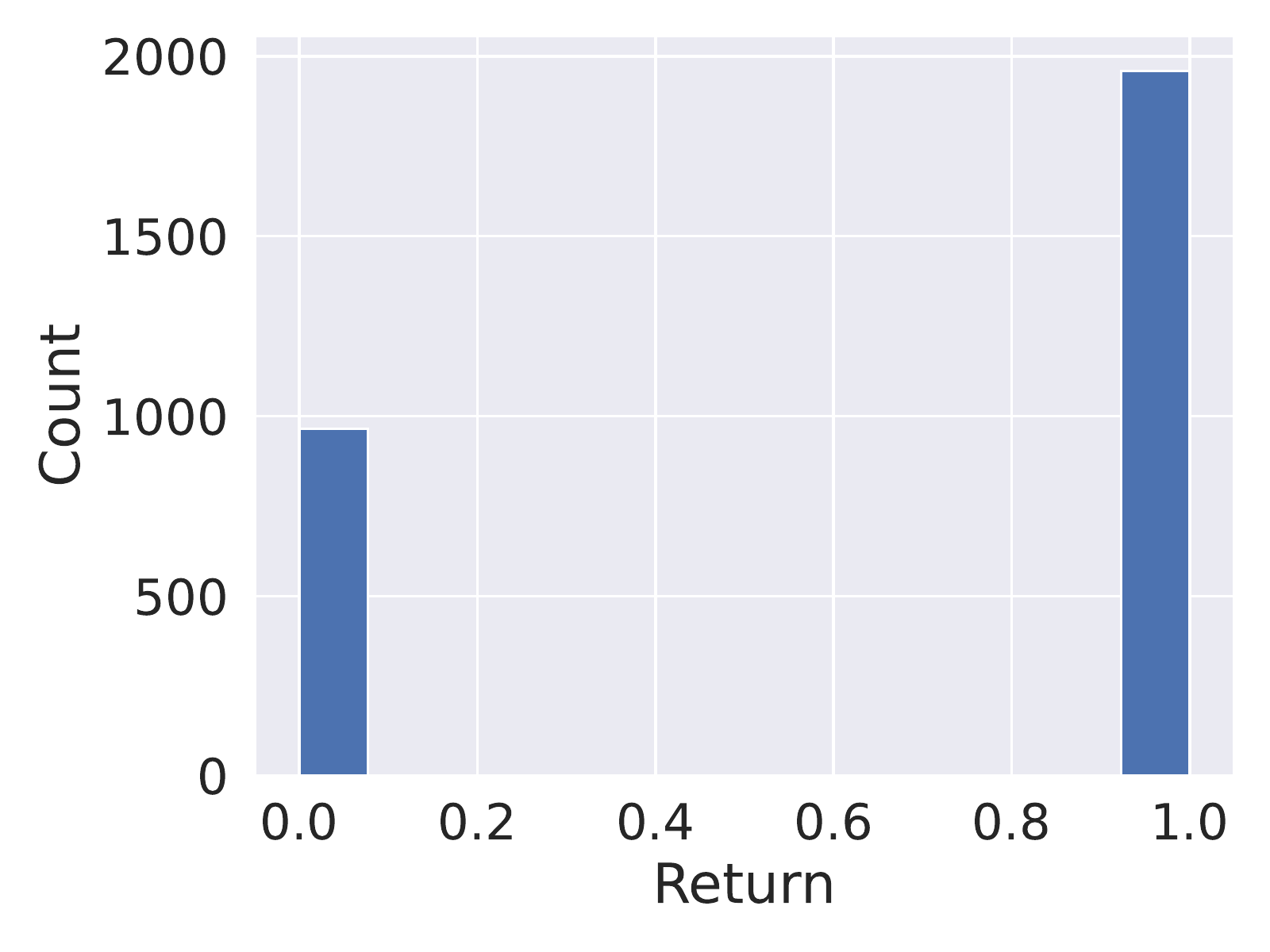}&
		\includegraphics[width=0.16\linewidth]{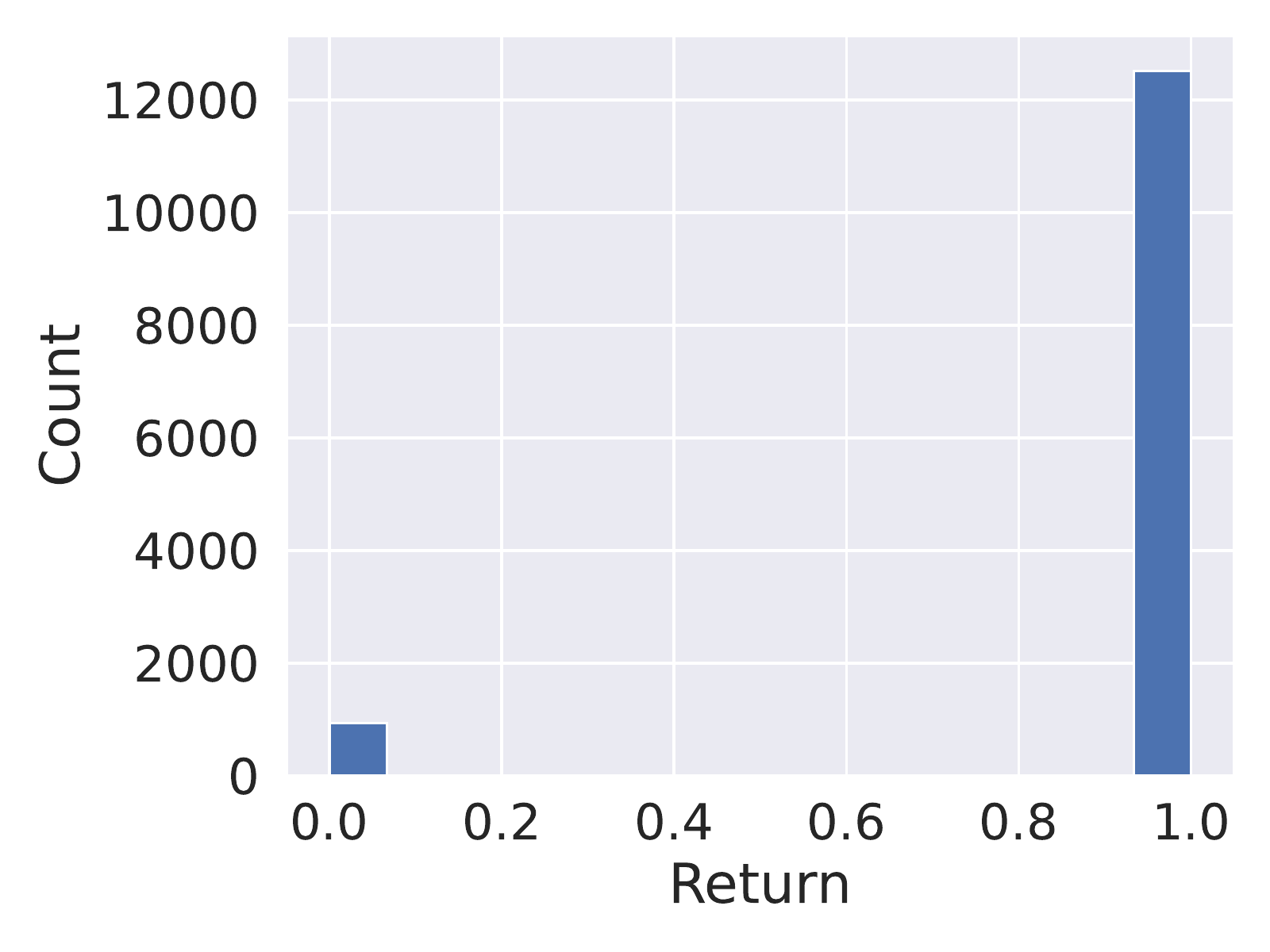}&
		\includegraphics[width=0.16\linewidth]{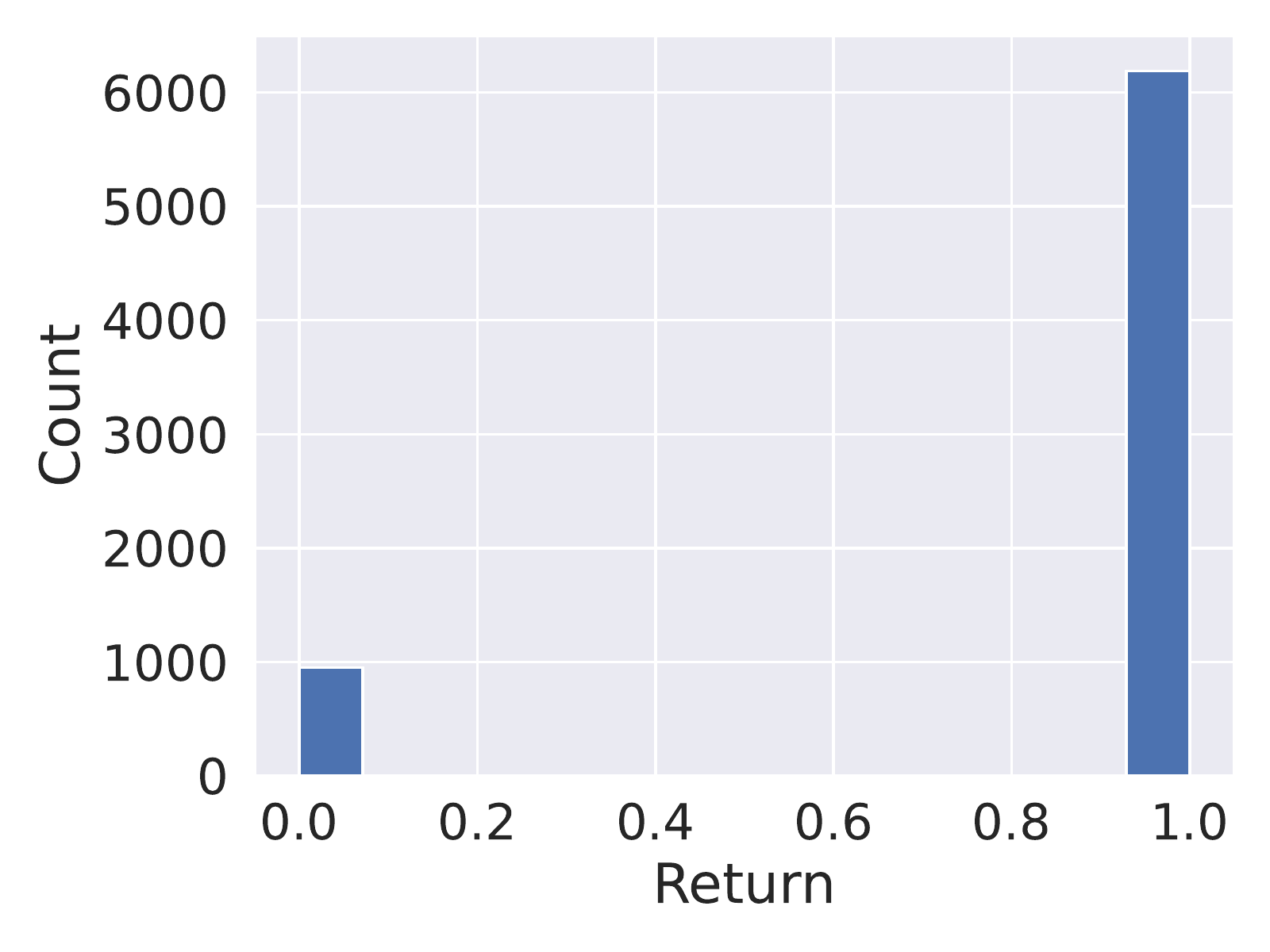} \\
		&(b) Antmanze & \\ \\
		
		 kitchen-complete-v0 & kitchen-partial-v0 & kitchen-mix-v0 \\
		
		\includegraphics[width=0.16\linewidth]{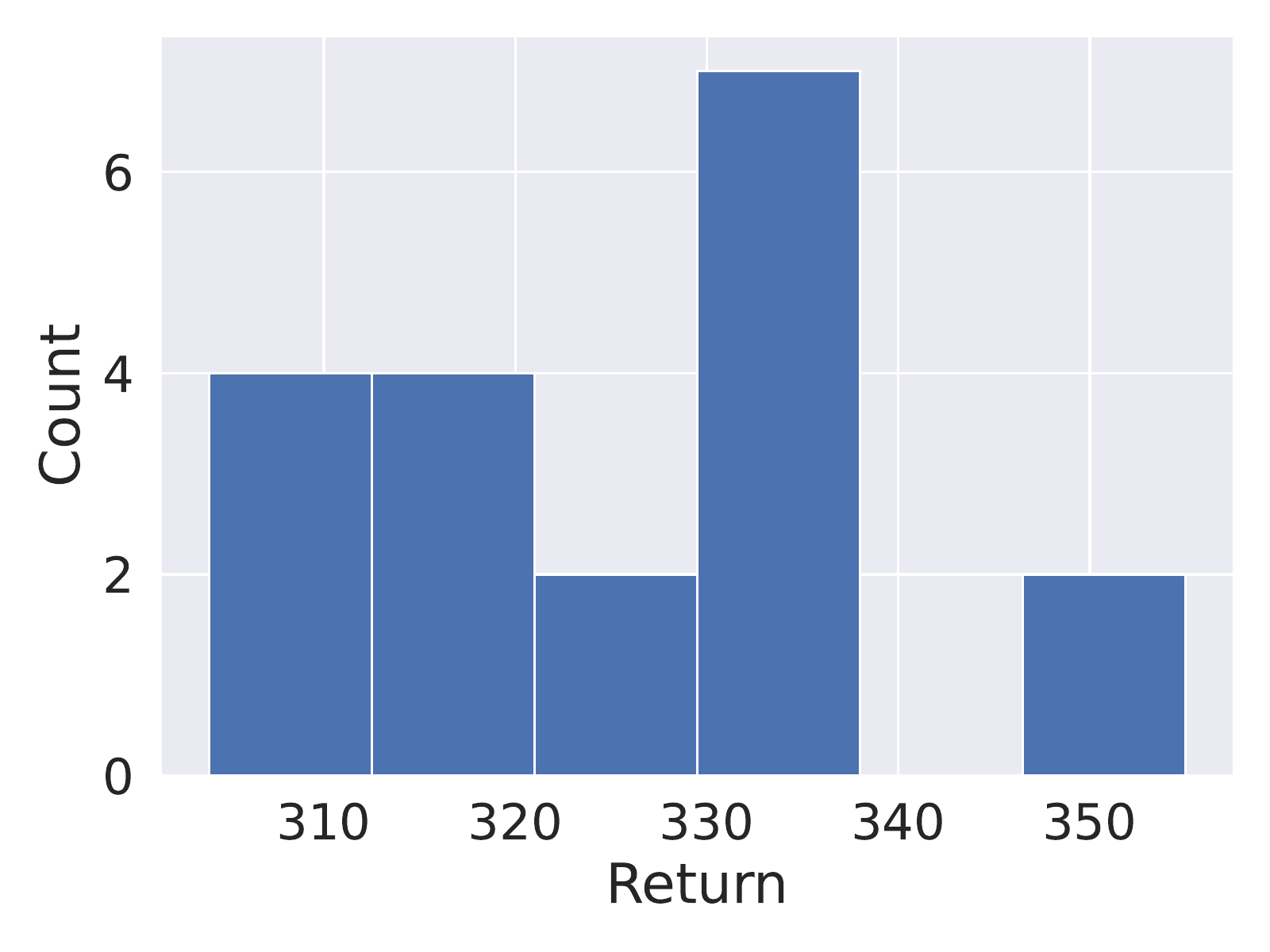}&
		\includegraphics[width=0.16\linewidth]{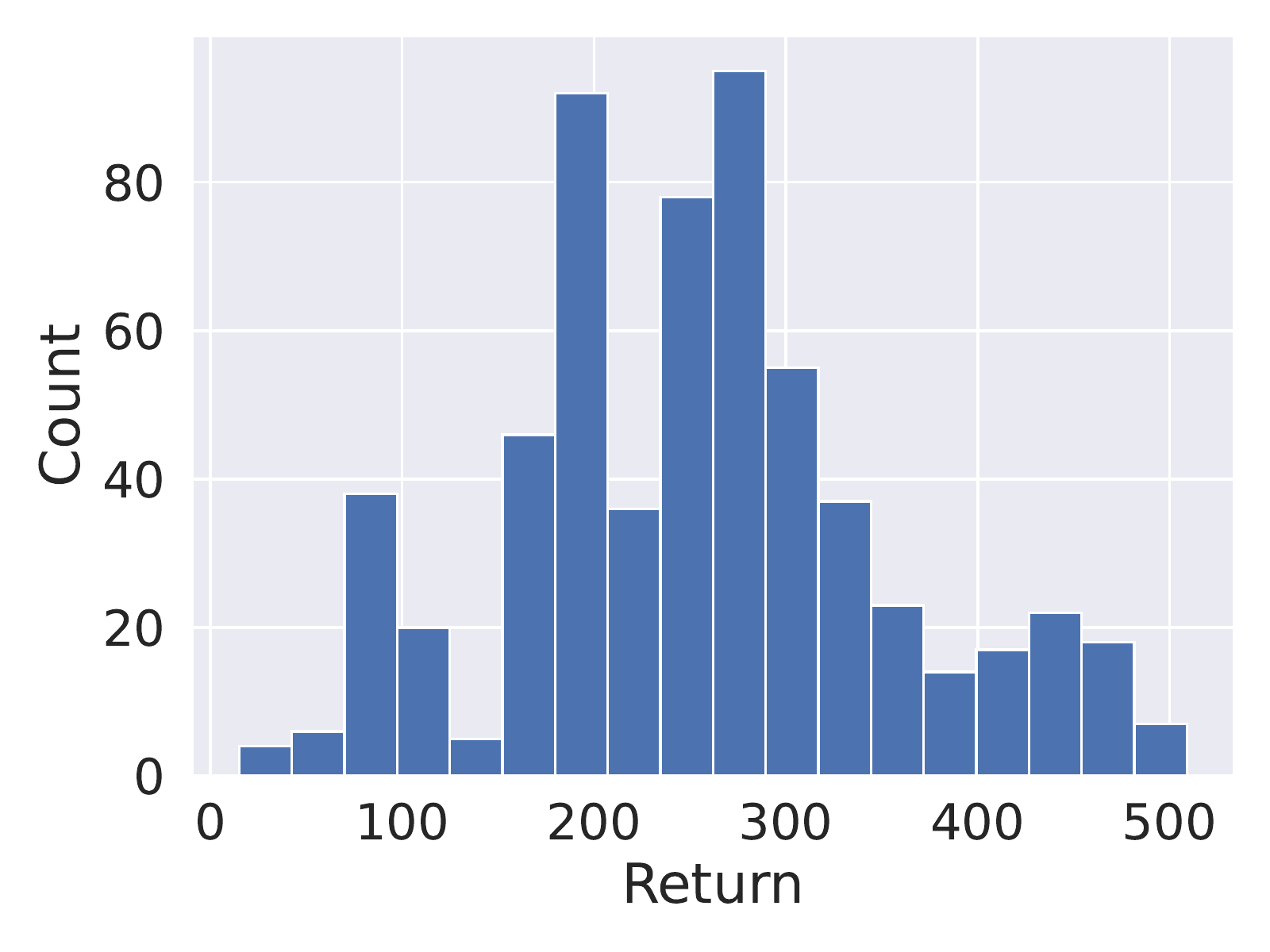}&
		\includegraphics[width=0.16\linewidth]{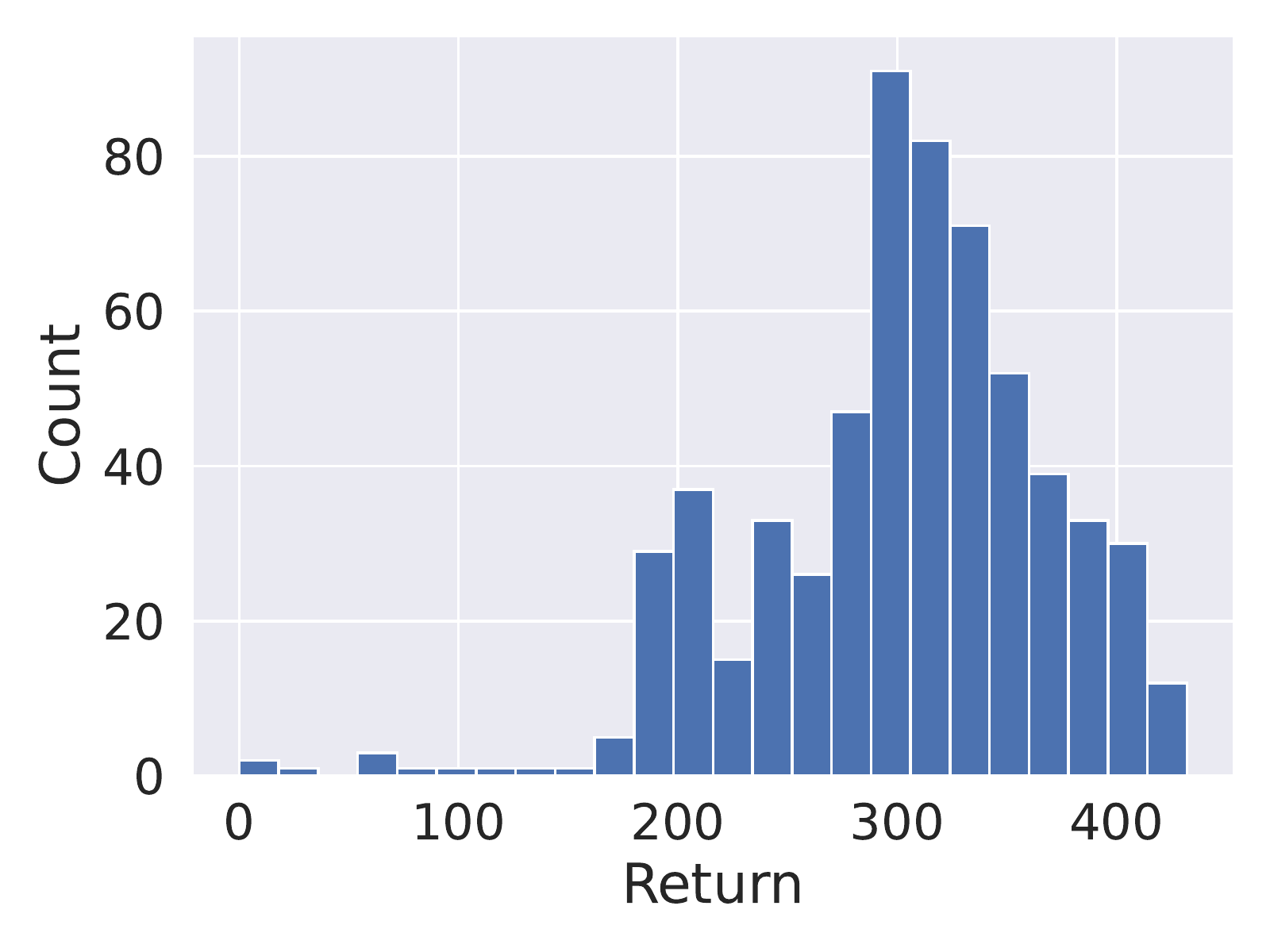}
        \\
		&(c)  Kitchen &\\ \\
		
		pen-human-v0  & hammer-human-v0 & door-human-v0  \\
		\includegraphics[width=0.16\linewidth]{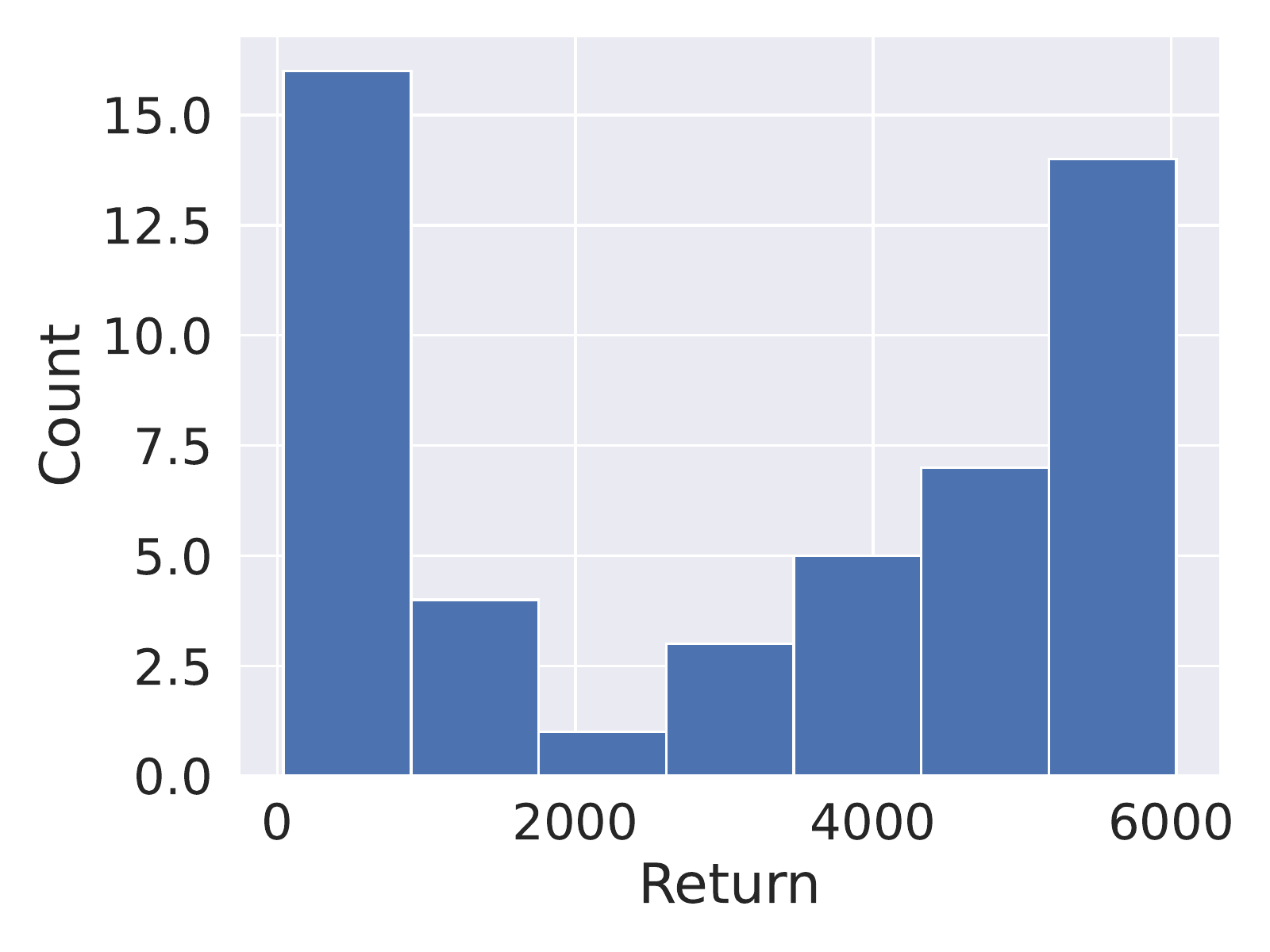} &
		\includegraphics[width=0.16\linewidth]{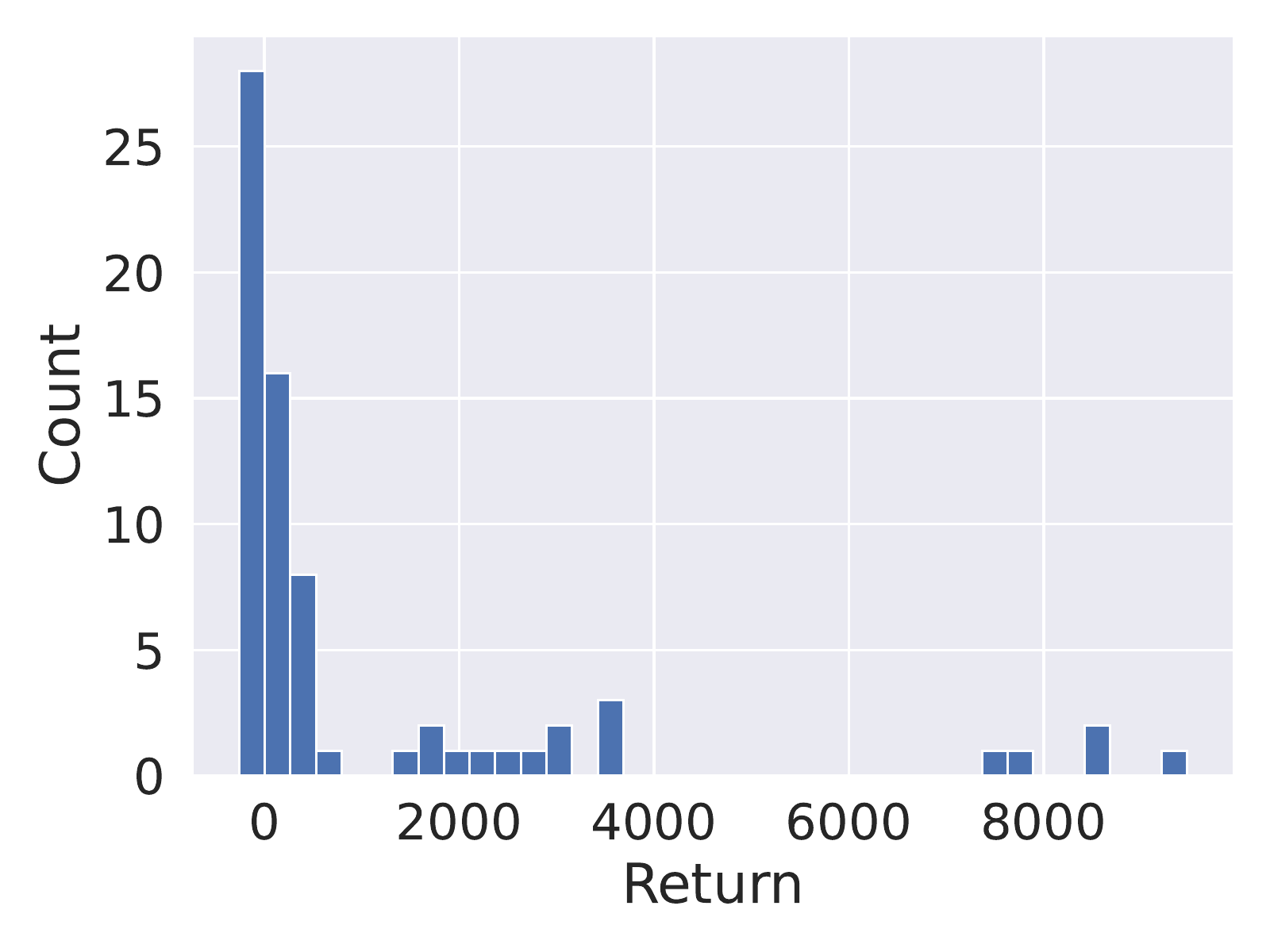}&
		\includegraphics[width=0.16\linewidth]{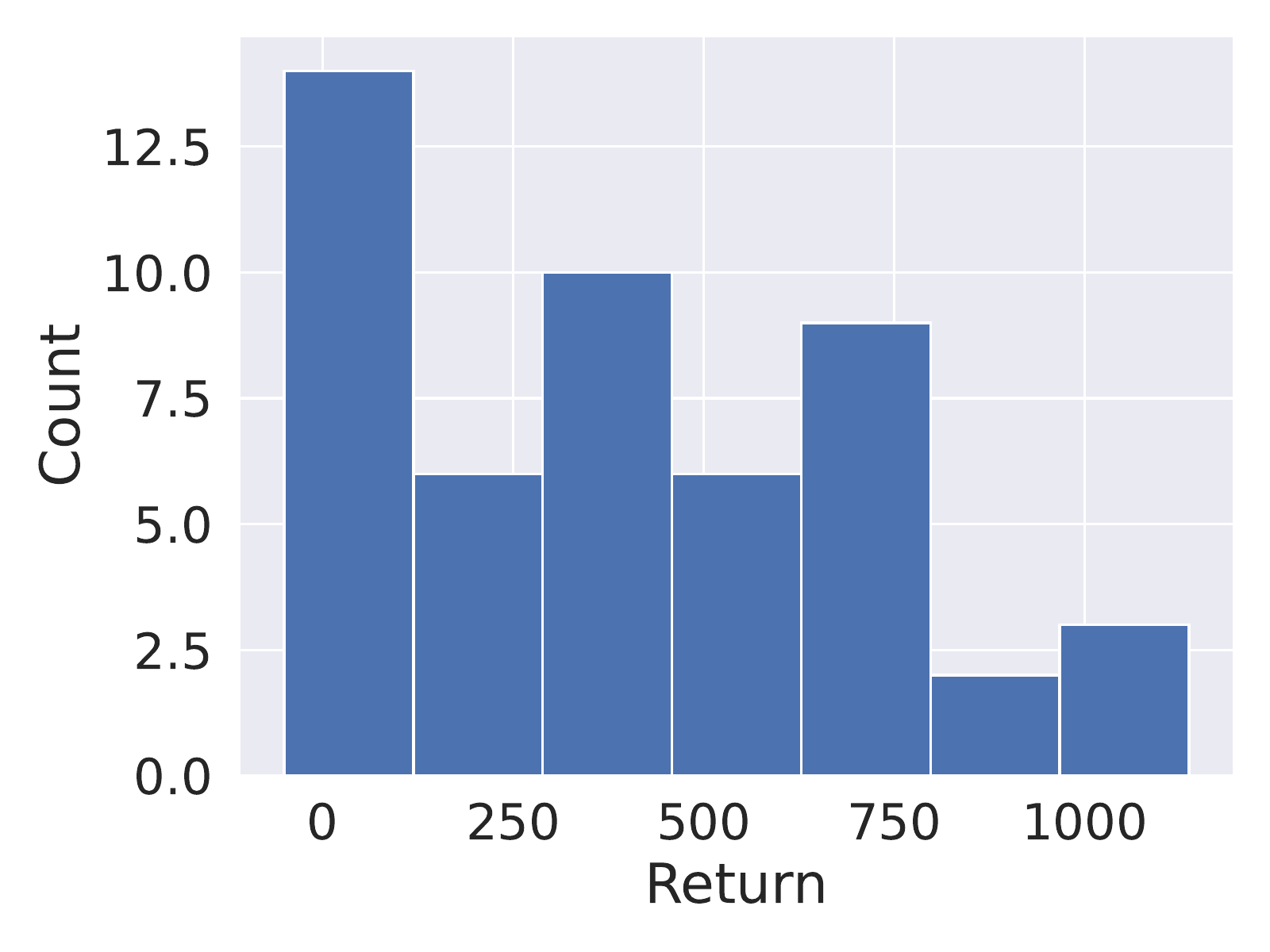} \\
		
		relocate-human-v0 & pen-cloned-v0 & 	hammer-cloned-v0 \\
		\includegraphics[width=0.16\linewidth]{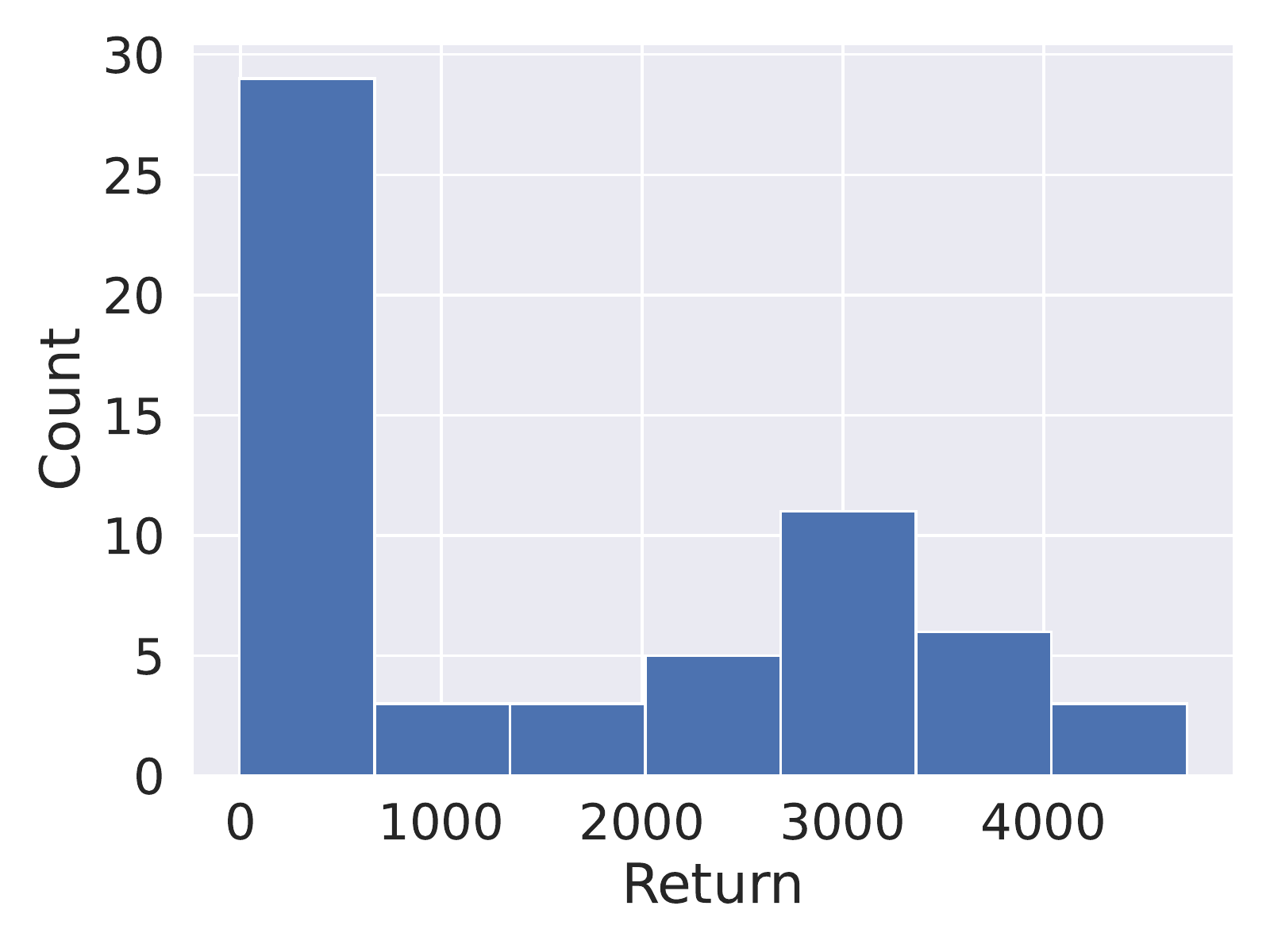}&
		\includegraphics[width=0.16\linewidth]{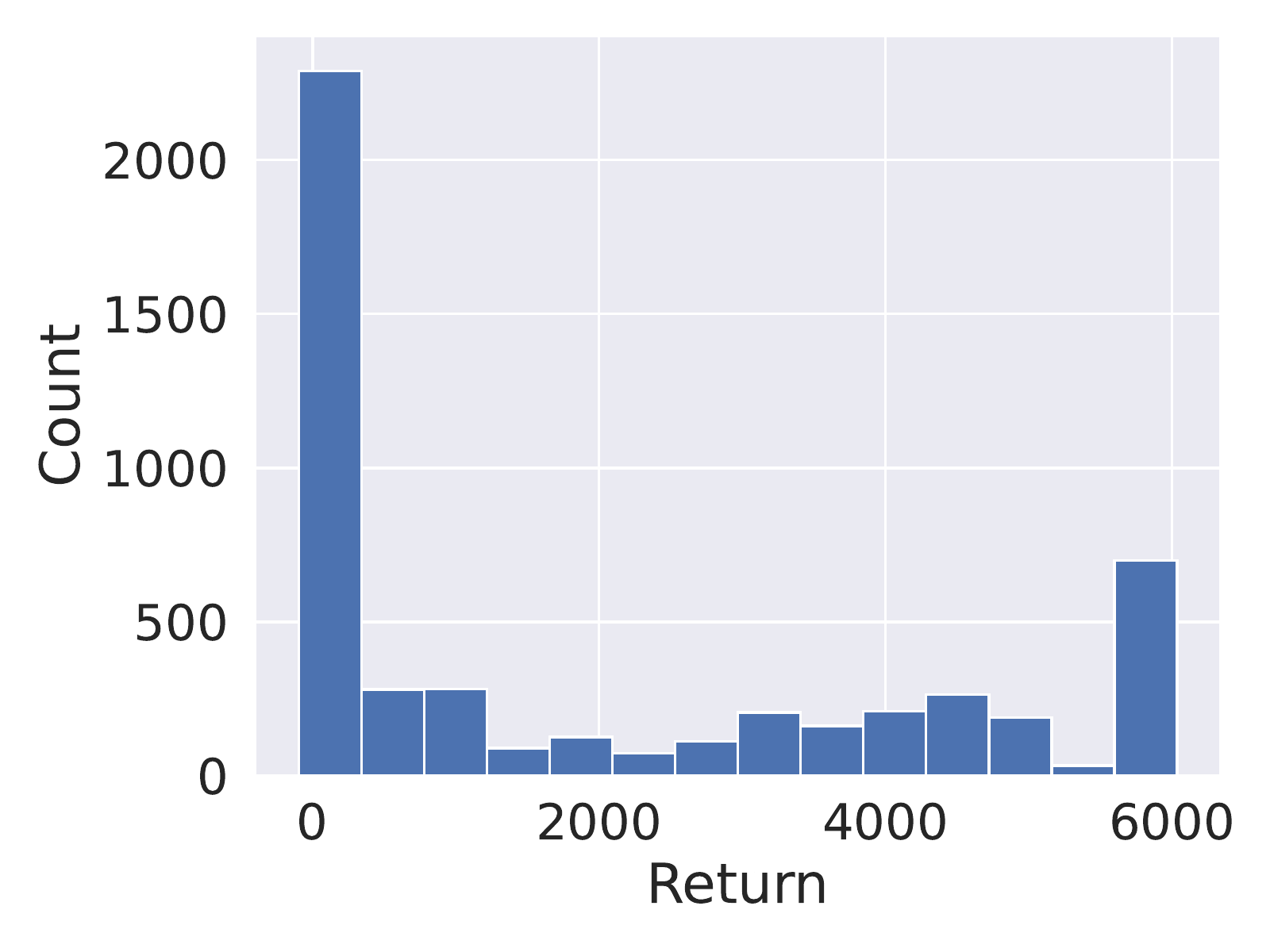} &
		\includegraphics[width=0.16\linewidth]{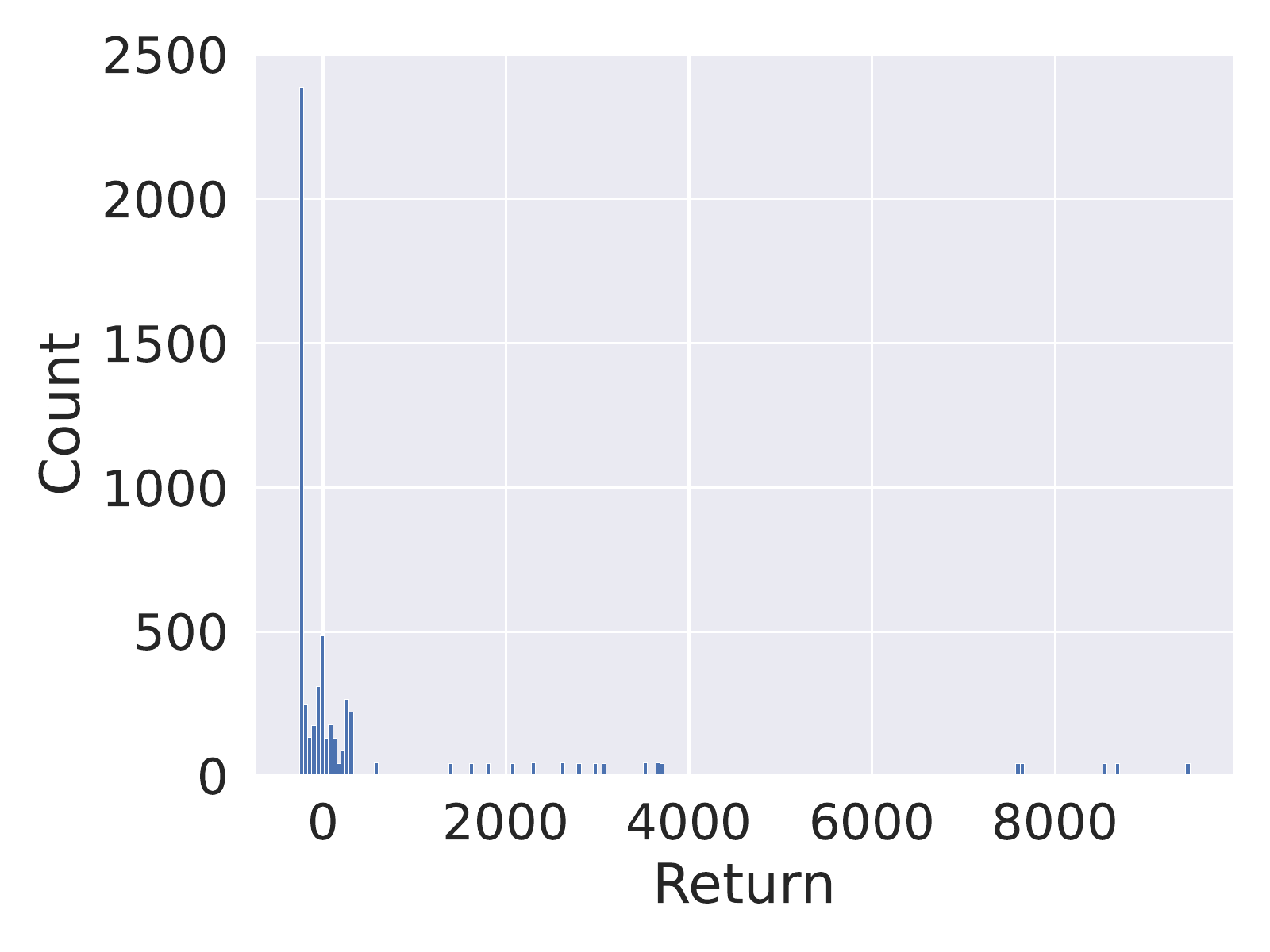} \\
		
	    door-cloned-v0 & relocate-cloned-v0 & \\

		\includegraphics[width=0.16\linewidth]{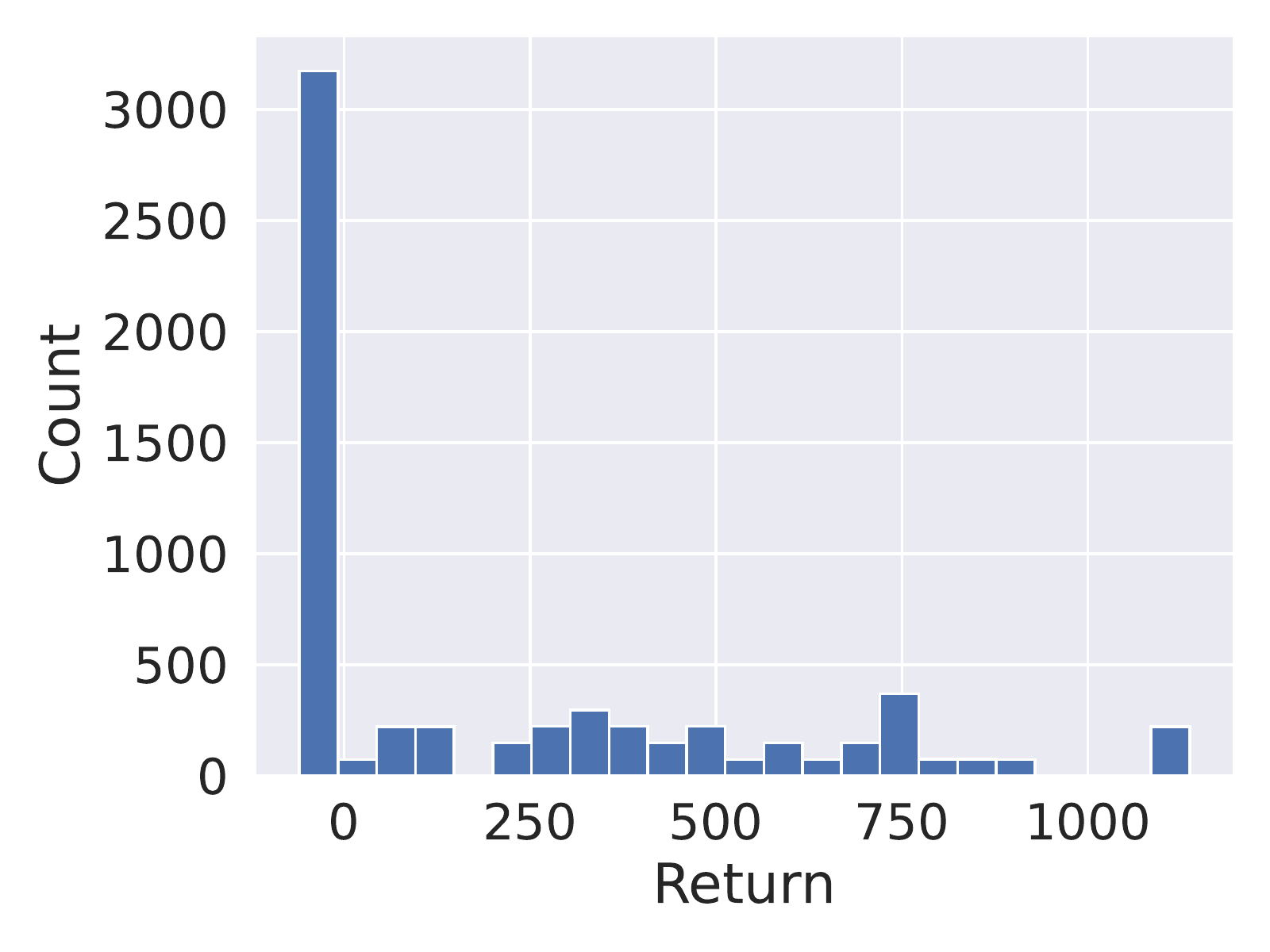}&
		\includegraphics[width=0.16\linewidth]{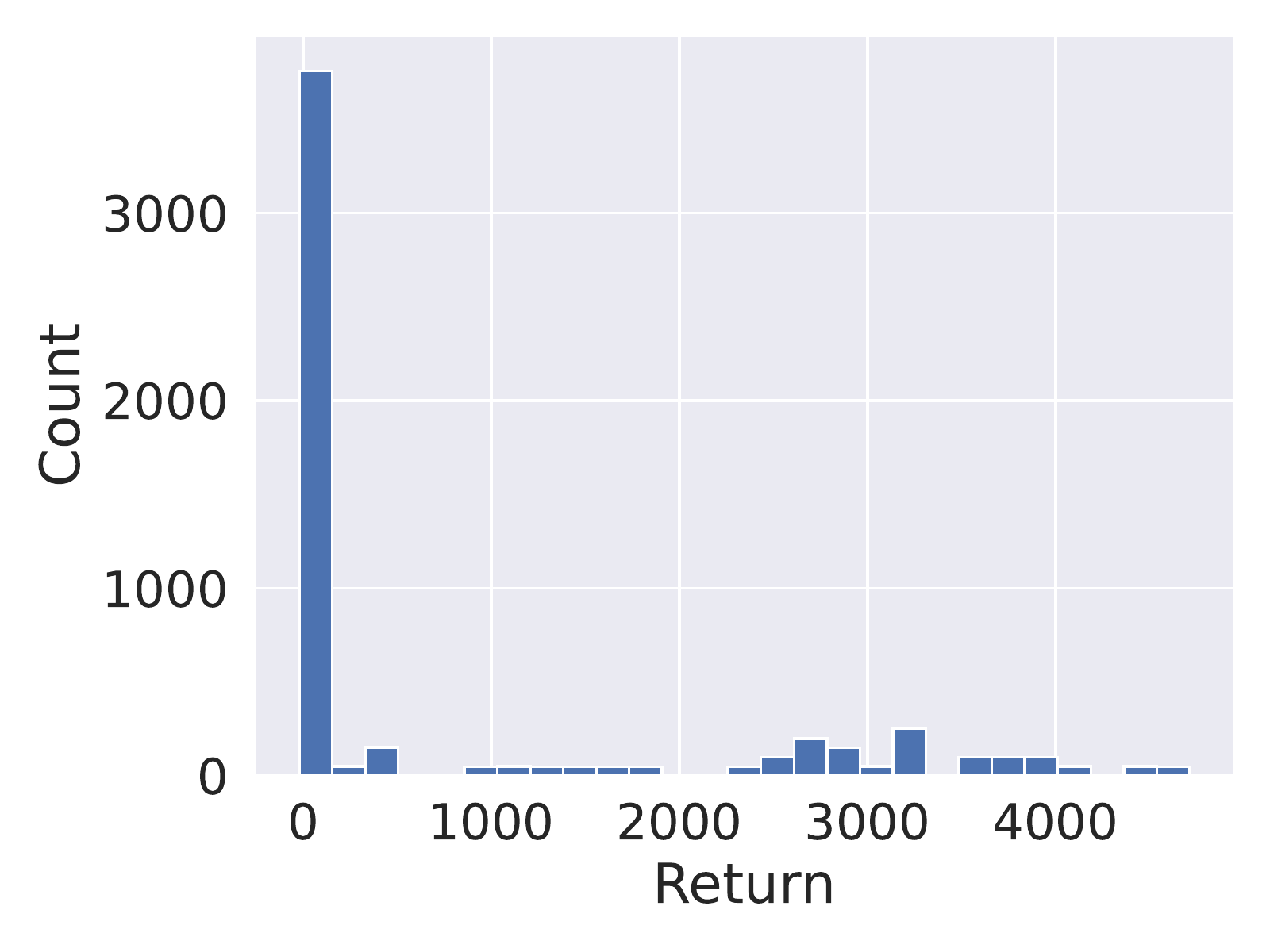}&
        \\
        &(d) Adroit&\\

	\end{tabular}
	\caption{Full Visualization of Trajectory Return Distributions. (a) Mujoco Locomotion. (b) Antmaze. (c) Kitchen. (d) Adroit.}
    \label{fig:dist_vis}
\end{figure*}

Resampling and reweighting are statistically equivalent with regards to the expected loss function. We provide implementations for both approaches in \shortname. As seen in \cref{tab:resample-and-reweight} and \cref{tab:resample-and-reweight2}, resampling and reweighting yield comparable scores on Mujoco locomotion, as well as Kitchen and Adroit tasks. These results indicate that both reweighting and resampling can successfully implement \shortname. Further, they suggest that the effectiveness of \shortname does not rely on the specific implementation, but rather arises from the prioritization of data itself.
The only exception we encountered is observed in the two Pen tasks, where for IQL with \shortalgoo, resampling performed well while reweighting was unable to achieve meaningful scores. Notably, in these two tasks, some priority weights of \shortalgoo are extremely large due to the presence of exceptionally high returns in the return distributions (see \cref{fig:dist_vis}). We hypothesize that these exceedingly large weights may alter the learning rate, thereby affecting the gradient descent process.

\begin{table}[htbp]
\centering
\small
\caption{Compare resampling and reweighting implementations for ShortName on Mujoco locomotion.}
\label{tab:resample-and-reweight}
\begin{tabular}{lcccc}
\toprule
\multirow{2}{*}{TD3+BC} & \multicolumn{2}{c}{\shortalgo} & \multicolumn{2}{c}{\shortalgoo} \\
\cmidrule(lr){2-3}\cmidrule(lr){4-5}
                         & resample & reweight & resample & reweight \\
\midrule
mujoco-v2 total          & 734.1 & 740.3 & 707.3 & 705.3 \\
\toprule
\multirow{2}{*}{IQL}    & \multicolumn{2}{c}{\shortalgo} & \multicolumn{2}{c}{\shortalgoo} \\
\cmidrule(lr){2-3}\cmidrule(lr){4-5}
                         & resample & reweight & resample & reweight \\
\midrule
mujoco-v2 total          & 727.8 & 725.4 & 726.7 & 727.0 \\  
\bottomrule
\end{tabular}
\end{table}



\begin{table}[hthp]
\small
\vspace{-1mm}
 \centering
\caption{Compare resampling and reweighting implementations of IQL (\shortalgo) on Adroit and Kitchen.}
\label{tab:resample-and-reweight2}
\begin{tabular}{ccc}
\toprule
 & reweighting  & resampling  \\
\midrule
kitchen-complete-v0   & 68.0   &     64.2    \\
kitchen-partial-v0    & 61.5 &    66.5     \\
kitchen-mixed-v0      & 45.3 &  52.1 \\ 
\midrule  
\rowcolor[HTML]{EFEFEF}
kitchen total     & 174.8 & 182.8 \\
\midrule
pen-human-v0     & 73.9 &    72.9          \\
pen-cloned-v0    & 61.6 &   61.2        \\
\bottomrule
\end{tabular}
\vspace{-6pt}
\end{table}

\bibliographystyle{IEEEtran}
\bibliography{ref.bib}

\begin{thebibliography}{10}
\providecommand{\url}[1]{#1}
\csname url@samestyle\endcsname
\providecommand{\newblock}{\relax}
\providecommand{\bibinfo}[2]{#2}
\providecommand{\BIBentrySTDinterwordspacing}{\spaceskip=0pt\relax}
\providecommand{\BIBentryALTinterwordstretchfactor}{4}
\providecommand{\BIBentryALTinterwordspacing}{\spaceskip=\fontdimen2\font plus
\BIBentryALTinterwordstretchfactor\fontdimen3\font minus \fontdimen4\font\relax}
\providecommand{\BIBforeignlanguage}[2]{{%
\expandafter\ifx\csname l@#1\endcsname\relax
\typeout{** WARNING: IEEEtran.bst: No hyphenation pattern has been}%
\typeout{** loaded for the language `#1'. Using the pattern for}%
\typeout{** the default language instead.}%
\else
\language=\csname l@#1\endcsname
\fi
#2}}
\providecommand{\BIBdecl}{\relax}
\BIBdecl

\bibitem{hong2023harnessing}
Z.-W. Hong, P.~Agrawal, R.~T. des Combes, and R.~Laroche, ``Harnessing mixed offline reinforcement learning datasets via trajectory weighting,'' in \emph{The Eleventh International Conference on Learning Representations}, 2023.

\bibitem{chen2021decision}
L.~Chen, K.~Lu, A.~Rajeswaran, K.~Lee, A.~Grover, M.~Laskin, P.~Abbeel, A.~Srinivas, and I.~Mordatch, ``Decision transformer: Reinforcement learning via sequence modeling,'' in \emph{Advances in neural information processing systems}, vol.~34, 2021, pp. 15\,084--15\,097.

\bibitem{lange2012batch}
S.~Lange, T.~Gabel, and M.~Riedmiller, ``Batch reinforcement learning,'' in \emph{Reinforcement learning}.\hskip 1em plus 0.5em minus 0.4em\relax Springer, 2012.

\bibitem{prudencio2023survey}
R.~F. Prudencio, M.~R. Maximo, and E.~L. Colombini, ``A survey on offline reinforcement learning: Taxonomy, review, and open problems,'' \emph{IEEE Transactions on Neural Networks and Learning Systems}, 2023.

\bibitem{fujimoto2019off}
S.~Fujimoto, D.~Meger, and D.~Precup, ``Off-policy deep reinforcement learning without exploration,'' in \emph{International conference on machine learning}, 2019, pp. 2052--2062.

\bibitem{jaques2019way}
N.~Jaques, A.~Ghandeharioun, J.~H. Shen, C.~Ferguson, A.~Lapedriza, N.~Jones, S.~Gu, and R.~Picard, ``Way off-policy batch deep reinforcement learning of implicit human preferences in dialog,'' \emph{arXiv preprint arXiv:1907.00456}, 2019.

\bibitem{peng2019advantage}
X.~B. Peng, A.~Kumar, G.~Zhang, and S.~Levine, ``Advantage-weighted regression: Simple and scalable off-policy reinforcement learning,'' \emph{arXiv preprint arXiv:1910.00177}, 2019.

\bibitem{wu2019behavior}
Y.~Wu, G.~Tucker, and O.~Nachum, ``Behavior regularized offline reinforcement learning,'' \emph{arXiv preprint arXiv:1911.11361}, 2019.

\bibitem{kumar2019stabilizing}
A.~Kumar, J.~Fu, M.~Soh, G.~Tucker, and S.~Levine, ``Stabilizing off-policy q-learning via bootstrapping error reduction,'' in \emph{Advances in neural information processing systems}, vol.~32, 2019.

\bibitem{td3+bc}
S.~Fujimoto and S.~S. Gu, ``A minimalist approach to offline reinforcement learning,'' in \emph{Proceedings of the 34th Conference on Neural Information Processing Systems (NeurIPS)}, vol.~34, 2021, pp. 20\,132--20\,145.

\bibitem{PER}
T.~Schaul, J.~Quan, I.~Antonoglou, and D.~Silver, ``Prioritized experience replay,'' in \emph{International Conference on Learning Representations}, 2016.

\bibitem{kumar2020discor}
A.~Kumar, A.~Gupta, and S.~Levine, ``Discor: Corrective feedback in reinforcement learning via distribution correction,'' in \emph{Advances in Neural Information Processing Systems}, vol.~33, 2020, pp. 18\,560--18\,572.

\bibitem{sinha2022lfiw}
S.~Sinha, J.~Song, A.~Garg, and S.~Ermon, ``Experience replay with likelihood-free importance weights,'' in \emph{Learning for Dynamics and Control Conference}, 2022.

\bibitem{brittain2019pser}
M.~Brittain, J.~Bertram, X.~Yang, and P.~Wei, ``Prioritized sequence experience replay,'' \emph{arXiv preprint arXiv:1905.12726}, 2019.

\bibitem{wang2020striving}
C.~Wang, Y.~Wu, Q.~Vuong, and K.~Ross, ``Striving for simplicity and performance in off-policy drl: Output normalization and non-uniform sampling,'' in \emph{International Conference on Machine Learning}, 2020, pp. 10\,070--10\,080.

\bibitem{liu2021regret}
X.-H. Liu, Z.~Xue, J.-C. Pang, S.~Jiang, F.~Xu, and Y.~Yu, ``Regret minimization experience replay in off-policy reinforcement learning,'' in \emph{Advances in Neural Information Processing Systems}, 2021.

\bibitem{oh2018self}
J.~Oh, Y.~Guo, S.~Singh, and H.~Lee, ``Self-imitation learning,'' in \emph{International conference on machine learning}, 2018, pp. 3878--3887.

\bibitem{singh2022offline}
A.~Singh, A.~Kumar, Q.~Vuong, Y.~Chebotar, and S.~Levine, ``Re{DS}: Offline {RL} with heteroskedastic datasets via support constraints,'' in \emph{Thirty-seventh Conference on Neural Information Processing Systems}, 2023.

\bibitem{chen2020bail}
X.~Chen, Z.~Zhou, Z.~Wang, C.~Wang, Y.~Wu, and K.~Ross, ``Bail: Best-action imitation learning for batch deep reinforcement learning,'' in \emph{Advances in Neural Information Processing Systems}, vol.~33, 2020, pp. 18\,353--18\,363.

\bibitem{wang2018exponentially}
Q.~Wang, J.~Xiong, L.~Han, H.~Liu, T.~Zhang \emph{et~al.}, ``Exponentially weighted imitation learning for batched historical data,'' in \emph{Advances in Neural Information Processing Systems}, vol.~31, 2018.

\bibitem{liu2021curriculum}
M.~Liu, H.~Zhao, Z.~Yang, J.~Shen, W.~Zhang, L.~Zhao, and T.-Y. Liu, ``Curriculum offline imitating learning,'' in \emph{Advances in Neural Information Processing Systems}, vol.~34, 2021, pp. 6266--6277.

\bibitem{fu2019diagnosing}
J.~Fu, A.~Kumar, M.~Soh, and S.~Levine, ``Diagnosing bottlenecks in deep q-learning algorithms,'' in \emph{International Conference on Machine Learning}, 2019, pp. 2021--2030.

\bibitem{van2018deadly_triad}
H.~Van~Hasselt, Y.~Doron, F.~Strub, M.~Hessel, N.~Sonnerat, and J.~Modayil, ``Deep reinforcement learning and the deadly triad,'' \emph{arXiv preprint arXiv:1812.02648}, 2018.

\bibitem{tsitsiklis1996deadly_tirad}
J.~Tsitsiklis and B.~Van~Roy, ``An analysis of temporal-difference learning with function approximationtechnical,'' \emph{Rep. LIDS-P-2322). Lab. Inf. Decis. Syst. Massachusetts Inst. Technol. Tech. Rep}, 1996.

\bibitem{CQL}
A.~Kumar, A.~Zhou, G.~Tucker, and S.~Levine, ``Conservative q-learning for offline reinforcement learning,'' in \emph{Advances in Neural Information Processing Systems}, vol.~33, 2020, pp. 1179--1191.

\bibitem{IQL}
I.~Kostrikov, A.~Nair, and S.~Levine, ``Offline reinforcement learning with implicit q-learning,'' \emph{arXiv preprint arXiv:2110.06169}, 2021.

\bibitem{onestep}
D.~Brandfonbrener, W.~Whitney, R.~Ranganath, and J.~Bruna, ``Offline rl without off-policy evaluation,'' in \emph{Advances in neural information processing systems}, vol.~34, 2021, pp. 4933--4946.

\bibitem{gym}
G.~Brockman, V.~Cheung, L.~Pettersson, J.~Schneider, J.~Schulman, J.~Tang, and W.~Zaremba, ``Openai gym,'' \emph{arXiv preprint arXiv:1606.01540}, 2016.

\bibitem{fu2020d4rl}
J.~Fu, A.~Kumar, O.~Nachum, G.~Tucker, and S.~Levine, ``D4rl: Datasets for deep data-driven reinforcement learning,'' 2020.

\bibitem{yue2022boosting}
Y.~Yue, B.~Kang, X.~Ma, Z.~Xu, G.~Huang, and S.~YAN, ``Boosting offline reinforcement learning via data rebalancing,'' in \emph{3rd NeurIPS Offline RL Workshop'}, 2022.

\bibitem{kakade2002approximately}
S.~Kakade and J.~Langford, ``Approximately optimal approximate reinforcement learning,'' in \emph{Proceedings of the Nineteenth International Conference on Machine Learning}, 2002, pp. 267--274.

\bibitem{sutton2018reinforcement}
R.~S. Sutton and A.~G. Barto, \emph{Reinforcement learning: An introduction}.\hskip 1em plus 0.5em minus 0.4em\relax MIT press, 2018.

\bibitem{wang2023diffusion}
Z.~Wang, J.~J. Hunt, and M.~Zhou, ``Diffusion policies as an expressive policy class for offline reinforcement learning,'' in \emph{The Eleventh International Conference on Learning Representations}, 2023.

\bibitem{yue2023understanding}
Y.~Yue, R.~Lu, B.~Kang, S.~Song, and G.~Huang, ``Understanding, predicting and better resolving q-value divergence in offline-{RL},'' in \emph{Thirty-seventh Conference on Neural Information Processing Systems}, 2023.

\bibitem{zhang2023efficient}
L.~Zhang, Y.~Feng, R.~Wang, Y.~Xu, N.~Xu, Z.~Liu, and H.~Du, ``Efficient experience replay architecture for offline reinforcement learning,'' \emph{Robotic Intelligence and Automation}, 2023.

\bibitem{wang2020critic}
Z.~Wang, A.~Novikov, K.~Zolna, J.~S. Merel, J.~T. Springenberg, S.~E. Reed, B.~Shahriari, N.~Siegel, C.~Gulcehre, N.~Heess \emph{et~al.}, ``Critic regularized regression,'' in \emph{Advances in Neural Information Processing Systems}, vol.~33, 2020, pp. 7768--7778.

\bibitem{nair2020awac}
A.~Nair, A.~Gupta, M.~Dalal, and S.~Levine, ``Awac: Accelerating online reinforcement learning with offline datasets,'' \emph{arXiv preprint arXiv:2006.09359}, 2020.

\bibitem{chen2022offline}
H.~Chen, C.~Lu, C.~Ying, H.~Su, and J.~Zhu, ``Offline reinforcement learning via high-fidelity generative behavior modeling,'' in \emph{The Eleventh International Conference on Learning Representations}, 2023.

\bibitem{yang2023hundreds}
Q.~Yang, S.~Wang, Q.~Zhang, G.~Huang, and S.~Song, ``Hundreds guide millions: Adaptive offline reinforcement learning with expert guidance,'' \emph{IEEE Transactions on Neural Networks and Learning Systems}, 2023.

\bibitem{buckman2020pessimism}
J.~Buckman, C.~Gelada, and M.~G. Bellemare, ``The importance of pessimism in fixed-dataset policy optimization,'' in \emph{International Conference on Learning Representations}, 2021.

\bibitem{yu2021combo}
T.~Yu, A.~Kumar, R.~Rafailov, A.~Rajeswaran, S.~Levine, and C.~Finn, ``Combo: Conservative offline model-based policy optimization,'' in \emph{Advances in neural information processing systems}, vol.~34, 2021, pp. 28\,954--28\,967.

\bibitem{sac-n}
G.~An, S.~Moon, J.-H. Kim, and H.~O. Song, ``Uncertainty-based offline reinforcement learning with diversified q-ensemble,'' in \emph{Advances in neural information processing systems}, vol.~34, 2021, pp. 7436--7447.

\bibitem{lb-sac}
A.~Nikulin, V.~Kurenkov, D.~Tarasov, D.~Akimov, and S.~Kolesnikov, ``Q-ensemble for offline rl: Don't scale the ensemble, scale the batch size,'' \emph{arXiv preprint arXiv:2211.11092}, 2022.

\bibitem{ghasemipour2022so}
K.~Ghasemipour, S.~S. Gu, and O.~Nachum, ``Why so pessimistic? estimating uncertainties for offline rl through ensembles, and why their independence matters,'' in \emph{Advances in Neural Information Processing Systems}, vol.~35, 2022, pp. 18\,267--18\,281.

\bibitem{ma2021conservative}
Y.~Ma, D.~Jayaraman, and O.~Bastani, ``Conservative offline distributional reinforcement learning,'' in \emph{Advances in neural information processing systems}, vol.~34, 2021, pp. 19\,235--19\,247.

\bibitem{lee2022offline}
S.~Lee, Y.~Seo, K.~Lee, P.~Abbeel, and J.~Shin, ``Offline-to-online reinforcement learning via balanced replay and pessimistic q-ensemble,'' in \emph{Conference on Robot Learning}, 2022.

\bibitem{pong2022offline}
V.~H. Pong, A.~V. Nair, L.~M. Smith, C.~Huang, and S.~Levine, ``Offline meta-reinforcement learning with online self-supervision,'' in \emph{International Conference on Machine Learning}, 2022.

\bibitem{Double-Q}
H.~Hasselt, ``Double q-learning,'' in \emph{Advances in neural information processing systems}, vol.~23, 2010.

\bibitem{td3}
S.~Fujimoto, H.~Hoof, and D.~Meger, ``Addressing function approximation error in actor-critic methods,'' in \emph{International conference on machine learning}, 2018, pp. 1587--1596.

\end{thebibliography}

\end{document}